\documentclass[a4paper,fleqn]{cas-sc}

\usepackage[numbers]{natbib}
\usepackage{amsmath,amssymb,amsfonts}
\usepackage{bm}
\usepackage{algorithm}
\usepackage{algpseudocode}
\usepackage[caption=false]{subfig}

\begin{document}

\shorttitle{History-aware adaptive ROMs via iSVD}
\shortauthors{Hedayat et al.}

\title[mode=title]{History-aware adaptive reduced-order models via incremental singular value decomposition}

\author[1]{Amirpasha Hedayat}[orcid=0009-0001-5229-9396]
\cormark[1]
\ead{ahedayat@umich.edu}

\author[2]{Ali Mohaghegh}[orcid=0009-0009-9321-5134]
\ead{ali.mohaghegh@ku.edu}

\author[3]{Laura Balzano}[orcid=0000-0003-2914-123X]
\ead{girasole@umich.edu}

\author[2]{Cheng Huang}[orcid=0000-0000-0000-0000]
\ead{chenghuang@ku.edu}

\author[1]{Karthik Duraisamy}[orcid=0000-0000-0000-0000]
\ead{kdur@umich.edu}

\affiliation[1]{organization={Department of Aerospace Engineering},
    addressline={University of Michigan},
    city={Ann Arbor},
    state={MI},
    country={USA}}

\affiliation[2]{organization={Department of Aerospace Engineering},
    addressline={University of Kansas},
    city={Lawrence},
    state={KS},
    country={USA}}

\affiliation[3]{organization={Department of Electrical Engineering and Computer Science},
    addressline={University of Michigan},
    city={Ann Arbor},
    state={MI},
    country={USA}}

\cortext[1]{Corresponding author}

\begin{abstract}
Reduced-order models (ROMs) offer the potential to accelerate the simulation of complex high-dimensional dynamical systems, but their predictive accuracy often deteriorates once the online dynamics depart from the regime represented in the offline training data. This limitation has motivated growing interest in adaptive ROMs that can update their low-dimensional representation during the online prediction. In this work, we develop a projection-based adaptive reduced-order modeling framework based on incremental singular value decomposition (iSVD), in which occasional evaluations of the full-order operator are used to update the reduced basis. The ROMs considered here are intrusive and fully parameterized by their basis, so each basis update naturally propagates to the remaining ingredients of the model, including the reduced operators and hyper-reduction machinery. The model retains an encoded representation of the history of the observed dynamics in its evolving singular structure, and  in this sense is \emph{history-aware}. We study the proposed framework on three nonlinear problems of increasing complexity: the one-dimensional viscous Burgers equation, the Sod shock tube, and a highly stiff one-dimensional ten-species rotating detonation engine (RDE). The Burgers problem is used to develop the method in detail and to compare iSVD with several alternative basis adaptation rules, showing that history-aware updates substantially outperform instantaneous updates and that iSVD provides the strongest overall performance. The Sod and RDE cases then demonstrate that these advantages persist in more challenging compressible flow settings. In particular, for the RDE problem, the proposed iSVD adaptive ROM improves upon the current state-of-the-art adaptive ROM baseline in both predictive accuracy and computational efficiency. A computational cost analysis further shows that the dominant online cost of the adaptive framework comes from the interactions with the full-order operator to obtain the correction snapshot, and the iSVD basis update itself remains negligible. These results identify iSVD as a particularly effective mechanism for online learning of reduced subspaces and suggest a path toward more reliable ROMs that remain predictive over long horizons, including cases where the prediction interval is several orders of magnitude longer than the initial training window.
\end{abstract}


\maketitle

\section{Introduction}
\label{sec:introduction}

High-fidelity simulations of large-scale nonlinear dynamical systems have become indispensable in modern science and engineering, but their computational cost often makes them impractical to embed directly in workflows such as design optimization, uncertainty quantification, control, and real-time decision making in digital twins~\citep{benner2015survey, antoulas2020interpolatory, rowley2017modelreduction}. Reduced-order models (ROMs) attempt to address this limitation by constructing a compact, low-dimensional surrogate that preserves the essential dynamics of the full-order model (FOM) while enabling orders-of-magnitude acceleration in computational time~\citep{carlberg2017galerkin, peherstorfer2016data, huang2022multiscale}. As a result, ROMs have been successfully applied across a wide range of problems in fluid dynamics, combustion, and control-oriented engineering~\citep{rowley2004modelreduction, barone2009stable, arnold2022les}.

Despite their success, most projection-based ROMs are developed in a fundamentally \emph{static} fashion. In this approach, an offline training phase collects FOM snapshots over a prescribed parameter or time window, extracts a low-dimensional trial subspace using proper orthogonal decomposition (POD)~\citep{lumley1967structure, sirovich1987turbulence, holmes1996turbulence} or related techniques, constructs the reduced operators in that subspace, and then deploys the resulting model online in frozen form~\citep{benner2015survey, carlberg2017galerkin, peherstorfer2016data}. This development paradigm mirrors the classical machine learning setting of training offline and deploying online, and it inherits the same fundamental limitation in that the model is only predictive as long as the online trajectory remains \emph{in-distribution} with respect to the offline training manifold. In this sense, static ROMs effectively act as \emph{interpolators} rather than genuine \emph{predictors}, and their accuracy degrades rapidly once the dynamics leave the training regime~\citep{hedayat2026toward, huang2023predictive}. This situation is quite common in systems with constantly evolving dynamics, such as convection-dominated flows, hypersonics, shock-dominated compressible flows, combustion, and detonations, all of which exhibit a slow decay of the Kolmogorov \(N\)-width and therefore cannot be captured efficiently by any fixed low-dimensional subspace~\citep{peherstorfer2022breaking}. In such cases, practitioners are often forced to fall back on the expensive full-order solver, which defeats the very purpose of model reduction.

A substantial body of work has sought to restore predictive capability in ROMs beyond the training regime, and these efforts can be broadly grouped into three directions. The first replaces the linear trial subspace with a nonlinear embedding, typically through deep convolutional autoencoders and related neural network architectures~\citep{lee2020modelreduction, fresca2021comprehensive, kim2022fast, conti2023reduced, otto2023learning}, or through analytically-defined quadratic manifolds~\citep{barnett2022quadratic, geelen2023operator}. A second direction retains a linear description but exploits problem-specific transformations or coordinate alignments to recover fast Kolmogorov decay, including freezing methods~\citep{ohlberger2013nonlinear}, shifted POD and symmetry reduction~\citep{rowley2003reduction, reiss2018shifted}, transported subspaces~\citep{nair2019transported, rim2023manifold}, and implicit feature tracking~\citep{mirhoseini2023model}. A third line of work, which the present paper pursues, is that of \emph{adaptive} model reduction, wherein the reduced subspace (and the corresponding reduced operators) is not fixed after offline training but is updated \emph{online} as the simulation proceeds, so that the model continuously re-aligns itself with the evolving dynamics~\citep{peherstorfer2015online, peherstorfer2020transport, huang2023predictive, mohaghegh2026feature, hedayat2026toward}.

Early adaptive approaches exploited interpolation across local reduced bases tailored to different regions of parameter space~\citep{amsallem2008interpolation, amsallem2011online, amsallem2012nonlinear}, and were later extended to hyper-reduction through localized DEIM~\citep{peherstorfer2014localized}. Adaptive frameworks that update the basis and hyper-reduction sampling points in a truly streaming fashion were then developed by Peherstorfer and collaborators via Dynamic Data-Driven ROMs~\citep{peherstorfer2015dynamic}, Adaptive DEIM~\citep{peherstorfer2015online}, and its extension to transport-dominated problems, AADEIM~\citep{peherstorfer2020transport}. Related developments include lookahead data-gathering strategies for online adaptation~\citep{singh2023lookahead}, rank-one adaptive sampling for chaotic reacting flows~\citep{huang2023predictive}, feature-guided sampling tailored to shock- and flame-dominated problems~\citep{mohaghegh2026feature}, time-dependent bases derived from variational principles~\citep{ramezanian2021fly, patil2023reduced, jung2025accelerating}, hybrid FOM--ROM snapshot strategies~\citep{zucatti2024adaptive, bai2022reduced, feng2021fom}, basis refinement and sieving methods~\citep{carlberg2015adaptive, etter2020online}, and on-the-fly adaptation for topology optimization~\citep{yano2021globally}. More recently, adaptive formulations have also been explored in the non-intrusive setting using online manifold optimization~\citep{hedayat2026toward}. Collectively, these works establish adaptive ROMs as a promising pathway toward truly predictive model reduction for systems whose dynamics constantly evolve.

Because projection-based ROMs operate within a low-dimensional subspace, the task of adapting them online is, at its core, a problem of \emph{online subspace tracking}---a topic with a long and rich history in the signal processing and streaming data literature~\citep{comon1990tracking, balzano2018streaming}. The central goal of subspace tracking is precisely what an adaptive ROM requires, namely to maintain an estimate of the dominant low-dimensional subspace from a sequence of streaming data vectors while balancing responsiveness to new observations against retention of information learned from the past. Two complementary viewpoints have emerged~\citep{balzano2018streaming}. Algebraic methods treat the problem as the incremental update of a singular or eigenvalue decomposition of a slowly-evolving (possibly exponentially-weighted) data matrix, and include Bunch and Nielsen's incremental SVD~\citep{bunch1978updating}, Brand's thin iSVD variants~\citep{brand2002incremental, brand2006fast}, Krasulina's method~\citep{krasulina1969method}, and noisy and block power methods~\citep{hardt2014noisy, allenzhu2017first}. Geometric methods, in contrast, formulate the update as a stochastic gradient step on a matrix manifold, with Oja's classical stochastic rule~\citep{oja1982simplified}, Grassmannian Rank-One Update Subspace Estimation (GROUSE)~\citep{balzano2010online, balzano2013grouse}, Projection Approximation Subspace Tracking (PAST)~\citep{yang1995projection}, and PETRELS~\citep{chi2013petrels} as notable examples. While subspace-tracking machinery has begun to be leveraged for adaptive model reduction~\citep{zimmermann2018geometric}, its full potential---in particular that of history-aware algebraic updates such as iSVD with a principled forgetting mechanism---has not yet been realized in the adaptive ROM context. Moreover, unlike the work in~\citep{zimmermann2018geometric}, the adaptation signal received in this paper relies on interactions with the main dynamical model.

Despite the significant progress described above, a substantial gap remains between current adaptive ROMs and their deployment in realistic, nonlinear multi-physics applications. Many existing adaptive schemes rely on rank-one updates that respond only to the most recent observation; we will call these methods \emph{instantaneous}, though technically the evolving basis itself is capturing some aspects of the recent system state. Alternatively, other schemes rely on windowed SVD re-computations that are accurate within a finite lookback window but computationally demanding and blind to the dynamics outside of that window. Between these two extremes lies the opportunity to use a subspace update mechanism that is simultaneously lightweight at every step and carries a structured, compressed memory of the system's past. Motivated by this opportunity, we propose in this work a new adaptive reduced-order modeling framework built around incremental singular value decomposition. The proposed framework has four distinctive features. First, it is \emph{history-aware} in the sense that the iSVD update carries forward an encoded representation of the entire observed trajectory within the spectral structure (singular values and left singular vectors) of the evolving basis, while a forgetting factor controls the balance between recent and past information. Second, the model acquires a \emph{lookahead} correction signal by advancing an independent coarse-time-step full-order solve one adaptation window into the future; thus, at every adaptation event the ROM receives approximate, and potentially noisy, information about how the system will behave in the near future. This property is not unique to the proposed framework in this study and have been utilized in recent adaptive ROM literature~\citep{singh2023lookahead, mohaghegh2026feature}. Third, the adaptation leverages this lookahead signal to update the basis and, through the basis, the reduced operators and hyper-reduction sampling machinery that are fully parameterized by it. Fourth, as a consequence of these mechanisms, the resulting model is \emph{predictive}, in that it is able to extrapolate far beyond its offline training regime, and in that sense it behaves as a true predictor rather than as an interpolator. In practice, the proposed framework amounts to a compact, continuously evolving surrogate that corrects itself on the fly through occasional interactions with the full-order solver, with negligible offline training cost, effectively acting as a physics-consistent solver accelerator.

We develop and evaluate the proposed framework on three nonlinear problems of increasing complexity. The one-dimensional viscous Burgers equation serves as a development platform where we study the predictive limitations of static ROMs, compare iSVD against several alternative basis adaptation rules---including a windowed SVD, the Direct method~\citep{mohaghegh2026feature}, the one-step approach~\citep{huang2023predictive}, Oja's rule, and GROUSE---and assess the sensitivity of the framework to its main hyperparameters. The Sod shock tube extends the evaluation to a compressible hyperbolic setting with moving discontinuities, and a one-dimensional ten-species rotating detonation engine (RDE) with detailed hydrogen--oxygen kinetics provides the most demanding, application-driven test. Across all three cases, history-aware updates substantially outperform instantaneous ones, and iSVD delivers the strongest overall performance. In particular, on the RDE problem, the proposed iSVD adaptive ROM improves upon the current state-of-the-art Direct adaptive ROM baseline~\citep{mohaghegh2026feature, Mohaghegh2026selfadaptive} in both predictive accuracy and acceleration factor, identifying iSVD as a particularly effective mechanism for online learning of reduced subspaces in predictive simulation beyond the training regime. A reference Python implementation of the proposed framework, including the experiments, is publicly available at \url{https://github.com/APHedayat/iSVD-ROM}.

The remainder of this paper is organized as follows. Section~\ref{sec:fom} describes the FOM and numerical time-discretization considered throughout. Section~\ref{sec:static_rom} briefly reviews the static projection-based reduced-order modeling framework, including projection and hyper-reduction. Section~\ref{sec:adaptive_rom} introduces the proposed history-aware adaptive framework, detailing the coarse-time-step correction signal, the iSVD-based online basis update, the alternative adaptation rules used for comparison, the full adaptive algorithm, and a computational complexity analysis. Section~\ref{sec:results} presents numerical results on the Burgers equation, the Sod shock tube, and the RDE. Finally, Section~\ref{sec:conclusion} summarizes the main findings and outlines directions for future work.

\section{Full-order model}
\label{sec:fom}

We consider the full-order model (FOM) as a generic nonlinear dynamical system of the form
\begin{equation}
\frac{d\bm q}{dt} = \bm f(\bm q,t),
\qquad
\bm q(0)=\bm q_0,
\label{eq:fom_continuous}
\end{equation}
where $t \in [0,T]$ denotes time, $\bm q : [0,T] \to \mathbb{R}^N$ is the high-dimensional state, $\bm q_0 \in \mathbb{R}^N$ is the initial condition, and $\bm f : \mathbb{R}^N \times [0,T] \to \mathbb{R}^N$ is a generally nonlinear operator describing the system dynamics. In the context of time-dependent partial differential equations, \eqref{eq:fom_continuous} represents the semi-discrete system obtained after spatial discretization. For conservation-law systems, $\bm q$ denotes the vector of conservative variables and $N$ is the total number of degrees of freedom. Typically,
$
N = N_{\mathrm{var}}\,N_{\mathrm{elem}},
$
where $N_{\mathrm{elem}}$ is the number of spatial degrees of freedom and $N_{\mathrm{var}}$ is the number of state variables stored at each spatial location.

To advance \eqref{eq:fom_continuous} in time, let
$
0=t^0<t^1<\cdots<t^{N_t}=T
$
with a uniform time step
$
\Delta t = t^i-t^{i-1}
$
(for $i = 1,2,\dots,N_t$), and denote by $\bm q^n = \bm q(t^n)$ the fully discrete state at time level $t^n$. A general $l$-step linear multistep discretization of \eqref{eq:fom_continuous} can be written in residual form as
\begin{equation}
\bm r^n(\bm q^n)
\coloneq
\bm q^n
+\sum_{j=1}^{l}\alpha_j \bm q^{n-j}
-\Delta t\,\beta_0\,\bm f(\bm q^n,t^n)
-\Delta t\sum_{j=1}^{l}\beta_j \bm f(\bm q^{n-j},t^{n-j}),
\qquad n\ge l,
\label{eq:fom_lmm}
\end{equation}
where $\alpha_j,\beta_j \in \mathbb{R}$ are the coefficients of the chosen time integration scheme. If $\beta_0=0$, we call the method explicit; otherwise, it is implicit. The state $\bm q^n$ is obtained at each time step by solving
\begin{equation}
\bm r^n(\bm q^n)=\bm 0.
\label{eq:fom_discrete_solve}
\end{equation}
When an implicit scheme is used, we denote the Jacobian of the semi-discrete operator by
$
\bm J_f(\bm q,t)
\coloneq
\frac{\partial \bm f}{\partial \bm q}(\bm q,t),
$
and the Jacobian of the fully discrete residual by
$
\bm J_r^n(\bm q^n)
\coloneq
\frac{\partial \bm r^n}{\partial \bm q^n}(\bm q^n).
$
These quantities will appear naturally in the reduced formulations introduced in the following sections.

\section{Static reduced-order modeling}
\label{sec:static_rom}

\subsection{Projection}
\label{subsec:static_projection_roms}

In a projection-based, static, reduced-order model (ROM), the high-dimensional state is approximated using a fixed low-dimensional trial subspace constructed offline. To this end, one first collects a set of full-order snapshots over a training time interval,
$
\bm Q_{\mathrm{train}}
\coloneq
\begin{bmatrix}
\bm q(t^1) & \bm q(t^2) & \cdots & \bm q(t^{N_s})
\end{bmatrix}
\in \mathbb{R}^{N\times N_s},
$
where \(N_s\) denotes the number of training snapshots. To preprocess these data, we introduce a reference state \(\bm q_{\mathrm{ref}} \in \mathbb{R}^N\) and a diagonal scaling matrix \(\bm D \in \mathbb{R}^{N\times N}\). The reference state is typically chosen as the initial solution,
while the scaling matrix \(\bm D\) is introduced so that different physical variables entering the snapshot matrix have comparable magnitudes before the basis is computed.
We define \(\bm D\) as a block-diagonal matrix
$
\bm D = 
\operatorname{diag}
\left(
\bm D_1,\bm D_2,\dots,\bm D_{N_{\mathrm{elem}}}
\right),
$
where each spatial block is given by
\begin{equation}
\bm D_i
\coloneq
\operatorname{diag}
\left(
\phi_{1,\mathrm{norm}}^{-1},
\phi_{2,\mathrm{norm}}^{-1},
\dots,
\phi_{N_{\mathrm{var}},\mathrm{norm}}^{-1}
\right),
\qquad
i=1,\dots,N_{\mathrm{elem}}.
\label{eq:scaling_matrix_local_block}
\end{equation}
Here, \(\phi_{v,\mathrm{norm}}\) is the characteristic magnitude of the \(v\)th state variable over the training interval, defined by
\begin{equation}
\phi_{v,\mathrm{norm}}
\coloneq
\frac{1}{\Delta T}
\int_{t_0}^{t_0+\Delta T}
\frac{1}{|\Omega|}
\int_{\Omega}
\left(\phi_v'(\bm x,t)\right)^2
\,d\bm x\,dt,
\qquad
v=1,\dots,N_{\mathrm{var}},
\label{eq:variable_normalization}
\end{equation}
where \(\phi_v'\) denotes the \(v\)th component of the centered training snapshot field.
The reduced trial basis is then typically obtained by computing the proper orthogonal decomposition (POD) of the preprocessed snapshot matrix
$
\bm D\!\left(\bm Q_{\mathrm{train}} - \bm q_{\mathrm{ref}}\bm 1^T\right),
$
and retaining the leading \(r\) left singular vectors associated with the dominant singular values. Let
$
\bm \Phi \in \mathbb{R}^{N \times r},
$
with
$
r \ll N
$
and
$
\bm \Phi^T \bm \Phi = \bm I_r,
$
denote the resulting reduced basis, whose columns span the trial subspace. The full-order state is then approximated as
\begin{equation}
\bm q(t) \approx \tilde{\bm q}(t) \coloneq \bm q_{\mathrm{ref}} + \bm D^{-1}\bm \Phi \bm a(t),
\label{eq:trial_approximation}
\end{equation}
where \(\bm a(t) \in \mathbb{R}^r\) denotes the reduced coordinates.
Substituting the reduced ansatz \eqref{eq:trial_approximation} into the semi-discrete FOM \eqref{eq:fom_continuous} yields
\begin{equation}
\bm \Phi \dot{\bm a}(t)
=
\bm D\,\bm f\!\left(\bm q_{\mathrm{ref}} + \bm D^{-1}\bm \Phi \bm a(t), t\right).
\label{eq:scaled_inserted_trial_ansatz}
\end{equation}
The system in \eqref{eq:scaled_inserted_trial_ansatz} is defined in \(\mathbb{R}^N\), so a test space is introduced to define the reduced dynamics that live in \(\mathbb{R}^r\). Let
$
\bm \Psi \in \mathbb{R}^{N \times r}
\label{eq:test_basis}
$
be this full-column-rank test basis. The projection condition requires the residual associated with \eqref{eq:scaled_inserted_trial_ansatz} to be orthogonal to the test space:
\begin{equation}
\bm \Psi^T
\left[
\bm \Phi \dot{\bm a}(t)
-
\bm D\,\bm f\!\left(\bm q_{\mathrm{ref}}+\bm D^{-1}\bm \Phi \bm a(t), t\right)
\right]
=
\bm 0.
\label{eq:pg_continuous_condition}
\end{equation}
The character of the resulting reduced model depends on the relationship between the trial basis \(\bm \Phi\) and the test basis \(\bm \Psi\). If the trial and test spaces coincide, the projection is orthogonal and one obtains a Galerkin ROM. If they differ, the projection is oblique and one obtains a Petrov--Galerkin ROM.

\subsubsection{Galerkin formulation}
\label{subsubsec:static_galerkin}

The Galerkin ROM is obtained by choosing the test basis equal to the trial basis,
$
\bm \Psi = \bm \Phi.
$
In this case, the projection condition \eqref{eq:pg_continuous_condition} becomes
\begin{equation}
\bm \Phi^T
\left[
\bm \Phi \dot{\bm a}(t)
-
\bm D\,\bm f\!\left(\bm q_{\mathrm{ref}}+\bm D^{-1}\bm \Phi \bm a(t), t\right)
\right]
=
\bm 0,
\label{eq:galerkin_continuous_condition}
\end{equation}
which, using \(\bm \Phi^T \bm \Phi = \bm I_r\), yields
\begin{equation}
\dot{\bm a}(t)
=
\bm \Phi^T \bm D\,\bm f\!\left(\bm q_{\mathrm{ref}}+\bm D^{-1}\bm \Phi \bm a(t), t\right).
\label{eq:galerkin_continuous_rom}
\end{equation}
Thus, the Galerkin ROM is obtained by orthogonally projecting the governing equations onto the trial subspace.

\subsubsection{Petrov--Galerkin formulation}
\label{subsubsec:static_petrov-galerkin}

More generally, one may choose a test basis \(\bm \Psi \neq \bm \Phi\), leading to an oblique projection of the governing equations. In this case, the reduced dynamics are defined by
\begin{equation}
\bm \Psi^T
\left[
\bm \Phi \dot{\bm a}(t)
-
\bm D\,\bm f\!\left(\bm q_{\mathrm{ref}}+\bm D^{-1}\bm \Phi \bm a(t), t\right)
\right]
=
\bm 0,
\label{eq:petrov_galerkin_continuous_condition}
\end{equation}
or, equivalently, assuming \(\bm \Psi^T \bm \Phi\) is nonsingular,
\begin{equation}
\dot{\bm a}(t)
=
\left[ \bm \Psi^T \bm \Phi \right]^{-1}
\bm \Psi^T
\bm D\,\bm f\!\left(\bm q_{\mathrm{ref}}+\bm D^{-1}\bm \Phi \bm a(t), t\right).
\label{eq:petrov_galerkin_continuous_rom}
\end{equation}

A particularly important Petrov--Galerkin method is the least-squares Petrov--Galerkin (LSPG) formulation~\citep{Carlberg2011, carlberg2017galerkin}, which is defined directly at the fully discrete level. Rather than projecting the semi-discrete equations first and then discretizing in time, LSPG seeks the reduced state at each time step by minimizing the norm of the scaled full-order discrete residual over the trial manifold:
\begin{equation}
\bm a^n
\coloneq
\arg\min_{\hat{\bm a} \in \mathbb{R}^r}
\left\|
\bm D\,\bm r^n\!\left(\bm q_{\mathrm{ref}}+\bm D^{-1}\bm \Phi \hat{\bm a}\right)
\right\|_2^2.
\label{eq:lspg_minimization}
\end{equation}
The first-order optimality condition associated with \eqref{eq:lspg_minimization} is
\begin{equation}
\left(\bm \Psi^n_{\mathrm{LSPG}}\right)^T
\bm D\,\bm r^n\!\left(\tilde{\bm q}^{\,n}\right)
=
\bm 0,
\label{eq:lspg_stationarity}
\end{equation}
with $\tilde{\bm q}$ defined in~\eqref{eq:trial_approximation} and test basis
\begin{equation}
\bm \Psi^n_{\mathrm{LSPG}}
\coloneq
\bm D\,\bm J_r^n(\tilde{\bm q}^{\,n})\,\bm D^{-1}\bm \Phi,
\label{eq:lspg_test_basis}
\end{equation}
where \(\bm J_r^n\) is the Jacobian of the fully discrete residual defined in \eqref{eq:fom_lmm}. Thus, LSPG is a state-dependent Petrov--Galerkin projection in which the test space is induced by the minimization of the scaled discrete residual.

For example, under backward Euler time integration,
\begin{equation}
\bm r^n(\bm q)
=
\bm q - \bm q^{n-1} - \Delta t\,\bm f(\bm q,t^n),
\label{eq:be_residual_rom_section}
\end{equation}
and therefore
\begin{equation}
\bm J_r^n(\bm q)
=
\bm I - \Delta t\,\bm J_f(\bm q,t^n),
\label{eq:be_residual_jacobian_rom_section}
\end{equation}
so that the LSPG test basis for this particular time integration scheme becomes
\begin{equation}
\bm \Psi^n_{\mathrm{LSPG}}
=
\left(
\bm I - \Delta t\,\bm D\,\bm J_f(\tilde{\bm q}^{\,n},t^n)\bm D^{-1}
\right)\bm \Phi.
\label{eq:lspg_test_basis_be}
\end{equation}
This state-dependent test basis is one of the key distinctions between LSPG and Galerkin projection. LSPG is particularly attractive for nonlinear implicit time integration, where its discrete residual minimization principle often provides improved robustness and accuracy relative to continuous projection methods~\citep{Carlberg2013gnat, carlberg2017galerkin, Parish2020}.

\subsection{Hyper-reduction}
\label{subsec:static_hyperreduction}

Although projection reduces the state dimension from \(N\) to \(r\), the evaluation of the nonlinear terms in the reduced equations generally still scales with the ambient dimension \(N\). Consequently, projection alone does not guarantee an efficient online model. Hyper-reduction addresses this issue by restricting the evaluation of the nonlinear function to a small set of sampled entries and reconstructing the required full-dimensional quantities from those samples. In this work, two hyper-reduction strategies are considered: the discrete empirical interpolation method and feature-guided sampling.

\subsubsection{Discrete empirical interpolation method}
\label{subsubsec:static_deim}

The discrete empirical interpolation method (DEIM)~\citep{Chaturantabut2010} approximates the scaled nonlinear term using an additional low-dimensional basis and a small number of sampled entries. Let $\bm U \in \mathbb{R}^{N\times m}$ with $m \ll N$ and $\bm U^T \bm U = \bm I_m$ denote the DEIM basis, and let
$\bm P
\coloneq
\begin{bmatrix}
\bm e_{p_1} & \bm e_{p_2} & \cdots & \bm e_{p_m}
\end{bmatrix}
\in \mathbb{R}^{N\times m}$
be the sampling matrix associated with the DEIM interpolation indices \(\{p_1,\dots,p_m\}\). The DEIM approximation is then
\begin{equation}
\bm D\,\bm f(\bm q,t)
\approx
\bm U (\bm P^T \bm U)^{\dagger} \bm P^T \bm D\,\bm f(\bm q,t).
\label{eq:deim_approximation}
\end{equation}
In the present work, we take the DEIM basis equal to the trial basis,
$
\bm U = \bm \Phi,
$
which has been found to provide sufficiently accurate hyper-reduction~\citep{Huang2022}. The scaled nonlinear term is therefore approximated as
\begin{equation}
\bm D\,\bm f(\bm q,t)
\approx
\bm \Phi (\bm P^T \bm \Phi)^{\dagger} \bm P^T \bm D\,\bm f(\bm q,t).
\label{eq:qdeim_approximation_phi}
\end{equation}
Substituting this approximation into the Galerkin ROM \eqref{eq:galerkin_continuous_rom} yields
\begin{equation}
\dot{\bm a}(t)
\approx
(\bm P^T \bm \Phi)^{\dagger} \bm P^T
\bm D\,\bm f\!\left(\bm q_{\mathrm{ref}}+\bm D^{-1}\bm \Phi \bm a(t),t\right),
\label{eq:deim_galerkin_rom}
\end{equation}
where the orthonormality of \(\bm \Phi\) has been used. Here, only the \(m\) sampled entries
$
\bm P^T \bm D\,\bm f\!\left(\bm q_{\mathrm{ref}}+\bm D^{-1}\bm \Phi \bm a(t),t\right)
$
are evaluated online, while the operator \((\bm P^T \bm \Phi)^{\dagger}\) is precomputed offline.

For LSPG, the hyper-reduced minimization problem may be written as
\begin{equation}
\bm a^n
\coloneq
\arg\min_{\hat{\bm a}\in\mathbb{R}^r}
\left\|
\bm \Phi(\bm P^T \bm \Phi)^\dagger
\bm P^T \bm D\,
\bm r^n\!\left(\bm q_{\mathrm{ref}}+\bm D^{-1}\bm \Phi \hat{\bm a}\right)
\right\|_2^2.
\label{eq:hyperreduced_lspg}
\end{equation}
The resulting hyper-reduced LSPG formulation is
\begin{equation}
\left(\bm P^T \bm \Psi^n_{\mathrm{LSPG}}\right)^T
\left[(\bm P^T \bm \Phi)^\dagger\right]^T
(\bm P^T \bm \Phi)^\dagger
\bm P^T \bm D\,\bm r^n(\tilde{\bm q}^{\,n})
=
\bm 0,
\label{eq:hyperreduced_lspg_stationarity}
\end{equation}
where $\bm \Psi^n_{\mathrm{LSPG}}$ is defined in \eqref{eq:lspg_test_basis}. As a result, DEIM restricts attention to the sampled test and trial bases.

In practice, the interpolation indices can be selected using QDEIM~\citep{Zlatko2016}, in which the sampling points are obtained from a column-pivoted QR factorization of the basis transpose,
\begin{equation}
\bm \Phi^T \bm \Pi = \bm Q \bm R,
\label{eq:qdeim_qr}
\end{equation}
where \(\bm \Pi\) is a permutation matrix. The first \(m\) pivots define the sampling indices \(\{p_1,\dots,p_m\}\).

\subsubsection{Feature-guided sampling}
\label{subsubsec:static_fgs}

We also consider the feature-guided sampling (FGS) strategy proposed by Mohaghegh and Huang~\citep{mohaghegh2026feature}. In this case, the hyper-reduced equations (\eqref{eq:deim_galerkin_rom} for Galerkin, \eqref{eq:hyperreduced_lspg_stationarity} for LSPG) remain exactly the same as in the DEIM setting, and only the set of sampling points is modified. More specifically, FGS augments the sampling set obtained by an algebraic method such as QDEIM with additional physics-informed points chosen to better capture important flow features.

Let
$
\Theta : \mathbb{R}^N \rightarrow \mathbb{R}^{N_\theta}
$
denote a feature map that extracts a physically meaningful quantity from the state, and define
$
\bm \theta_n \coloneq \Theta({\bm y}^{n+z}),
$
where \({\bm y}^{n+z}\) is an estimated future state, which will be defined in the next section (see equation \eqref{eq:coarse_fom_backward_euler}). The additional FGS points are then selected in regions where the feature gradient is strongest, namely through
\begin{equation}
\bm S_n
\coloneq
\arg\max_{\bm S_n \in \mathcal{S}_{n_s}}
\|\nabla \bm \theta_n\|_2^2.
\label{eq:fgs_selection}
\end{equation}
Thus, FGS may be viewed as DEIM/QDEIM with additional physics-informed oversampling. This strategy is especially useful in problems with localized structures, such as shocks, reaction fronts, or detonation waves, where purely algebraic sampling may fail to adequately resolve the most dynamically important regions.

The static formulations presented above serve as the building blocks for the adaptive framework considered in this work, with both LSPG--QDEIM and Galerkin--FGS arising as particular instances in the results section.

\section{History-aware adaptive reduced-order modeling}
\label{sec:adaptive_rom}

The static ROMs described in Section~\ref{sec:static_rom} rely on a fixed trial basis learned offline. While such models can be highly accurate when the online dynamics remain close to the training regime, their performance may deteriorate when the solution manifold evolves beyond the subspace represented by the initial basis. To address this limitation, we now develop an adaptive reduced-order modeling framework in which the model is updated online and evolves with the system. This framework is built around two design questions:
\begin{enumerate}
    \item \emph{where the online correction information comes from}, and
    \item \emph{how the reduced basis is updated once that information is received}.
\end{enumerate}
In the following, we will answer both of these questions.

\subsection{Correction signal}
\label{subsec:adaptive_correction_signal}

We obtain the correction signal from a coarse-time-step FOM, which we refer to as the coarse FOM throughout the manuscript. More precisely, recall the fully discrete full-order residual introduced in \eqref{eq:fom_lmm}. At adaptation events, instead of advancing the FOM on the fine time step \(\Delta t\), we advance it with the larger step \(z\Delta t\), where \(z \in \mathbb{N}\) is the adaptation frequency. Thus, the correction signal is generated by the same full-order equations, but on a coarser temporal grid. For a general linear multistep discretization, this simply amounts to replacing \(\Delta t\) by \(z\Delta t\) in \eqref{eq:fom_lmm}. We employ the backward-Euler scheme to obtain the adaptation signal in this work, so the coarse interaction takes the form
\begin{equation}
\bm y^{n+z} - \bm y^{n} - z\Delta t\,\bm f(\bm y^{n+z},t^{n+z}) = \bm 0,
\label{eq:coarse_fom_backward_euler}
\end{equation}
where \(\bm y^{n}\) denotes the most recently available coarse full-order state and \(\bm y^{n+z}\) is the new correction snapshot.

\begin{figure}
\centering
\includegraphics[width=0.8\linewidth]{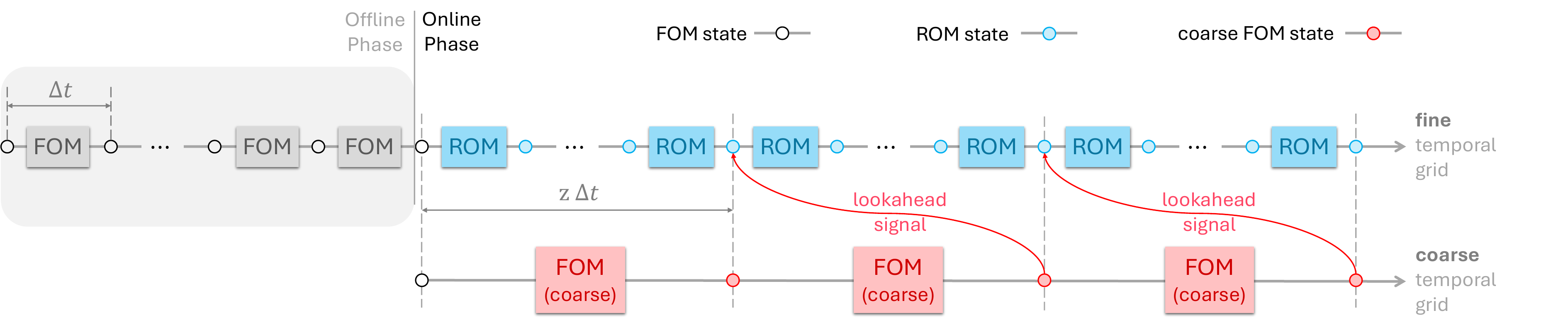}
\caption{Schematic of the lookahead correction signal mechanism used in the proposed adaptive ROM framework. The ROM advances on the fine time grid, while a separate coarse-time-step full-order trajectory is evolved independently using step size \(z\Delta t\). At each adaptation event, the coarse FOM provides a lookahead snapshot approximating the system state \(z\) fine steps into the future, which is then used as the correction signal for basis adaptation. The newly generated coarse snapshot is subsequently stored and advanced again to provide the next lookahead signal.}
\label{fig:lookahead}
\end{figure}

Equation \eqref{eq:coarse_fom_backward_euler} shows that the signal provided to the adaptive ROM is inherently a \emph{lookahead} signal, meaning that every time the model is adapted, the available information corresponds to an approximation of the system state \(z\) time steps into the future. The use of lookahead information for online ROM adaptation has been emphasized in prior adaptive model reduction work, including \citep{singh2023lookahead} and \citep{mohaghegh2026feature}. In the present work, we adopt the same lookahead principle to generate correction snapshots. A second important feature of the present construction is that the correction signal trajectory is generated independently of the ROM trajectory. At the beginning of the online phase, the coarse FOM is initialized with the last full-order snapshot available from the offline stage. Thereafter, each new correction snapshot \(\bm y^{n+z}\) is obtained by advancing the coarse FOM from the previously available coarse state \(\bm y^{n}\). This prevents ROM prediction errors from feeding back into the correction signal trajectory. The resulting signal is still approximate and may be noisy, but its errors arise from the coarse full-order evolution rather than from accumulated ROM prediction error. Prior adaptive ROM studies~\citep{huang2023predictive, mohaghegh2026feature} have demonstrated that such approximate information obtained from the coarse FOM can be highly useful for online adaptation. A schematic of this lookahead correction mechanism is shown in Figure~\ref{fig:lookahead}.

\subsection{Online basis adaptation}
\label{subsec:adaptive_basis_updates}

Once the correction signal has been obtained, the next step is to use it to adapt the ROM. In the intrusive formulations considered here, the model is fully parameterized by its reduced basis, and therefore adaptation reduces to updating the underlying trial subspace. The problem thus naturally becomes one of online subspace tracking: given a newly acquired snapshot, update the current low-dimensional subspace so that it continues to represent the evolving dominant dynamics of the system. This viewpoint connects adaptive reduced-order modeling to a broad literature on streaming and online subspace estimation. Among the online subspace tracking methods, the primary choice in this work is Brand's incremental singular value decomposition (iSVD)~\citep{brand2002incremental, brand2006fast}. The appeal of iSVD in the present setting is that it provides an online mechanism for updating the trial subspace using newly acquired correction snapshots, while simultaneously retaining a compressed spectral summary of previously observed dynamics. This means that iSVD carries forward historical information through its singular structure, and hence is \emph{history-aware} in that sense. This feature is central to the adaptive framework developed here.

Since the trial basis is constructed in the scaled coordinates introduced in Section~\ref{sec:static_rom}, the correction snapshots must be processed in the same manner before they are used for basis adaptation. Accordingly, for each newly received correction snapshot \(\bm y_{k+1}\), we define the preprocessed snapshot
$
\widehat{\bm y}_{k+1}
\coloneq
\bm D\left(\bm y_{k+1}-\bm q_{\mathrm{ref}}\right).
$
To formalize the update, let
$
\widehat{\bm Y}_k
\coloneq
\begin{bmatrix}
\widehat{\bm y}_1 & \widehat{\bm y}_2 & \cdots & \widehat{\bm y}_k
\end{bmatrix}
\in \mathbb{R}^{N\times k}
$
denote the matrix of preprocessed correction snapshots accumulated up to adaptation step \(k\). Note that we distinguish between the fine-time-step index $n$ and the adaptation event index $k$. In order to bias the representation toward recent information while retaining a memory of the past, we introduce the exponentially weighted snapshot matrix
\begin{equation}
\widehat{\bm Y}_k^{(\lambda)}
\coloneq
\begin{bmatrix}
\lambda^{k-1}\widehat{\bm y}_1 &
\lambda^{k-2}\widehat{\bm y}_2 &
\cdots &
\lambda \widehat{\bm y}_{k-1} &
\widehat{\bm y}_k
\end{bmatrix},
\qquad
\lambda \in [0,1],
\label{eq:weighted_snapshot_matrix}
\end{equation}
where \(\lambda\) is the forgetting factor. When \(\lambda=1\), all past snapshots are treated equally and no forgetting occurs; when \(\lambda<1\), older snapshots are progressively discounted in favor of newer observations.

At adaptation step \(k\), the goal is to maintain a rank-\(r\) approximation of the dominant left singular subspace of \(\widehat{\bm Y}_k^{(\lambda)}\) without recomputing a full singular value decomposition from scratch. Let
\begin{equation}
\widehat{\bm Y}_k^{(\lambda)}
\approx
\bm \Phi_k \bm \Sigma_k \bm V_k^T,
\qquad
\bm \Phi_k \in \mathbb{R}^{N\times r},
\qquad
\bm \Phi_k^T \bm \Phi_k = \bm I_r,
\qquad
\bm \Sigma_k = \mathrm{diag}(\bm \sigma_k),
\label{eq:isvd_current_factorization}
\end{equation}
denote the current reduced representation, where \(\bm \sigma_k \in \mathbb{R}^r\) collects the singular values retained by the update. Upon receipt of a new correction snapshot \(\widehat{\bm y}_{k+1}\), the weighted history matrix is augmented according to
\begin{equation}
\widehat{\bm Y}_{k+1}^{(\lambda)}
=
\begin{bmatrix}
\lambda \widehat{\bm Y}_k^{(\lambda)} & \widehat{\bm y}_{k+1}
\end{bmatrix}.
\label{eq:isvd_augmented_history}
\end{equation}
Thus, each basis update can be interpreted as incorporating one new column into an exponentially weighted snapshot history. To update the left singular subspace efficiently, the new snapshot is first decomposed into its component inside the current trial subspace and its orthogonal residual
\begin{equation}
\bm p_{k+1}
\coloneq
\arg\min_{\bm p \in \mathbb{R}^r}
\left\|
\widehat{\bm y}_{k+1} - \bm \Phi_k \bm p
\right\|_2^2
=
\bm \Phi_k^{\dagger}\widehat{\bm y}_{k+1},
\qquad
\bm q_{k+1}
\coloneq
\widehat{\bm y}_{k+1} - \bm \Phi_k \bm p_{k+1}.
\label{eq:isvd_residual}
\end{equation}
Here, \(\bm p_{k+1}\) measures the representation of the incoming snapshot in the current reduced basis, and \(\bm q_{k+1}\) captures the component not represented by the current subspace. Using \eqref{eq:isvd_current_factorization}--\eqref{eq:isvd_residual}, the augmented weighted snapshot matrix \eqref{eq:isvd_augmented_history} admits the factorization
\begin{equation}
\widehat{\bm Y}_{k+1}^{(\lambda)}
\approx
\begin{bmatrix}
\bm \Phi_k & \displaystyle \frac{\bm q_{k+1}}{\|\bm q_{k+1}\|_2}
\end{bmatrix}
\bm K_{k+1}
\begin{bmatrix}
\bm V_k^T & \bm 0 \\
\bm 0   & 1
\end{bmatrix},
\label{eq:isvd_factorization}
\end{equation}
where the small core matrix is
\begin{equation}
\bm K_{k+1}
\coloneq
\begin{bmatrix}
\lambda \bm \Sigma_k & \bm p_{k+1} \\
\bm 0^T & \|\bm q_{k+1}\|_2
\end{bmatrix}
\in \mathbb{R}^{(r+1)\times(r+1)}.
\label{eq:isvd_core_matrix}
\end{equation}
Since the outer factors in \eqref{eq:isvd_factorization} have orthonormal columns, the updated singular values and left singular vectors of \(\widehat{\bm Y}_{k+1}^{(\lambda)}\) are obtained by computing the singular value decomposition of the small and nearly diagonal matrix \(\bm K_{k+1}\):
\begin{equation}
\bm K_{k+1}
=
\overline{\bm U}_{k+1}
\overline{\bm \Sigma}_{k+1}
\overline{\bm V}_{k+1}^T.
\label{eq:isvd_core_svd}
\end{equation}
The updated left singular basis is therefore
\begin{equation}
\widetilde{\bm \Phi}_{k+1}
\coloneq
\begin{bmatrix}
\bm \Phi_k & \displaystyle \frac{\bm q_{k+1}}{\|\bm q_{k+1}\|_2}
\end{bmatrix}
\overline{\bm U}_{k+1},
\label{eq:isvd_updated_basis_full}
\end{equation}
with associated singular values given by the diagonal entries of \(\overline{\bm \Sigma}_{k+1}\). To maintain a reduced space of dimension \(r\), the update is truncated by retaining only the leading \(r\) singular directions:
\begin{equation}
\bm \Phi_{k+1}
\coloneq
\widetilde{\bm \Phi}_{k+1}(:,1\!:\!r),
\qquad
\bm \Sigma_{k+1}
\coloneq
\overline{\bm \Sigma}_{k+1}(1\!:\!r,1\!:\!r),
\qquad
\bm \sigma_{k+1}
\coloneq
\mathrm{diag}(\bm \Sigma_{k+1}).
\label{eq:isvd_truncation}
\end{equation}
By design, iSVD preserves orthonormality of the basis, so no additional orthonormalization is required after \(\bm \Phi_{k+1}\) is computed.

The outlined procedure above reveals why iSVD is inherently history-aware. The new basis is not determined solely by the incoming snapshot \(\widehat{\bm y}_{k+1}\); rather, it is obtained from the interaction of this snapshot with the compressed spectral representation \((\bm \Phi_k,\bm \Sigma_k)\) of the entire weighted history. This can be seen more explicitly by considering the associated weighted second-moment matrix
$
\bm C_k
\coloneq
\widehat{\bm Y}_k^{(\lambda)} \widehat{\bm Y}_k^{(\lambda)T}.
$
From \eqref{eq:isvd_augmented_history}, one obtains the recursion
$
\bm C_{k+1}
=
\lambda^2 \bm C_k + \widehat{\bm y}_{k+1}\widehat{\bm y}_{k+1}^T.
$
Thus, iSVD tracks the dominant eigenspace of an exponentially weighted covariance operator. Persistent directions in the past history are encoded in the singular values carried by \(\bm \Sigma_k\), while new directions enter through the residual component \(\bm q_{k+1}\). The forgetting factor \(\lambda\) acts as a memory valve in this interaction. This mechanism is particularly well suited to adaptive reduced-order modeling. In evolving nonlinear systems, new correction snapshots may contain local transient information that should influence the basis, but not necessarily dominate it entirely. The iSVD update strikes this balance naturally, since the coordinates \(\bm p_{k+1}\) measure compatibility with the current subspace, the residual norm \(\|\bm q_{k+1}\|_2\) quantifies genuinely new information, and the singular values encode the cumulative importance of previously learned directions. As a result, the updated basis reflects both recent observations and an encoding of the strength of the past data correlations.

To assess the behavior of the proposed iSVD-based update against other online adaptation strategies, we also consider several alternative basis update rules. These methods differ mainly in how they balance newly received information against previously learned subspace structure.

\begin{enumerate}

    \item \textit{Windowed singular value decomposition}:\\
        A natural baseline is to recompute the basis from scratch using a rolling window of the most recent correction snapshots. Let
        \begin{equation}
        \widehat{\bm Y}^{(w)}_{k+1}
        \coloneq
        \begin{bmatrix}
        \widehat{\bm y}_{k-w+2} &
        \widehat{\bm y}_{k-w+3} &
        \cdots &
        \widehat{\bm y}_{k+1}
        \end{bmatrix}
        \in \mathbb{R}^{N\times w},
        \label{eq:wsvd_window_matrix}
        \end{equation}
        where \(w\) is the number of coarse FOM correction snapshots retained in the window. The window is initialized using the last \(w\) full-order snapshots from the offline training dataset, and is updated at adaptation events, where the oldest snapshot is discarded and the newly received coarse FOM signal is appended. The updated basis is then defined as the leading \(r\) left singular vectors of \(\widehat{\bm Y}^{(w)}_{k+1}\),
        \begin{equation}
        \widehat{\bm Y}^{(w)}_{k+1}
        =
        \bm U^{(w)}_{k+1}
        \bm \Sigma^{(w)}_{k+1}
        \bm V^{(w)T}_{k+1},
        \qquad
        \bm \Phi_{k+1}
        \coloneq
        \bm U^{(w)}_{k+1}(:,1\!:\!r).
        \label{eq:wsvd_update}
        \end{equation}
        This strategy retains a finite recent history, but unlike iSVD it does not propagate past information recursively through a compressed spectral state. Instead, the basis is rebuilt from scratch on a moving window at every update.

    \item \textit{Direct method}:\\
        The Direct basis adaptation method~\cite{mohaghegh2026feature} updates the basis from the same correction snapshot window used above for windowed SVD. Here, the basis update is posed as
        \begin{equation}
        \arg\min_{\bm \Phi_{k+1}\in\mathbb{R}^{N\times r}}
        \left\|
        \bm \Phi_{k+1}\bm \Phi_k^\dagger \widehat{\bm Y}^{(w)}_{k+1}
        -
        \widehat{\bm Y}^{(w)}_{k+1}
        \right\|_F^2.
        \label{eq:direct_optimization}
        \end{equation}
        One feasible closed-form update is given by
        \begin{equation}
        \bm \Phi_{k+1}
        \coloneq
        \widehat{\bm Y}^{(w)}_{k+1}
        \left(
        \bm \Phi_k^\dagger \widehat{\bm Y}^{(w)}_{k+1}
        \right)^\dagger .
        \label{eq:direct_update}
        \end{equation}
        Thus, the direct method updates the basis by fitting it directly to a recent data window.

        \item \textit{One-step approach}:\\
        We now turn to basis adaptation methods that act \emph{instantaneously} on the newly available information and do not preserve system history in the manner of the previous methods. Since these updates are designed to react to the state at the current time level, it is more natural to use the previously received coarse FOM signal, which approximates the \emph{current} state of the system, rather than the newly generated lookahead signal for the future time level. We denote the previous preprocessed correction snapshot by \(\widehat{\bm y}_k\).
        The one-step basis adaptation method of Huang and Duraisamy~\cite{huang2023predictive} applies a rank-one correction to the basis to eliminate the projection error of this estimated full state. Let $\bm \Phi_k \bm a_{n+1}^{-}$
        denote the ROM-predicted state in the scaled coordinates before basis adaptation. The update takes the form
        \begin{equation}
        \bm \Phi_{k+1}
        \coloneq
        \operatorname{orth}
        \left(
        \bm \Phi_k
        +
        \frac{
        (\widehat{\bm y}_k-\bm \Phi_k \bm a_{n+1}^{-})
        (\bm a_{n+1}^{-})^T
        }{
        \|\bm a_{n+1}^{-}\|_2^2
        }
        \right),
        \label{eq:onestep_update}
        \end{equation}
        where \(\operatorname{orth}(\cdot)\) denotes orthonormalization. This update is compact and highly responsive to \(\widehat{\bm y}_k\), but it does not maintain a spectral summary of the past history.

    \item \textit{Oja's rule}:\\
        Oja's rule~\citep{oja1982simplified} is a classical streaming update that may be viewed as a stochastic gradient step for principal subspace estimation. Given the current basis \(\bm \Phi_k\) and the current snapshot \(\widehat{\bm y}_k\), the update is
        \begin{equation}
        \bm \Phi_{k+1}
        \coloneq
        \operatorname{orth}
        \left(
        \bm \Phi_k
        +
        \eta\,
        \widehat{\bm y}_k
        \bigl((\widehat{\bm y}_k)^T \bm \Phi_k\bigr)
        \right),
        \label{eq:oja_update}
        \end{equation}
        where \(\eta>0\) is a learning rate. Oja's method is computationally simple and memory-efficient, but just like the one-step approach, it reacts primarily to the current snapshot.

    \item \textit{Grassmannian Rank-One Update Subspace Estimation}:\\
        Grassmannian Rank-One Update Subspace Estimation (GROUSE)~\citep{balzano2010online} is a geometric subspace tracking method that updates the basis along a rank-one geodesic direction on the Grassmann manifold. Let
        \begin{equation}
        \bm w_{k+1}
        \coloneq
        \arg\min_{\bm w\in\mathbb{R}^r}
        \left\|
        \widehat{\bm y}_k - \bm \Phi_k \bm w
        \right\|_2^2,
        \qquad
        \bm p_{k+1} \coloneq \bm \Phi_k \bm w_{k+1},
        \qquad
        \bm r_{k+1} \coloneq \widehat{\bm y}_k - \bm p_{k+1}.
        \label{eq:grouse_projection}
        \end{equation}
        Defining
        $
        \alpha_{k+1} \coloneq \eta \|\bm p_{k+1}\|_2 \|\bm r_{k+1}\|_2,
        $
        the update can be written as
        \begin{equation}
        \bm \Phi_{k+1}
        \coloneq
        \bm \Phi_k
        +
        \left[
        \bigl(\cos\alpha_{k+1}-1\bigr)\frac{\bm p_{k+1}}{\|\bm p_{k+1}\|_2}
        +
        \sin\alpha_{k+1}\frac{\bm r_{k+1}}{\|\bm r_{k+1}\|_2}
        \right]
        \frac{\bm w_{k+1}^T}{\|\bm w_{k+1}\|_2}.
        \label{eq:grouse_update}
        \end{equation}
        GROUSE preserves orthonormality by construction and is especially attractive from a geometric subspace tracking viewpoint, but like one-step approach and Oja's rule it is primarily driven by the instantaneous snapshot rather than an explicitly maintained spectral history. In \cite{balzano2022equivalence}, this method was shown to be equivalent to Oja's rule when one maps the step sizes between the two algorithms with a specific data-dependent nonlinear relationship. We compare the two methods with static step size choices. 

\end{enumerate}

\subsection{Adaptive ROM algorithm}
\label{subsec:adaptive_algorithm_complexity}

We now explain the complete history-aware adaptive ROM framework proposed in this work. The overall procedure is presented in Algorithm~\ref{alg:adaptive_rom}, and may be summarized as follows:
\begin{itemize}
    \item [--] \textit{Offline data collection:} Advance the FOM \eqref{eq:fom_lmm}--\eqref{eq:fom_discrete_solve} over a short training window (typically for \(r\) time steps, where \(r\) is also the target ROM dimension), and assemble the corresponding training snapshot matrix.
    
    \item[--] \textit{Offline preprocessing and initialization:} Compute the reference state \(\bm q_{\mathrm{ref}}\), the scaling matrix \(\bm D\), the initial basis and singular values from the POD of the preprocessed data, the initial sampling points, and the basis-dependent reduced quantities required by the chosen ROM.
    
    \item[--] \textit{Initialization of the coarse correction trajectory:} Using the last full-order snapshot from the offline stage as initial condition, advance the coarse FOM ~\eqref{eq:coarse_fom_backward_euler} once with step size \(z\Delta t\) and store the resulting coarse snapshot. This initializes the correction signal trajectory.
    
    \item[--] \textit{Online ROM prediction:} Starting from the final offline state, advance the ROM one time step $\Delta t$ at a time using either the hyper-reduced Galerkin system \eqref{eq:deim_galerkin_rom} or the hyper-reduced LSPG system \eqref{eq:hyperreduced_lspg_stationarity}.
    
    \item[--] \textit{Adaptation event:} Every \(z\) steps, trigger model adaptation. First, advance the stored coarse full-order state once more using \eqref{eq:coarse_fom_backward_euler}. The newly obtained coarse snapshot is the correction signal used for the current adaptation event.
    
    \item[--] \textit{Basis update:} Preprocess the new correction signal and use it, together with the current basis and singular values, in the iSVD update described in Algorithm~\ref{alg:isvd_update}.
    
    \item[--] \textit{Sampling and operator update:} Since the basis has changed, recompute the sampling points using either QDEIM \eqref{eq:qdeim_qr} or FGS \eqref{eq:fgs_selection}, and update the remaining reduced quantities that depend on the basis and the sampling set.
    
    \item[--] \textit{State transfer:} Lift the current reduced state to the ambient space and project it onto the updated basis so that the ROM can continue its evolution on the new subspace.
\end{itemize}

\begin{algorithm}
\caption{History-aware adaptive reduced-order model}
\label{alg:adaptive_rom}
\begin{algorithmic}[1]
\State \textbf{Input:} target ROM dimension \(r\), number of hyper-reduction sampling points \(n_s\), offline window \(w_{\mathrm{init}}\), prediction horizon \(N_t\), adaptation frequency \(z\), forgetting factor \(\lambda\)

\State Solve the FOM \eqref{eq:fom_lmm}--\eqref{eq:fom_discrete_solve} for the first \(w_{\mathrm{init}}\) time steps and collect the training data:
\vspace{-7pt}
$$
\bm Q_{\mathrm{train}}
\coloneq
\begin{bmatrix}
\bm q^1 & \cdots & \bm q^{w_{\mathrm{init}}}
\end{bmatrix}
$$

\State Preprocess the training data:
$$
(\bm q_{\mathrm{ref}},\bm D,\widehat{\bm Q}_{\mathrm{train}})
\leftarrow
\textsc{Preprocess}(\bm Q_{\mathrm{train}})
$$

\State Initialize the basis and singular values from the POD of the preprocessed data:
\vspace{-7pt}
$$
(\bm \Phi,\bm \sigma)
\leftarrow
\textsc{POD}(\widehat{\bm Q}_{\mathrm{train}},r)
$$

\State Initialize the ROM trajectory from the last offline snapshot:
\vspace{-7pt}
$$
\bm a^{w_{\mathrm{init}}}
\leftarrow
\bm \Phi^T \bm D\left(\bm q^{w_{\mathrm{init}}}-\bm q_{\mathrm{ref}}\right)
$$

\State Initialize the correction signal trajectory by one coarse FOM step \eqref{eq:coarse_fom_backward_euler} from the last offline snapshot:
\vspace{-7pt}
$$
\bm y_{\mathrm{corr}}
\leftarrow
\textsc{CoarseFOM\_Step}_{z\Delta t}\!\left(\bm q^{w_{\mathrm{init}}}\right)
$$

\State Initialize the sampling points using \eqref{eq:qdeim_qr} or \eqref{eq:fgs_selection}:
\vspace{-7pt}
$$
\mathcal{P}
\leftarrow
\textsc{QDEIM}(\bm \Phi,n_s)
\qquad  \text{or} \qquad
\mathcal{P}
\leftarrow
\textsc{FGS}(\bm \Phi, \bm y_{\mathrm{corr}}, n_s)
$$

\State Assemble the basis-dependent quantities required by Galerkin \eqref{eq:deim_galerkin_rom} or LSPG \eqref{eq:hyperreduced_lspg_stationarity}

\For{\(n=w_{\mathrm{init}},w_{\mathrm{init}}+1,\dots,N_t-1\)}

    \State Advance the ROM by one fine step using Galerkin \eqref{eq:deim_galerkin_rom} or LSPG \eqref{eq:hyperreduced_lspg_stationarity}:
    \vspace{-7pt}
    $$
    \bm a^{n+1}
    \leftarrow
    \textsc{ROM\_Step}_{\Delta t}(\bm \Phi,\mathcal{P},\bm a^{n})
    $$

    \State Reconstruct the current ROM state:
    $$
    \tilde{\bm q}^{\,n+1}
    \leftarrow
    \bm q_{\mathrm{ref}}+\bm D^{-1}\bm \Phi \bm a^{n+1}
    $$

    \If{\(\mathrm{mod}(n+1-w_{\mathrm{init}},z)=0\)}

        \State Advance the previous correction signal one coarse FOM step using \eqref{eq:coarse_fom_backward_euler} to obtain the new correction signal:
        \vspace{-7pt}
        $$
        \bm y_{\mathrm{corr}}
        \leftarrow
        \textsc{CoarseFOM\_Step}_{z\Delta t}\!\left(\bm y_{\mathrm{corr}}\right)
        $$

        \State Preprocess the new correction signal:
        \vspace{-7pt}
        $$
        \widehat{\bm y}
        \leftarrow
        \bm D\left(\bm y_{\mathrm{corr}}-\bm q_{\mathrm{ref}}\right)
        $$

        \State Update the basis and singular values using iSVD (Algorithm \ref{alg:isvd_update}):
        \vspace{-7pt}
        $$
        (\bm \Phi,\bm \sigma)
        \leftarrow
        \textsc{iSVD}(\bm \Phi,\bm \sigma,\widehat{\bm y},\lambda,r)
        $$

        \State Update the sampling points using \eqref{eq:qdeim_qr} or \eqref{eq:fgs_selection}:
        \vspace{-7pt}
        $$
        \mathcal{P}
        \leftarrow
        \textsc{QDEIM}(\bm \Phi,n_s)
        \qquad \text{or} \qquad
        \mathcal{P}
        \leftarrow
        \textsc{FGS}(\bm \Phi, \bm y_{\mathrm{corr}}, n_s)
        $$

        \State Update the remaining basis-dependent quantities required by Galerkin \eqref{eq:deim_galerkin_rom} or LSPG \eqref{eq:hyperreduced_lspg_stationarity}

        \State Transfer the current ROM state to the updated subspace:
        \vspace{-7pt}
        $$
        \bm a^{n+1}
        \leftarrow
        \bm \Phi^T \bm D\left(\tilde{\bm q}^{\,n+1}-\bm q_{\mathrm{ref}}\right)
        $$

    \EndIf

\EndFor
\State \textbf{Return:} predicted trajectory \(\{\tilde{\bm q}^{\,n}\}_{n=w_{\mathrm{init}}+1}^{N_t}\)
\end{algorithmic}
\end{algorithm}

The iSVD basis update routine appearing in Algorithm~\ref{alg:adaptive_rom} is given in Algorithm~\ref{alg:isvd_update}. All comparison methods (gathered in Appendix~\ref{app:app_basis_update_algorithms}) considered in this work can be incorporated into the same adaptive framework simply by replacing the iSVD update with the corresponding alternative basis adaptation rule. Note that for the instantaneous methods, namely one-step, Oja's, and GROUSE, lines 13 and 14 of Algorithm~\ref{alg:adaptive_rom} should be placed at the end of the loop rather than the beginning so that the update uses the previously acquired coarse FOM signal associated with the current time level, rather than the newly generated lookahead signal.

\begin{algorithm}
\caption{iSVD update with forgetting}
\label{alg:isvd_update}
\begin{algorithmic}[1]
\State \textbf{Input:} current basis \(\bm \Phi_k \in \mathbb{R}^{N\times r}\), singular values \(\bm \sigma_k \in \mathbb{R}^r\), preprocessed correction snapshot \(\widehat{\bm y}_{k+1} \in \mathbb{R}^N\), forgetting factor \(\lambda \in [0,1]\), target rank \(r\)

\State Form \(\bm \Sigma_k = \mathrm{diag}(\bm \sigma_k)\)
\State Compute
$$
\bm p_{k+1}
\coloneq
\arg\min_{\bm p\in\mathbb{R}^r}
\left\|
\widehat{\bm y}_{k+1}-\bm \Phi_k\bm p
\right\|_2^2
$$
\State Compute the residual:
$$
\bm q_{k+1}
\coloneq
\widehat{\bm y}_{k+1}-\bm \Phi_k\bm p_{k+1}
$$
\State Form the core matrix
$$
\bm K_{k+1}
\coloneq
\begin{bmatrix}
\lambda \bm \Sigma_k & \bm p_{k+1} \\
\bm 0^T & \|\bm q_{k+1}\|_2
\end{bmatrix}
$$
\State Compute the small singular value decomposition
$$
\bm K_{k+1}
=
\overline{\bm U}_{k+1}
\overline{\bm \Sigma}_{k+1}
\overline{\bm V}_{k+1}^T
$$
\State Form the updated basis
$$
\widetilde{\bm \Phi}_{k+1}
\coloneq
\begin{bmatrix}
\bm \Phi_k & \displaystyle \frac{\bm q_{k+1}}{\|\bm q_{k+1}\|_2}
\end{bmatrix}
\overline{\bm U}_{k+1}
$$
\State Truncate to rank \(r\):
$$
\bm \Phi_{k+1}
\coloneq
\widetilde{\bm \Phi}_{k+1}(:,1\!:\!r),
\qquad
\bm \sigma_{k+1}
\coloneq
\mathrm{diag}
\bigl(
\overline{\bm \Sigma}_{k+1}(1\!:\!r,1\!:\!r)
\bigr)
$$
\State \textbf{Return:} updated basis \(\bm \Phi_{k+1}\), updated singular values \(\bm \sigma_{k+1}\)
\end{algorithmic}
\end{algorithm}

\subsection{Adaptive ROM computational complexity}
\label{subsubsec:adaptive_complexity}

We now examine the computational cost of the proposed history-aware adaptive ROM. The offline stage is intentionally minimal. If the initial training window contains \(w_{\mathrm{init}}\) fine full-order steps, then the offline cost consists of \(w_{\mathrm{init}}\) full-order solves together with the preprocessing, singular value decomposition, and initial sampling point construction. Writing the cost of one (presumably implicit) full-order step as
\begin{equation}
C_{\mathrm{FOM}}=\mathcal{O}(N^\alpha),
\qquad
\alpha > 1,
\label{eq:fom_step_cost}
\end{equation}
the offline stage scales as
\begin{equation}
C_{\mathrm{off}}
=
w_{\mathrm{init}}\,C_{\mathrm{FOM}}
+
\mathcal{O}(Nw_{\mathrm{init}}^2 + w_{\mathrm{init}}^3)
+
\mathcal{O}(Nr^2).
\label{eq:offline_cost}
\end{equation}
Here, the \(\mathcal{O}(Nw_{\mathrm{init}}^2 + w_{\mathrm{init}}^3)\) term comes from the singular value decomposition of the \(N\times w_{\mathrm{init}}\) preprocessed training matrix, while the \(\mathcal{O}(Nr^2)\) term is the cost of the initial QDEIM update through pivoted QR. In the setting considered in this work, \(w_{\mathrm{init}}\) is chosen to be very small, typically of the same order as the target ROM dimension \(r\). As a result, the offline initialization cost is negligible compared to the cost of a large-data offline training phase in typical static ROMs. The main cost of the method is therefore the online stage. Between two adaptation events, the model advances on the reduced trajectory and incurs only the cost of one fine ROM step, which we denote by \(C_{\mathrm{ROM}}\). Every \(z\) fine steps, however, an adaptation event is triggered. At such an event, the stored coarse full-order state is first advanced once by the coarse FOM to generate a new correction signal. This coarse solve scales the same as that of a fine full-order step, namely \(\mathcal{O}(N^\alpha)\). After the new correction signal has been obtained, it must be preprocessed, passed through the iSVD update, and used to refresh the basis-dependent quantities of the ROM. The preprocessing step requires only vector operations in \(\mathbb{R}^N\), and therefore scales like \(\mathcal{O}(N)\). The iSVD update is more substantial. Computing the projection coefficients \(\bm p_{k+1}\) by least squares requires multiplying the \(N\times r\) basis by reduced vectors and solving a small reduced least-squares problem, which scales like \(\mathcal{O}(Nr^2)\). Forming the residual \(\bm q_{k+1}\) and its norm is \(\mathcal{O}(Nr)\). The singular value decomposition of the \((r+1)\times(r+1)\) core matrix \(\bm K_{k+1}\) costs \(\mathcal{O}(r^3)\). Finally, rotating the augmented basis by \(\overline{\bm U}_{k+1}\) and truncating back to rank \(r\) is a matrix-matrix product involving an \(N\times(r+1)\) matrix and an \((r+1)\times(r+1)\) matrix, which again scales like \(\mathcal{O}(Nr^2)\). Altogether,
\begin{equation}
C_{\mathrm{update}}^{\mathrm{iSVD}}
=
\mathcal{O}(Nr^2+r^3).
\label{eq:isvd_cost}
\end{equation}

Once the basis has been updated, the sampling points and the remaining reduced quantities must be refreshed. In the DEIM/QDEIM setting, updating the sampling points requires a pivoted QR factorization of \(\bm \Phi^T\), where \(\bm \Phi \in \mathbb{R}^{N\times r}\). Since \(\bm \Phi^T\) has dimension \(r\times N\), this operation scales like
\begin{equation}
C_{\mathrm{sample}}^{\mathrm{QDEIM}}
=
\mathcal{O}(Nr^2).
\label{eq:qdeim_cost}
\end{equation}
The sampled-basis pseudoinverse \((\bm P^T\bm \Phi)^\dagger\) must then be recomputed. Since \(\bm P^T\bm \Phi\in\mathbb{R}^{n_s\times r}\), this involves dense reduced linear algebra with cost
\begin{equation}
C_{\mathrm{refresh}}^{\mathrm{DEIM}}
=
\mathcal{O}(n_s r^2),
\label{eq:deim_refresh_cost}
\end{equation}
and the remaining basis-dependent reduced quantities are of the same or lower order. Finally, transferring the current ROM state to the updated subspace requires lifting the reduced state to the ambient space and projecting it back onto the new basis, which costs
\begin{equation}
C_{\mathrm{transfer}}
=
\mathcal{O}(Nr).
\label{eq:transfer_cost}
\end{equation}
Combining these contributions, the cost of one adaptation event may be written as
\begin{equation}
\begin{aligned}
C_{\mathrm{adapt}}
&=
C_{\mathrm{FOM}}
+
C_{\mathrm{update}}^{\mathrm{iSVD}}
+
C_{\mathrm{sample}}^{\mathrm{QDEIM}}
+
C_{\mathrm{refresh}}^{\mathrm{DEIM}}
+
C_{\mathrm{transfer}} \\
&=
\mathcal{O}
\left(
N^\alpha + Nr^2 + r^3 + n_s r^2 + Nr
\right).
\end{aligned}
\label{eq:adapt_event_cost}
\end{equation}
For many realistic high-fidelity solvers, especially those involving sparse linear algebra, implicit nonlinear solves, or multi-physics coupling, one expects \(\alpha>1\). Moreover, in practice, we often have \(r,\, n_s\ll N\), and therefore, the reduced linear-algebra terms remain much smaller than a full-order solve that scales like \(\mathcal{O}(N^\alpha)\). In other words, in the proposed framework of Algorithm~\ref{alg:adaptive_rom}, the dominant cost of adaptation is not the subspace update itself but rather the single full-order interaction used to generate the new correction signal. This means that the adaptive framework is still substantially cheaper than advancing the FOM at every time step, because the expensive full-order interaction occurs only once every \(z\) steps, while the system otherwise evolves on the reduced trajectory. The principal operations involved in one iSVD-based adaptation event are summarized in Table~\ref{tab:adaptive_event_cost}.

\begin{table}
\centering
\caption{Cost of one adaptation event in the proposed iSVD adaptive ROM. Here, \(N\) is the FOM dimension, \(r\) is the ROM dimension, and \(n_s\) is the number of sampling points.}
\label{tab:adaptive_event_cost}
\begin{tabular}{ll}
\toprule
Operation & Cost \\
\midrule
One coarse full-order step to generate new correction signal & \(\mathcal{O}(N^\alpha);\,\alpha>1\) \\
Snapshot preprocessing & \(\mathcal{O}(N)\) \\
iSVD projection and residual formation & \(\mathcal{O}(Nr^2)\) \\
SVD of the \((r+1)\times(r+1)\) core matrix & \(\mathcal{O}(r^3)\) \\
Basis rotation and truncation & \(\mathcal{O}(Nr^2)\) \\
QDEIM sampling update & \(\mathcal{O}(Nr^2)\) \\
Update of sampled-basis pseudoinverse and reduced data & \(\mathcal{O}(n_s r^2)\) \\
Transfer of ROM state to updated basis & \(\mathcal{O}(Nr)\) \\
\bottomrule
\end{tabular}
\end{table}

To place iSVD in context, we now compare its complexity with that of the alternative basis adaptation methods considered in this work. The scaling of each method follows directly from its update formula. Windowed SVD recomputes a singular value decomposition of an \(N\times w\) data matrix at every adaptation event, which yields \(\mathcal{O}(Nw^2)\) cost. The direct method forms and processes the reduced coefficient matrix \(\bm \Phi_k^\dagger \widehat{\bm Q}_{k+1}\in\mathbb{R}^{r\times w}\), so the two dominant matrix-matrix products each scale like \(\mathcal{O}(Nwr)\), while the pseudoinverse of the \(r\times w\) reduced matrix costs \(\mathcal{O}(r^2w)\). The one-step method forms a rank-one outer-product correction in \(\mathbb{R}^{N\times r}\), which is only \(\mathcal{O}(Nr)\), but then requires orthonormalization of the updated basis, leading to the dominant \(\mathcal{O}(Nr^2)\) cost. Oja's method has the same behavior, where the streaming update itself is \(\mathcal{O}(Nr)\), while the orthonormalization step again dominates with \(\mathcal{O}(Nr^2)\). GROUSE differs in that its geodesic update preserves orthonormality by construction, so no additional orthonormalization is required; 
however, to maintain numerically orthonormal columns the projection coefficients should be obtained by least squares, and this leads to a dominant \(\mathcal{O}(Nr^2)\) cost. Finally, iSVD combines a least-squares projection, a small \((r+1)\times(r+1)\) singular value decomposition, and a basis rotation, resulting in the \(\mathcal{O}(Nr^2+r^3)\) complexity already derived above.

\begin{table}
\centering
\caption{Complexity of the basis adaptation methods considered in this work. Here, \(w\) denotes the window size for the windowed methods. The ``stored history'' column counts only the direct historical information retained beyond the current basis itself.}
\label{tab:basis_update_costs}
\begin{tabular}{lll}
\toprule
Method & Cost & Stored history \\
\midrule
Windowed SVD & \(\mathcal{O}(Nw^2)\) & \(\mathcal{O}(Nw)\) \\
Direct method & \(\mathcal{O}(Nwr+r^2w)\) & \(\mathcal{O}(Nw)\) \\
One-step method & \(\mathcal{O}(Nr^2)\) & --- \\
Oja's method & \(\mathcal{O}(Nr^2)\) & --- \\
GROUSE & \(\mathcal{O}(Nr^2)\) & --- \\
iSVD & \(\mathcal{O}(Nr^2+r^3)\) & \(\mathcal{O}(r)\) \\
\bottomrule
\end{tabular}
\end{table}

Table~\ref{tab:basis_update_costs} compares the computational complexity of all basis adaptation methods considered in this work. The main point is not only how expensive each update is, but also what direct historical information each method explicitly retains. This is summarized in the ``stored history'' column, which accounts only for information stored beyond the current basis itself. Windowed SVD and the Direct method both retain an explicit short recent history and therefore require storing and repeatedly processing a data matrix of size \(\mathcal{O}(Nw)\). In contrast, the one-step approach, Oja's rule, and GROUSE react only to the current snapshot and thus carry no explicit representation of past system behavior beyond the current basis. The proposed iSVD update occupies an intermediate and practically attractive position: it retains historical information, but only in compressed form through the vector of singular values, requiring storage of only \(\mathcal{O}(r)\). In this sense, iSVD combines the memory of history-aware methods with a storage cost that is much closer to that of instantaneous updates.

\section{Numerical experiments}
\label{sec:results}

In this section, we assess the proposed adaptive ROM framework on three problems: the one-dimensional viscous Burgers equation, the Sod shock tube, and a one-dimensional ten-species rotating detonation engine. In all cases, the focus is on future-state prediction from a deliberately short initial training window. The ROM is initialized using only a small number of early full-order snapshots and is then tasked with predicting the subsequent dynamics over a much longer horizon, thereby testing whether online basis adaptation enables prediction beyond the initial training regime.

\subsection{Burgers equation}
\label{subsec:results_burgers}

We begin with the one-dimensional viscous Burgers equation, which serves as the main development and analysis testbed for the proposed history-aware adaptive ROM. This problem contains the essential ingredients that make predictive reduced-order modeling challenging, namely nonlinear convection, diffusion, and rapidly evolving dominant solution structures. It therefore provides a controlled, yet nontrivial setting in which to motivate adaptation, compare alternative basis update mechanisms, and study the sensitivity of the proposed framework to its hyperparameters.

\subsubsection{Problem setup}

We consider the one-dimensional viscous Burgers equation
\begin{equation}
\frac{\partial u}{\partial t}
+
u\frac{\partial u}{\partial x}
=
\nu \frac{\partial^2 u}{\partial x^2},
\qquad
x \in [0,L],\qquad t \in [0,T],
\label{eq:burgers_pde_results}
\end{equation}
with periodic boundary conditions and a Gaussian initial condition
$
u(x,0)
=
\exp\!\left(
-\frac{1}{2}
\left(
\frac{x-L/2}{\sigma}
\right)^2
\right)
$
with
$
\sigma = 0.1.
$
The domain length is set to \(L=1.0\), and the FOM is discretized on \(N_x=1000\) uniform grid points, so that \(\Delta x=L/N_x\). The viscosity is fixed at \(\nu=0.01\), and the fine time step is \(\Delta t=10^{-3}\). The full prediction horizon consists of \(N_t=500\) time steps, corresponding to \(t\in[0,0.5]\). All quantities in this test case are reported in nondimensional form. The FOM is advanced with backward Euler time integration, and the nonlinear system at each time step is solved by Newton iterations with tolerance \(10^{-8}\) and at most 10 iterations. Spatial derivatives are approximated using a first-order upwind discretization for the convective term and a centered second-order finite-difference discretization for the diffusive term, both with periodic boundary conditions. The fine-time-step FOM trajectory over the full horizon is taken as the ground truth solution against which the reduced models are compared.

\subsubsection{Static LSPG--QDEIM}
\label{subsubsec:results_burgers_static}

We first construct a static ROM using the LSPG--QDEIM formulation developed in Section~\ref{sec:static_rom}. In a conventional static ROM workflow, one typically collects a relatively rich offline dataset and then chooses the ROM dimension \(r\) so as to retain a prescribed percentage of the total POD energy, often \(99\%\) or \(99.99\%\). That philosophy is not the one pursued here. The goal of this work is not to construct the best possible static ROM from an extensive offline training phase, but rather to develop adaptive ROMs that can begin from a very small initial basis and then learn online as the system evolves. In that setting, the offline phase is intentionally minimal. If the target ROM dimension is \(r\), then the smallest number of full-order snapshots capable of generating an \(r\)-dimensional trial subspace is \(r\). Accordingly, throughout this work we often take the offline training window to contain exactly \(r\) full-order steps and retain all \(r\) resulting POD modes. In that sense, we capture \(100\%\) of the energy in the available offline data.

For the baseline static ROM, we therefore choose
$
r = n_s = w_{\mathrm{init}} = 4,
$
so that the ROM is trained on only the first four full-order steps and is then asked to predict the remaining steps without any basis update. The DEIM basis is taken equal to the trial basis, and the interpolation points are obtained by QDEIM.

\begin{figure}
\centering
  \subfloat[Solution profiles for \(r=n_s=w_{\mathrm{init}}\in\{4,8,16\}\).\label{fig:burgers_static_profiles_dim}]{%
    \includegraphics[width=0.8\columnwidth]{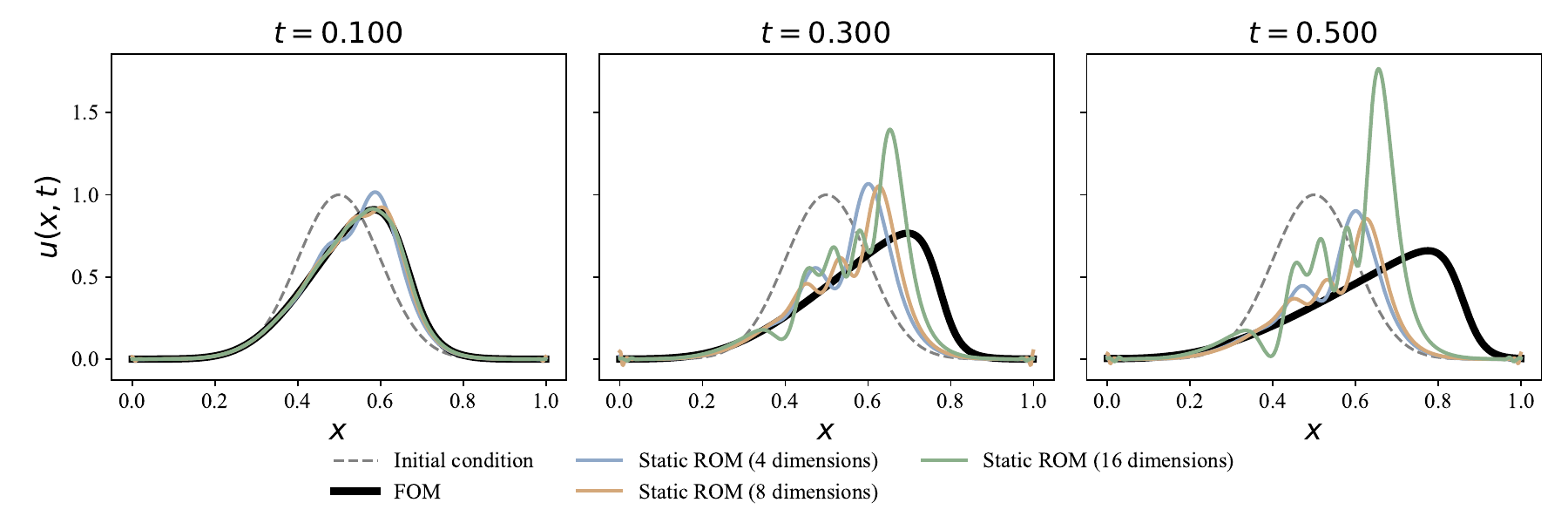}%
  }\\ \vspace{-10pt}
  \subfloat[Solution profiles for fixed \(r=n_s=4\) and varying offline training window length \(w_{\mathrm{init}}\in\{4,8,16,32,64,128,256\}\).\label{fig:burgers_static_profiles_training}]{%
    \includegraphics[width=0.8\columnwidth]{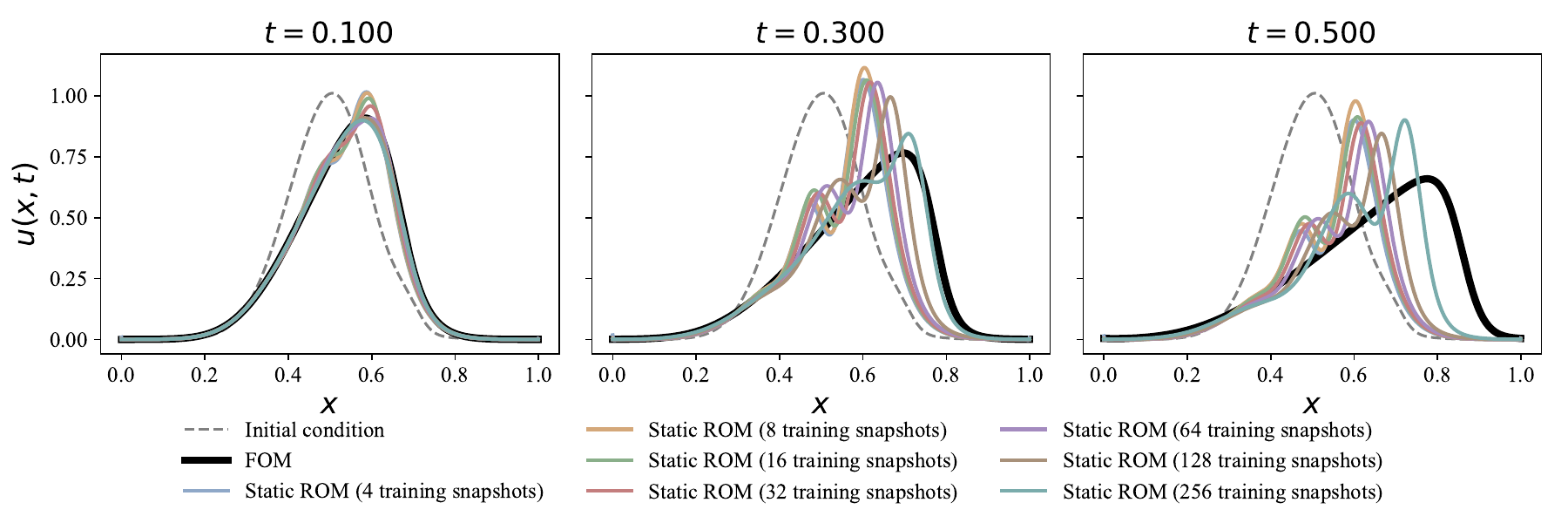}%
  }
  \caption{Static LSPG--QDEIM results for the Burgers problem. For all models, the training interval is \(t\in[0,w_{\mathrm{init}}\Delta t]\), and the testing interval is \(t\in[w_{\mathrm{init}}\Delta t,0.5]\), with $\Delta t = 10^{-3}$.}
  \label{fig:burgers_static_profiles}
\end{figure}

Figure~\ref{fig:burgers_static_profiles_dim} compares the resulting static LSPG--QDEIM trajectories against the FOM for three choices of ROM dimension,
$
r=n_s=w_{\mathrm{init}}\in\{4,8,16\}.
$
The key observation is that increasing the reduced dimension improves the approximation locally but does not resolve the underlying predictive failure. The \(r=4\) model fails earliest, as expected, while the \(r=8\) and \(r=16\) models remain accurate for somewhat longer. However, they still eventually lose the true trajectory once the solution evolves beyond the span of the training data. Thus, merely enriching the fixed trial space offline does not remove the fundamental limitation of a static basis. Even a larger static ROM remains frozen and limited to its training distribution, and therefore remains unable to adapt once the solution manifold moves away from the one found in the offline data.

To further reinforce this point, we next fix the reduced dimension and the number of DEIM points at their baseline values, \(r=n_s=4\), and vary only the offline training window length,
$
w_{\mathrm{init}}\in\{4,8,16,32,64,128,256\}.
$
In other words, we keep the ROM itself small but simply provide it with more offline data, to see if a static basis becomes sufficiently predictive. Figure~\ref{fig:burgers_static_profiles_training} depicts the results of this experiment. As the training window is extended, the static ROM is indeed exposed to a larger portion of the trajectory, and its performance improves over a longer interval. However, the essential conclusion remains unchanged. Even when the basis is trained on a substantial fraction of the trajectory, the model still struggles once it is required to extrapolate beyond the temporal region represented in the offline data. In particular, even the case \(w_{\mathrm{init}}=256\), where the model is trained on roughly half of the trajectory, does not remain predictive over the second half.

\begin{figure}
\centering
  \subfloat[Error histories for \(r=n_s=w_{\mathrm{init}}\in\{4,8,16\}\).\label{fig:burgers_static_errors_dim}]{%
    \includegraphics[width=0.4\columnwidth]{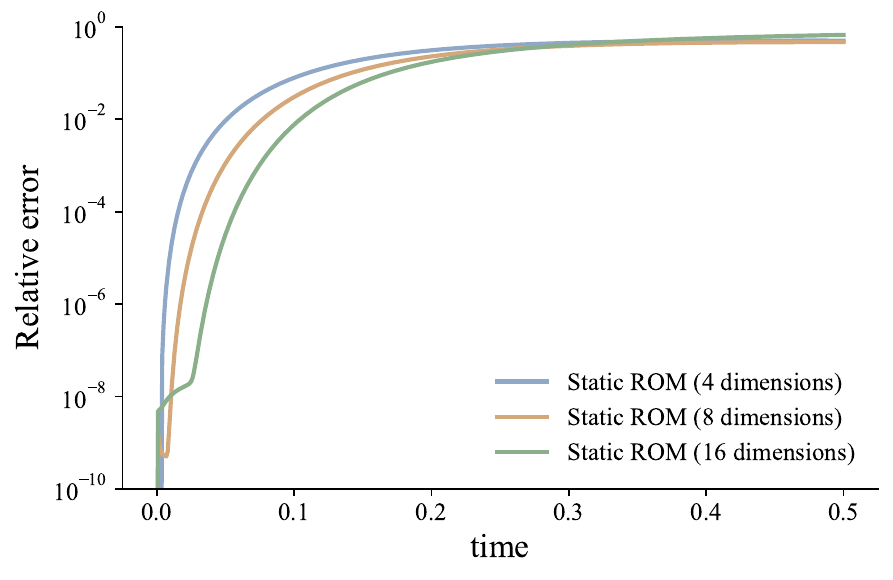}%
  }
  \hspace{10pt}
  \subfloat[Error histories for fixed \(r=n_s=4\) and varying offline training window length \(w_{\mathrm{init}}\in\{4,8,16,32,64,128,256\}\).\label{fig:burgers_static_errors_training}]{%
    \includegraphics[width=0.4\columnwidth]{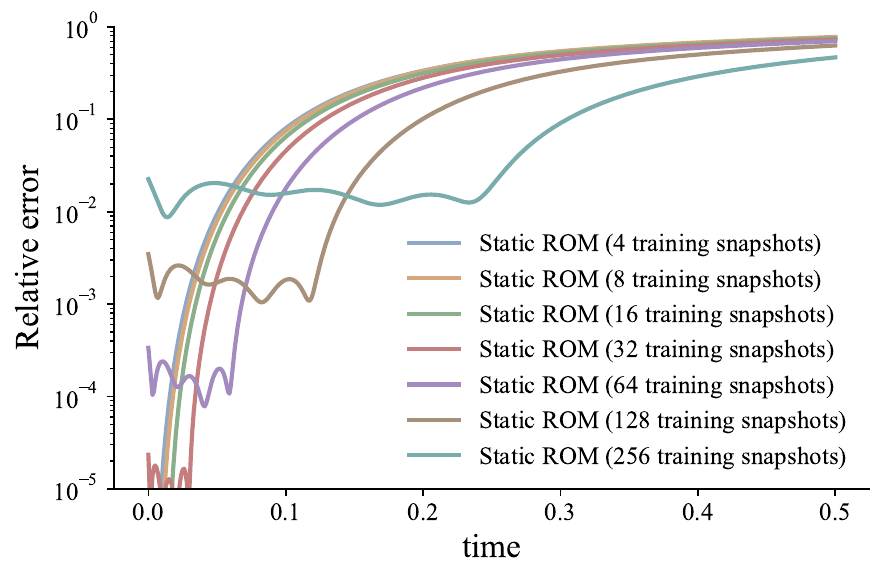}%
  }
  \caption{Relative error of the static LSPG--QDEIM ROM for the Burgers problem. For all models, the training interval is \(t\in[0,w_{\mathrm{init}}\Delta t]\), and the testing interval is \(t\in[w_{\mathrm{init}}\Delta t,0.5]\), with $\Delta t = 10^{-3}$.}
  \label{fig:burgers_static_errors}
\end{figure}

The corresponding quantitative error histories are shown in Figure~\ref{fig:burgers_static_errors}.
Throughout this work, the relative error at time level \(t^n\) is computed as
\begin{equation}
\varepsilon^n
\coloneq
\frac{
\left\|
\tilde{\bm q}^{\,n}-\bm q^n
\right\|_2
}{
\left\|
\bm q^n
\right\|_2
},
\label{eq:relative_error}
\end{equation}
where \(\bm q^n\) denotes the full-order reference state and \(\tilde{\bm q}^{\,n}\) denotes the corresponding ROM prediction.
Figure~\ref{fig:burgers_static_errors_dim} confirms that larger static models delay the breakdown but do not prevent it, and Figure~\ref{fig:burgers_static_errors_training} shows that increasing the amount of offline data has the same effect; the onset of failure is postponed, but the failure itself is not removed. Taken together, these two studies make the role of adaptation clear. Increasing the ROM dimension by enriching the offline basis or simply increasing the offline training window helps only to a limited extent. Neither addresses the core issue that the solution manifold itself evolves during prediction. Since the settings of interest in this paper are precisely those in which the offline phase is minimal and the online phase requires substantial extrapolation, the only viable path is to allow the ROM to learn online and evolve with the system.

\subsubsection{Adaptive LSPG--QDEIM}
\label{subsubsec:results_burgers_adaptive}

The adaptive ROM is built by taking the same static LSPG--QDEIM used in Section~\ref{subsubsec:results_burgers_static} and updating its basis every \(z\) time steps using the coarse FOM correction signal. The baseline adaptive configuration throughout this section is
$
r=n_s=w_{\mathrm{init}}=4.
$
Thus, the adaptive ROM begins from exactly the same minimal offline information as the static ROM, and any improvement in predictive performance is attributable entirely to online adaptation.

\begin{figure}
\centering
  \subfloat[Adaptation window $z = 5$.\label{fig:burgers_adaptive_profiles_z5}]{%
    \includegraphics[width=0.7\columnwidth]{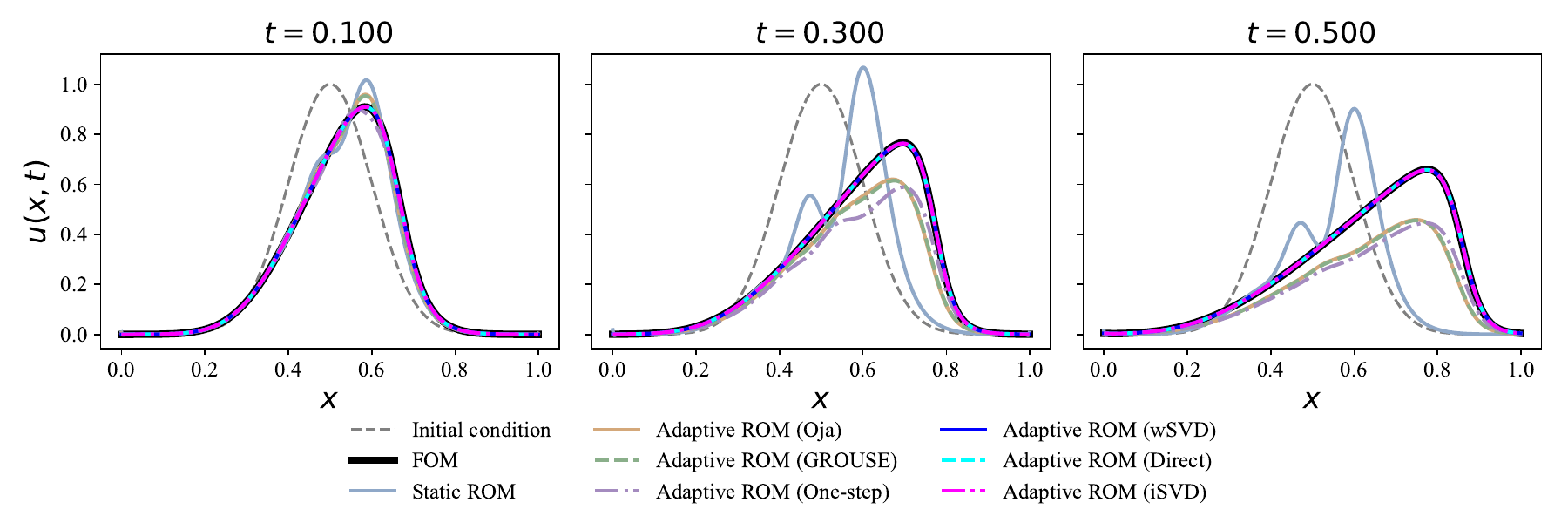}%
  }\\[-2pt]
  \subfloat[Adaptation window $z = 10$.\label{fig:burgers_adaptive_profiles_z10}]{%
    \includegraphics[width=0.7\columnwidth]{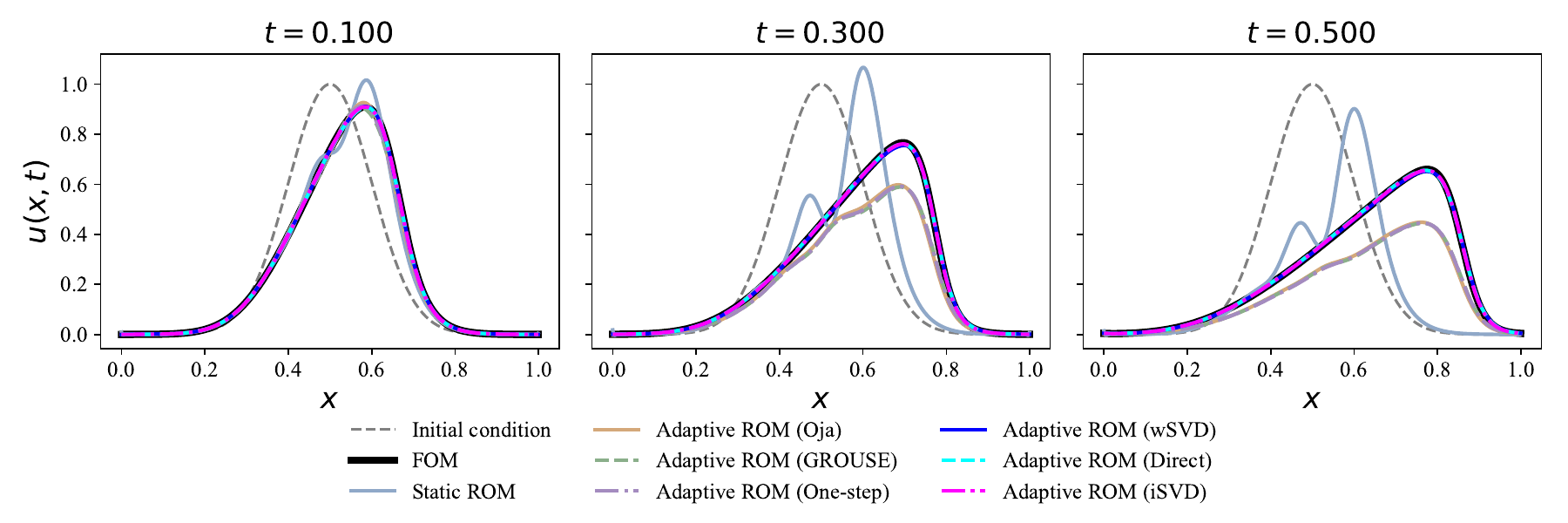}%
  }\\[-2pt]
  \subfloat[Adaptation window $z = 25$.\label{fig:burgers_adaptive_profiles_z25}]{%
    \includegraphics[width=0.7\columnwidth]{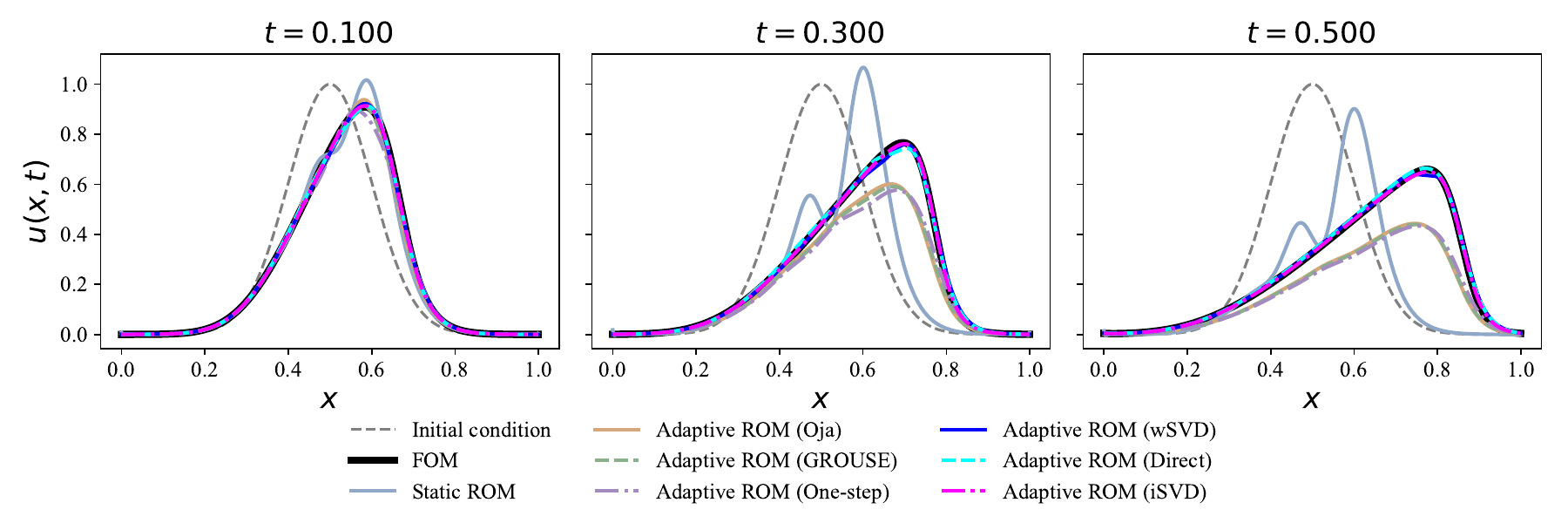}%
  }\\[-2pt]
  \subfloat[Adaptation window $z = 50$.\label{fig:burgers_adaptive_profiles_z50}]{%
    \includegraphics[width=0.7\columnwidth]{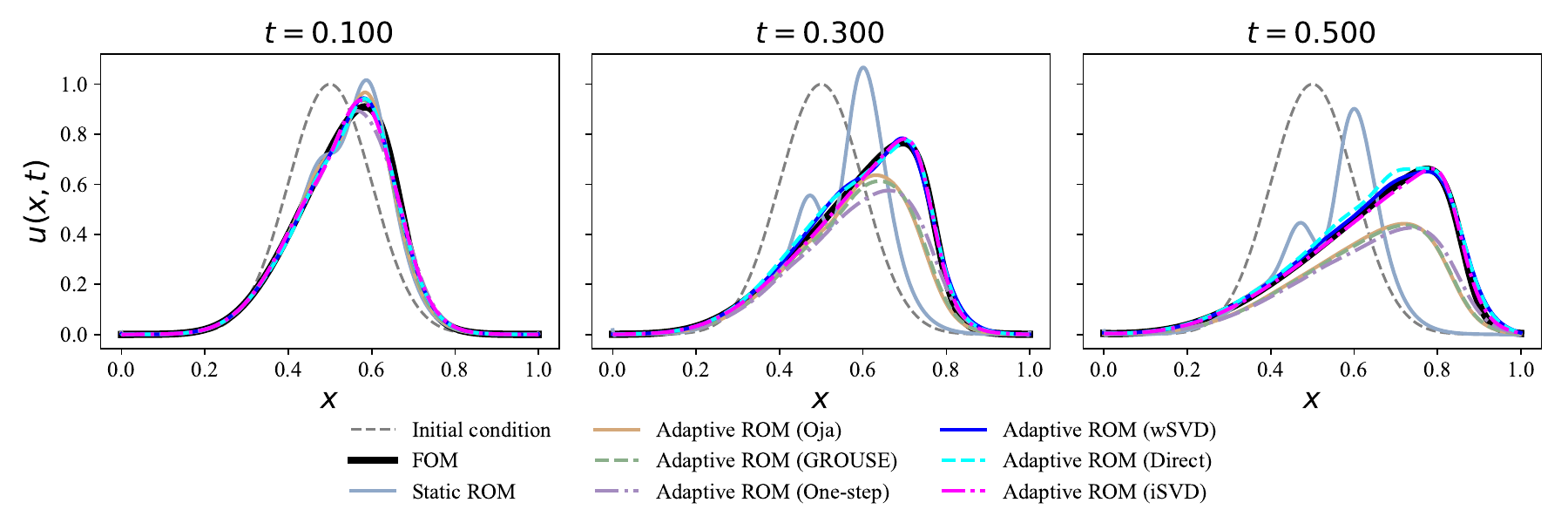}%
  }
  \caption{Adaptive LSPG--QDEIM solution profiles for the Burgers problem at different adaptation windows \(z\). For each \(z\), each method is shown with its best-performing method-specific hyperparameter. All ROMs are trained on the interval \(t\in[0,0.004]\), and tested over \(t\in[0.004,0.5]\).}
  \label{fig:burgers_adaptive_profiles}
\end{figure}

We first compare all basis adaptation strategies considered in this work, namely iSVD, windowed SVD, the Direct method, one-step updates, Oja’s rule, and GROUSE, for the following choices of adaptation window 
$
z \in \{5,10,25,50\},
$
while keeping the remaining baseline configuration fixed. Representative solution profiles are shown in Figure~\ref{fig:burgers_adaptive_profiles}. For every curve shown, the method-specific hyperparameter is chosen by sweeping over a range of candidate values and reporting the best performing configuration for that method at that value of \(z\). In particular, this means sweeping over the forgetting factor for iSVD, the window size for windowed SVD and the Direct method, and the learning rate for Oja’s method and GROUSE. The one-step method has no such additional hyperparameter. The tuned parameters are reported in Table~\ref{tab:burgers_tuning}.

\begin{table}
\centering
\caption{Tuned method-specific hyperparameters for the Burgers case at different adaptation windows \(z\). For each value of \(z\), the reported parameter is the best-performing value selected from a sweep over candidate settings for the corresponding adaptive-ROM model.}
\label{tab:burgers_tuning}
\begin{tabular}{lcccc}
\toprule
Model & $z=5$ & $z=10$ & $z=25$ & $z=50$ \\
\midrule
Oja (learning rate $\eta$) & $0.001$ & $0.01$ & $0.1$ & $1.0$ \\
GROUSE (learning rate $\eta$) & $0.001$ & $0.01$ & $0.01$ & $0.01$ \\
windowed SVD (window size $w$) & $16$ & $32$ & $64$ & $64$ \\
Direct (window size $w$) & $16$ & $8$ & $8$ & $4$ \\
iSVD (forgetting factor $\lambda$) & $0.5$ & $0.1$ & $10^{-7}$ & $10^{-7}$ \\
\bottomrule
\end{tabular}
\end{table}

A clear separation emerges between two classes of basis adaptation methods. The first class consists of methods that retain some notion of the system’s past, namely iSVD, windowed SVD, and the Direct method. The second class consists of methods that respond primarily to the current correction signal alone, namely the one-step update, Oja’s rule, and GROUSE. In particular, the iSVD adaptive ROM tracks the location, shape, and amplitude of the advecting wave profile most faithfully across the reported times, with the windowed SVD and Direct methods following the same qualitative behavior but with visibly larger discrepancies when $z$ grows. This is especially clear at later times, where the iSVD adaptive ROM continues to capture the dominant structure accurately. By contrast, the weaker methods exhibit a different failure pattern. The one-step update, Oja’s rule, and GROUSE may still respond to the newest correction snapshot, but their updates are too localized and short-sighted to maintain a coherent long-time representation of the full trajectory. As time progresses, their solution profiles show increasing shape distortion relative to the FOM, reflecting the fact that the subspace is being pulled too strongly by the most recent signal without adequately preserving the directions that remained important over the preceding evolution. The figure therefore makes clear the advantage of the history-aware methods, where they preserve a subspace that continues to represent the physically relevant manifold of the solution, whereas the instantaneous methods gradually lose that alignment.

\begin{figure}
\centering
  \subfloat[Adaptation window $z = 5$.\label{fig:burgers_adaptation_window5}]{%
    \includegraphics[width=0.4\columnwidth]{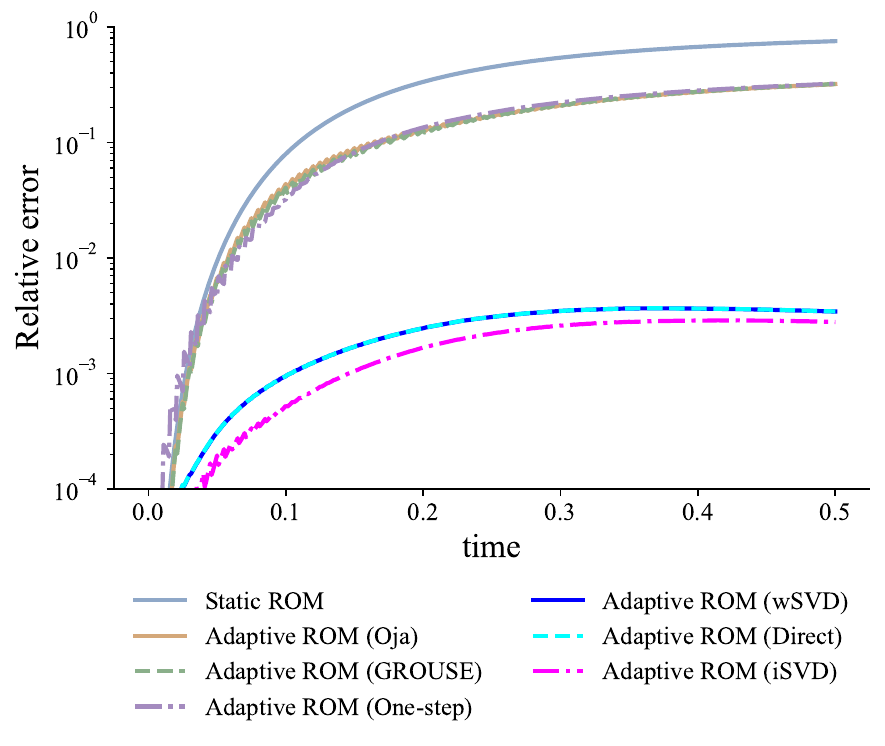}%
  }
  \hspace{10pt}
  \subfloat[Adaptation window $z = 10$.\label{fig:burgers_adaptation_window10}]{%
    \includegraphics[width=0.4\columnwidth]{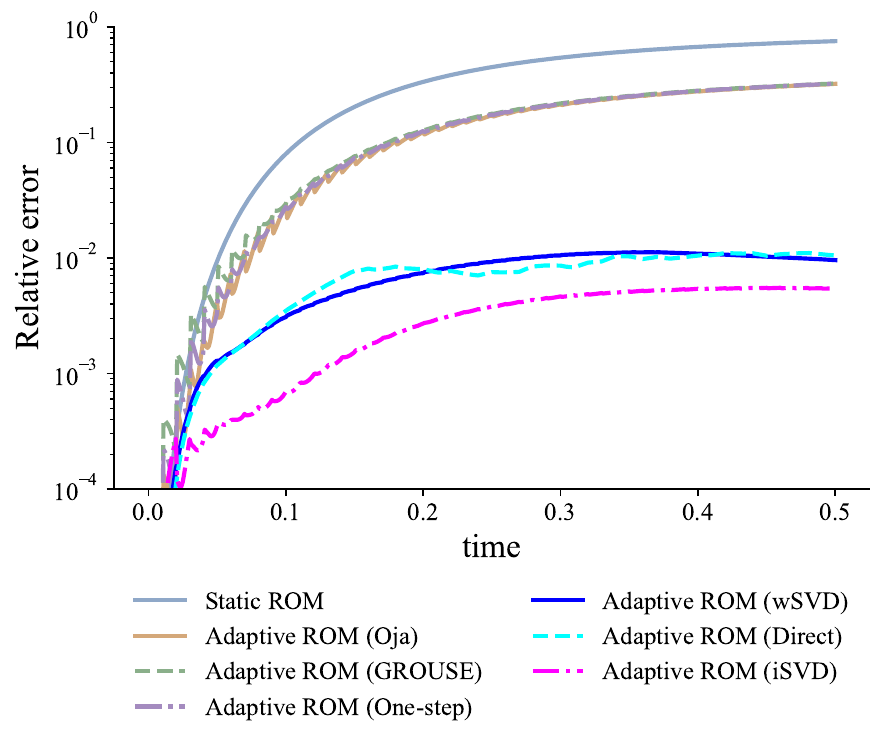}%
  }\\ \vspace{-10pt}
  \subfloat[Adaptation window $z = 25$.\label{fig:burgers_adaptation_window25}]{%
    \includegraphics[width=0.4\columnwidth]{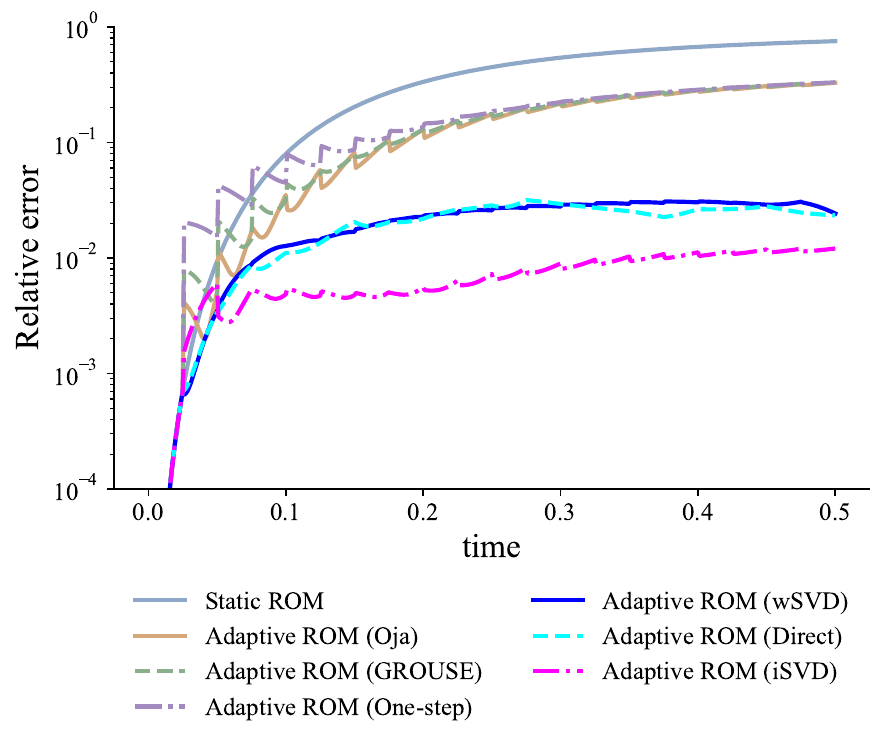}%
  }
  \hspace{10pt}
  \subfloat[Adaptation window $z = 50$.\label{fig:burgers_adaptation_window50}]{%
    \includegraphics[width=0.4\columnwidth]{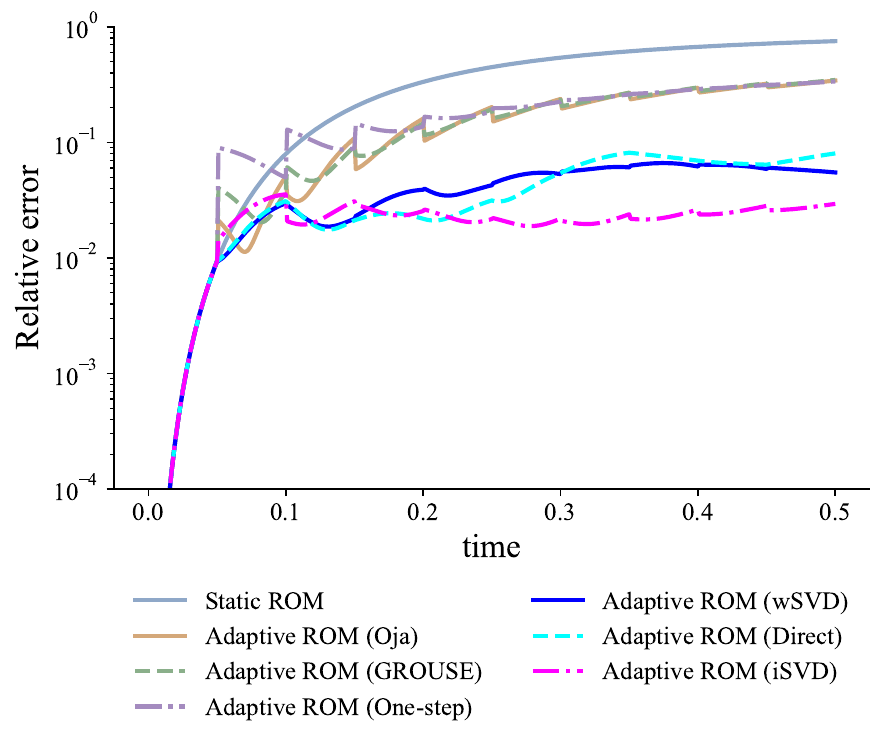}%
  }
  \caption{Relative error histories of the Burgers adaptive ROMs for different adaptation windows \(z\). For each \(z\), each method is shown with its best-performing method-specific hyperparameter. All ROMs are trained on the interval \(t\in[0,0.004]\), and tested over \(t\in[0.004,0.5]\).}
  \label{fig:burgers_adaptive_errors}
\end{figure}

The corresponding error histories are shown in Figure~\ref{fig:burgers_adaptive_errors}. The results show that the history-aware methods substantially outperform the instantaneous ones over the full prediction horizon, confirming that long-horizon predictive adaptation benefits strongly from retaining information beyond the newest observation. Among all methods, the proposed iSVD adaptive ROM delivers the best overall performance. Its error remains the smallest and grows the most slowly in time, indicating that the basis is able to evolve with the dominant solution manifold while still preserving the most relevant information accumulated from the past dynamics. Windowed SVD and the Direct method also perform significantly better than the instantaneous methods, which further supports the central idea that some explicit mechanism for carrying historical information is essential in this problem. However, both remain inferior to iSVD, suggesting that simply retaining a recent window of past states is not as effective as recursively compressing and propagating that history through the singular structure. This is a particularly important point because iSVD is also the lightest among the history-aware approaches from a memory standpoint. Windowed SVD and the Direct method both require storing a moving data matrix of size \(N\times w\), which becomes increasingly burdensome as the full-order dimension \(N\) grows. By contrast, iSVD stores only the associated vector of \(r\) singular values, so that the historical information is encoded in a highly compressed form. Computationally, iSVD also remains attractive, since its only singular value decomposition is performed on a small \((r+1)\times(r+1)\) core matrix. Thus, the results indicate not only that history matters, but that iSVD provides the most efficient way among the tested methods to carry and exploit that history.

\subsubsection{Effect of forgetting factor in iSVD}

\begin{figure}
\centering
  \subfloat[\label{burgers_forgetting_factor}]{%
    \includegraphics[width=0.45\columnwidth]{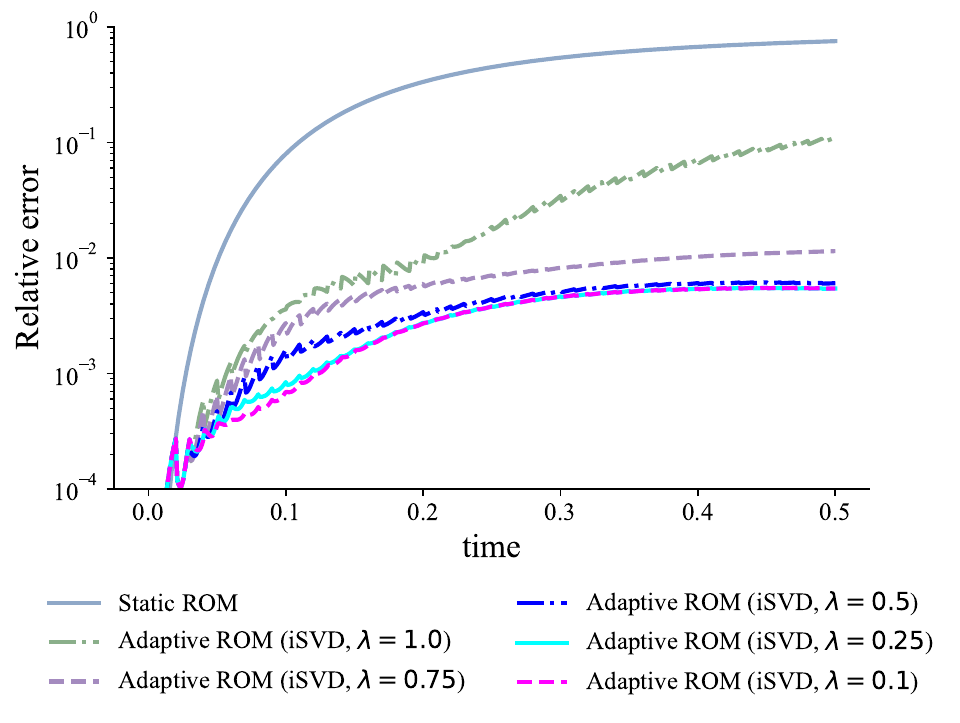}%
  }
  \caption{Effect of the forgetting factor \(\lambda\) on the iSVD adaptive ROM for the Burgers problem at fixed \(z=10\). All ROMs are trained on the interval \(t\in[0,0.004]\), and tested over \(t\in[0.004,0.5]\).}
  \label{fig:burgers_isvd_forgetting}
\end{figure}

Having established iSVD as the strongest basis update rule in the previous comparisons, we now study the role of its main hyperparameter, the forgetting factor \(\lambda\). In this experiment, we fix the adaptation window at
$
z=10
$
and vary only the forgetting factor in the iSVD update. In Figure~\ref{fig:burgers_isvd_forgetting}, we see \(\lambda=0.1\) delivers the best performance, followed by \(\lambda=0.25\), while larger values such as \(\lambda=0.75\) and \(\lambda=1.0\) perform distinctly worse. Thus, for this problem, more aggressive forgetting (smaller \(\lambda\)) is clearly beneficial. This behavior is physically intuitive, as the Burgers dynamics move rapidly away from the minimal initial training manifold, and so the basis must adapt quickly to newly emerging structures.
More generally, it can be expected that rapidly changing transport-dominated systems benefit from stronger forgetting, while more slowly evolving diffusive systems may be more compatible with longer memory.

\subsubsection{The role of the correction signal}

\begin{figure}
\centering
  \subfloat[Adaptation window $z = 5$.]{%
    \includegraphics[width=0.4\columnwidth]{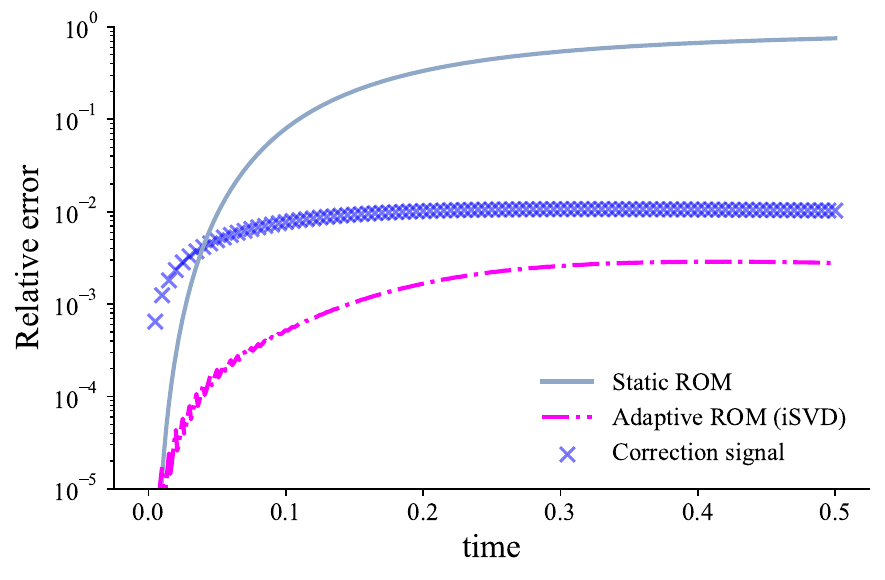}%
  }
  \hspace{10pt}
  \subfloat[Adaptation window $z = 10$.]{%
    \includegraphics[width=0.4\columnwidth]{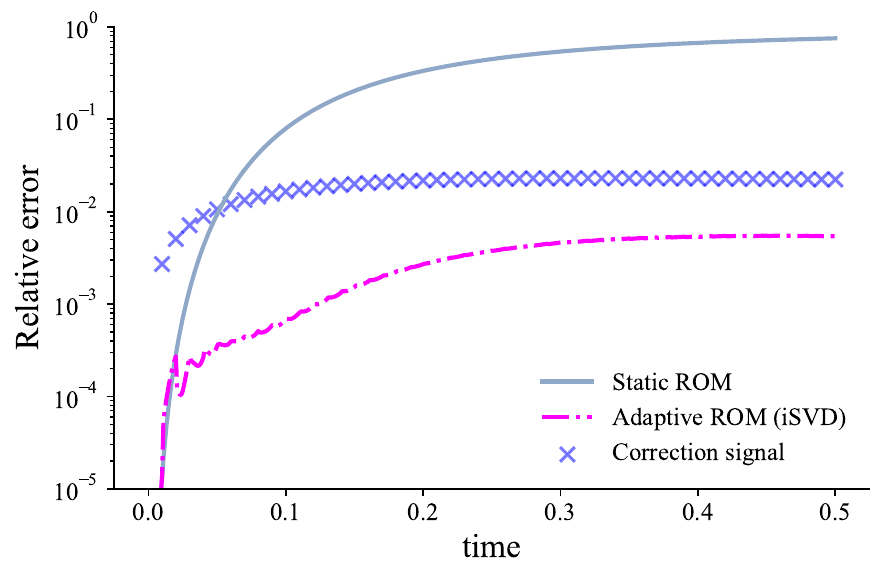}%
  }\\ \vspace{-10pt}
  \subfloat[Adaptation window $z = 25$.]{%
    \includegraphics[width=0.4\columnwidth]{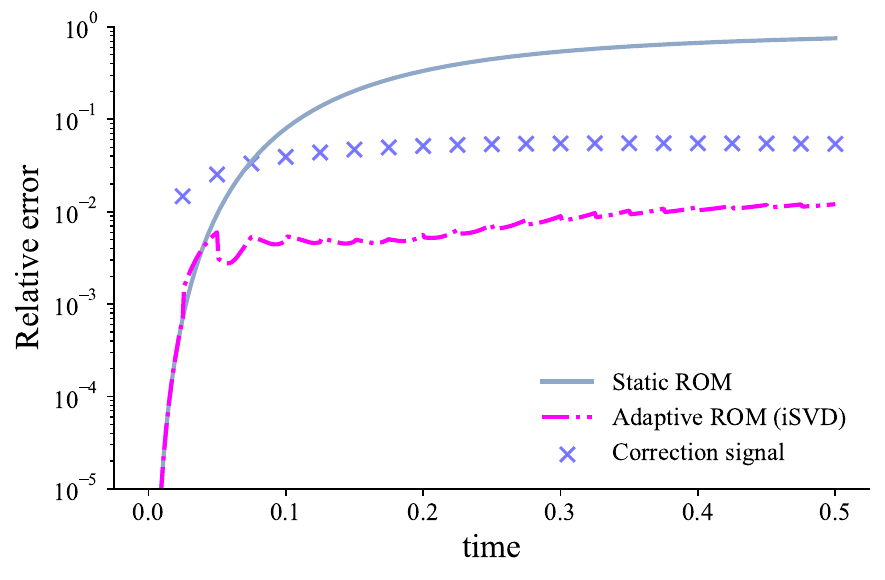}%
  }
  \hspace{10pt}
  \subfloat[Adaptation window $z = 50$.]{%
    \includegraphics[width=0.4\columnwidth]{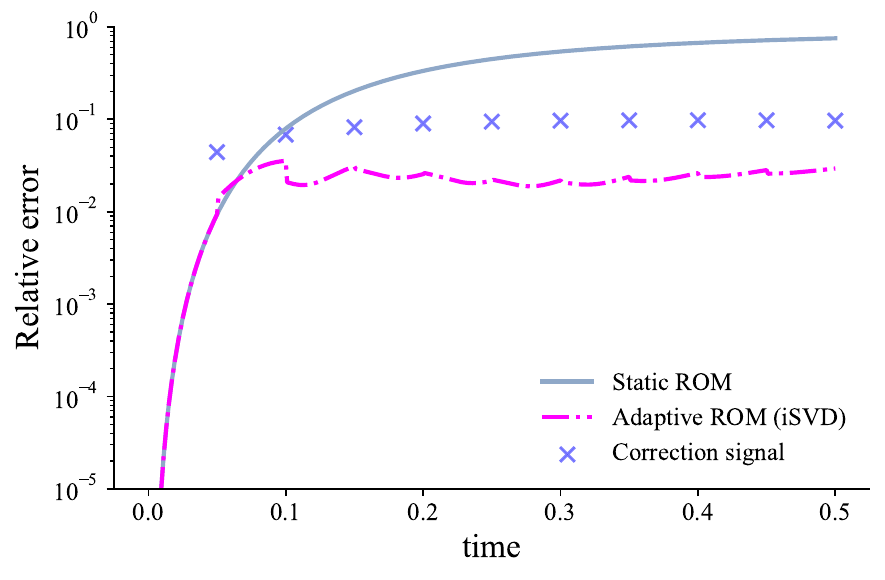}%
  }
  \caption{Comparison of the correction signal error and the iSVD adaptive ROM error for the Burgers problem over different adaptation windows \(z\). All ROMs are trained on the interval \(t\in[0,0.004]\), and tested over \(t\in[0.004,0.5]\). This figure shows that the adaptive ROM remains more accurate than the signal that informs its basis updates.}
  \label{fig:burgers_signal_vs_rom}
\end{figure}

An especially important part of the present framework is the role played by the correction signal. One may naturally ask why, if a separate coarse FOM trajectory must be run anyway, one should not simply use that coarse trajectory itself as the simulation of interest and discard the ROM altogether. To answer that, we compare the error history of the correction signal against the error history of the iSVD adaptive ROM for each adaptation window \(z\). Figure~\ref{fig:burgers_signal_vs_rom} shows that the adaptive ROM remains more accurate than the correction signal throughout the prediction horizon, even though the signal is precisely what is used to inform the basis update. This is a central conceptual point. The correction signal should not be regarded as the ground truth and used to overwrite the current reduced state. Instead, it should be viewed as a source of useful off-manifold information that indicates how the reduced subspace should rotate and evolve.
This distinction is subtle but essential. A state may be less accurate than the current ROM prediction in the ambient norm and yet still contain information that the current reduced basis lacks. In the present framework, the coarse FOM signal plays exactly this role.

\subsubsection{Runtime and acceleration}

In addition to predictive accuracy, we also report the computational performance of the adaptive models. Let \(t_{\mathrm{FOM}}\) denote the measured wall-clock time of the full-order simulation over the prediction horizon, and let \(t_{\mathrm{ROM}}\) denote the wall-clock time of a reduced-order simulation over the same horizon. We define the acceleration factor as
\begin{equation}
\lambda_{\mathrm{acc}}
\coloneq
\frac{t_{\mathrm{FOM}}}{t_{\mathrm{ROM}}}.
\label{eq:acceleration_factor}
\end{equation}
Table~\ref{tab:burgers_timing} summarizes the wall-clock times and acceleration factors of the FOM, the static LSPG--QDEIM ROM, and the adaptive ROMs (with $z=10$) obtained using the different basis update rules. The static LSPG--QDEIM ROM is the fastest model by a large margin, but this speed comes at the cost of poor long-horizon predictive accuracy, as shown earlier in this section. Thus, the static result should be interpreted as a reference for the maximum speed attainable when no online adaptation is performed. Among the adaptive models, all methods achieve acceleration factors of roughly \(4\) to \(4.5\), confirming that occasional coarse FOM interactions and the online basis adaptation that follows remain cheaper than repeated full-order simulation. The windowed SVD method is the slowest adaptive approach, which is consistent with its need to repeatedly store and process a moving data matrix. By contrast, iSVD, one-step, Oja's method, and GROUSE all exhibit very similar runtimes. This is particularly important, as it indicates that the proposed iSVD adaptive ROM delivers the strongest predictive performance while retaining essentially the same online cost as the lightest instantaneous update methods.

It should be noted that the wall-clock timings reported here are primarily informative rather than definitive measures of the acceleration potential of the framework. The Burgers problem is a lightweight model problem and is used in this work mainly as a platform for algorithmic development. In the final test case presented in this paper, we will consider a more realistic and computationally heavier simulation, whose cost better reflects the practical acceleration capabilities of the proposed adaptive ROM framework in more demanding settings.

\begin{table}
\centering
\caption{Wall-clock timings and acceleration factors for the Burgers problem (with $z=10$). All times were measured on a local machine equipped with an Apple M3
Pro chip, and are averaged over 10 runs to mitigate measurement noise.}
\label{tab:burgers_timing}
\begin{tabular}{lcc}
\toprule
Model & Wall-clock time (seconds) & Acceleration factor \(\lambda_{\mathrm{acc}}\) \\
\midrule
FOM & \(1.806\) & \(1.0\) \\
Static LSPG--QDEIM & \(0.052\) & $34.73$ \\
Adaptive ROM (iSVD) & \(0.404\) & $4.47$ \\
Adaptive ROM (windowed SVD) & \(0.483\) & $3.74$ \\
Adaptive ROM (Direct) & \(0.424\) & $4.26$ \\
Adaptive ROM (one-step) & \(0.408\) & $4.43$ \\
Adaptive ROM (Oja) & \(0.408\) & $4.43$ \\
Adaptive ROM (GROUSE) & \(0.402\) & $4.49$ \\
\bottomrule
\end{tabular}
\end{table}

\subsection{Sod shock tube}
\label{subsec:results_sod}

We next consider the one-dimensional Sod shock tube as a more challenging compressible-flow test case for the proposed history-aware adaptive ROM. In contrast to Burgers equation, this problem contains propagating discontinuities and interacting wave structures, including a shock wave, a contact discontinuity, and an expansion fan. These moving sharp features are classical failure modes for static ROMs.

\subsubsection{Problem setup}

We consider the one-dimensional compressible Euler equations on the domain
$
x \in [0,1],
$
with conserved state
$
\bm U(x,t)
=
\begin{bmatrix}
\rho \quad
m \quad
E
\end{bmatrix}^T,
$
where \(\rho\), \(m=\rho u\), and \(E\) denote the density, momentum, and total energy, respectively. The ratio of specific heats is fixed at
$
\gamma = 1.4.
$
The initial condition is the standard Sod Riemann problem, with a discontinuity at the midpoint of the domain:
\begin{equation*}
(\rho,u,p)(x,0)
=
\begin{cases}
(1.0,\,0,\,1.0), & x < 0.5,\\[1mm]
(0.125,\,0,\,0.1), & x > 0.5.
\end{cases}
\label{eq:sod_ic_results}
\end{equation*}

The FOM is discretized on
$
N_x = 256
$
uniform finite-volume cells over the unit domain, so that \(\Delta x = 1/N_x\). Time integration is performed with backward Euler using the time step
$
\Delta t = 2.5\times 10^{-4},
$
and the solution is evolved for
$
N_t = 500
$
time steps. All quantities are reported in nondimensional form. At each time step, the resulting nonlinear system is solved by Newton iterations with tolerance \(10^{-8}\) and at most 10 iterations. The spatial discretization uses a cell-centered finite-volume formulation with a local Lax--Friedrichs (Rusanov-type) numerical flux. Primitive variables are recovered from the conserved state at each iteration. At the boundaries, the end states are copied into ghost values, which corresponds to a simple zero-gradient extension. A concise statement of the governing equations is given in Appendix~\ref{app:app_sod}.

\subsubsection{Adaptive LSPG--QDEIM}

For this case, we again employ the intrusive LSPG--QDEIM framework developed in Sections~\ref{sec:static_rom} and \ref{sec:adaptive_rom}. The reduced basis is constructed from an initially short full-order training window and then updated online using Algorithm~\ref{alg:adaptive_rom}. We choose the baseline dimensions as
$
r=n_s=w_{\mathrm{init}}=4.
$
Based on the conclusions already established in the Burgers study, we focus here on the strongest basis update rules, namely iSVD and the Direct method. We also include the one-step approach as a representative of the instantaneous basis adaptation methods. We report the results for the adaptation window
$
z=10.
$
Results for the other adaptation windows,
$
z\in\{5,15,20\},
$
are collected in Appendix~\ref{app:app_sod_additional_results}.
For each value of \(z\), the method-specific hyperparameter of each adaptive update is selected by sweeping over a range of candidate values and retaining the best performing configuration. The tuned parameters are reported in Table~\ref{tab:sod_tuning}.
For the Direct method, the window size \(w\) controls a rolling state history that is updated at every adaptation event; increasing \(w\) can stabilize the update, but it can also retain older snapshots in which the shock, contact discontinuity, and expansion fan occupy different spatial locations. Thus, a larger window is not necessarily better, and the optimal value reflects a balance between temporal smoothing and the staleness of past wave configurations. For iSVD, smaller forgetting factors become favorable at larger \(z\) because each correction event occurs after a longer time interval and the incoming lookahead snapshot can differ more substantially from the previously encoded subspace. In that regime, the best update is more aggressive, placing greater emphasis on the newly received correction information.

\begin{table}
\centering
\caption{Tuned method-specific hyperparameters for the Sod shock tube case at different adaptation windows \(z\). For each value of \(z\), the reported parameter is the best-performing value selected from a sweep over candidate settings for the corresponding adaptive-ROM model.}
\label{tab:sod_tuning}
\begin{tabular}{lcccc}
\toprule
Model & $z=5$ & $z=10$ & $z=15$ & $z=20$ \\
\midrule
Direct (window size $w$) & $32$ & $8$ & $32$ & $16$ \\
iSVD (forgetting factor $\lambda$) & $0.25$ & $0.01$ & $10^{-8}$ & $10^{-8}$ \\
\bottomrule
\end{tabular}
\end{table}

\begin{figure}
\centering
  \subfloat[Density profiles.]{%
    \includegraphics[width=0.9\columnwidth]{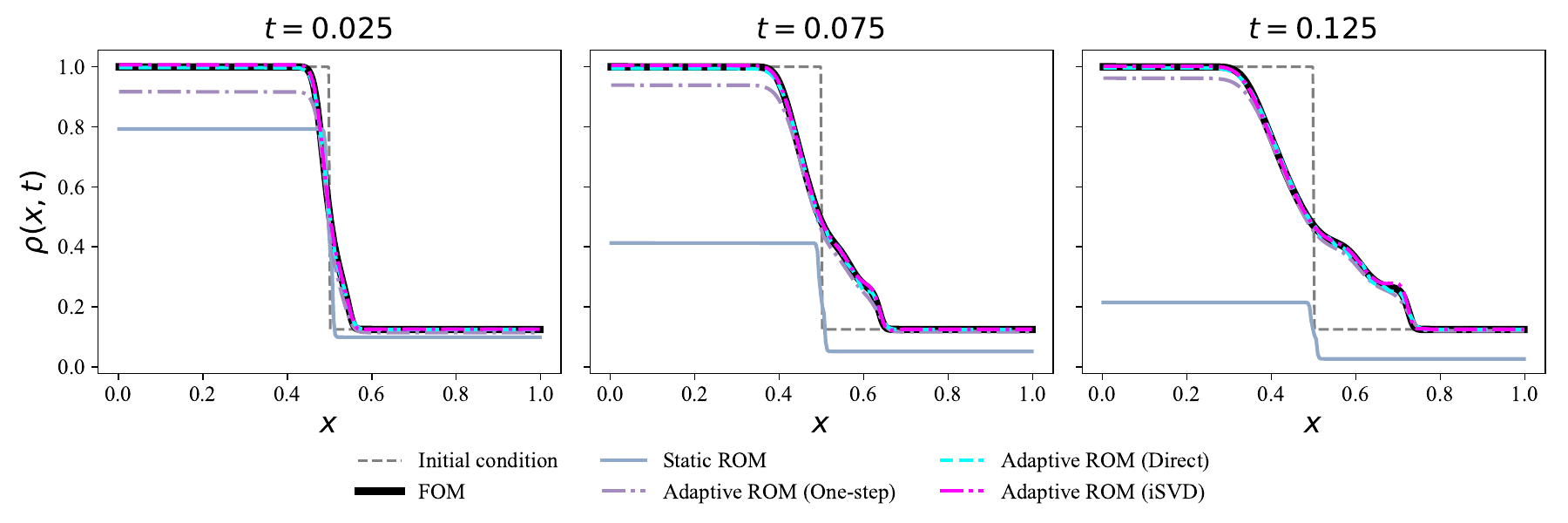}%
  }\\ \vspace{-10pt}
  \subfloat[Velocity profiles.]{%
    \includegraphics[width=0.9\columnwidth]{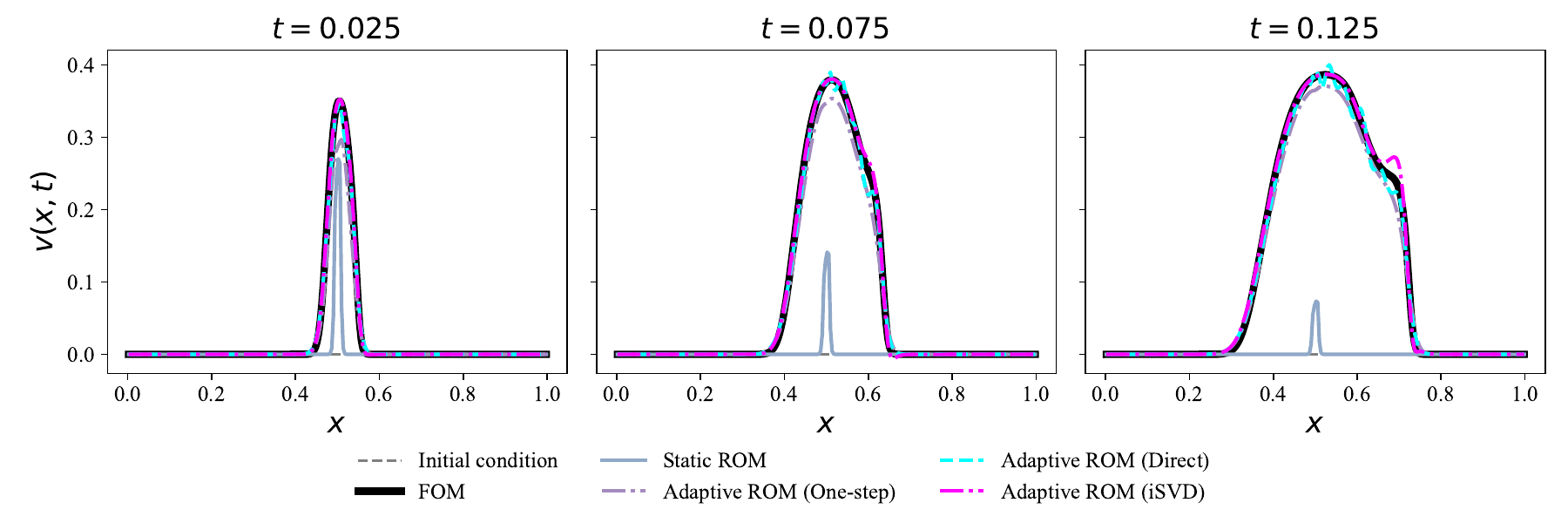}%
  }\\ \vspace{-10pt}
  \subfloat[Pressure profiles.]{%
    \includegraphics[width=0.9\columnwidth]{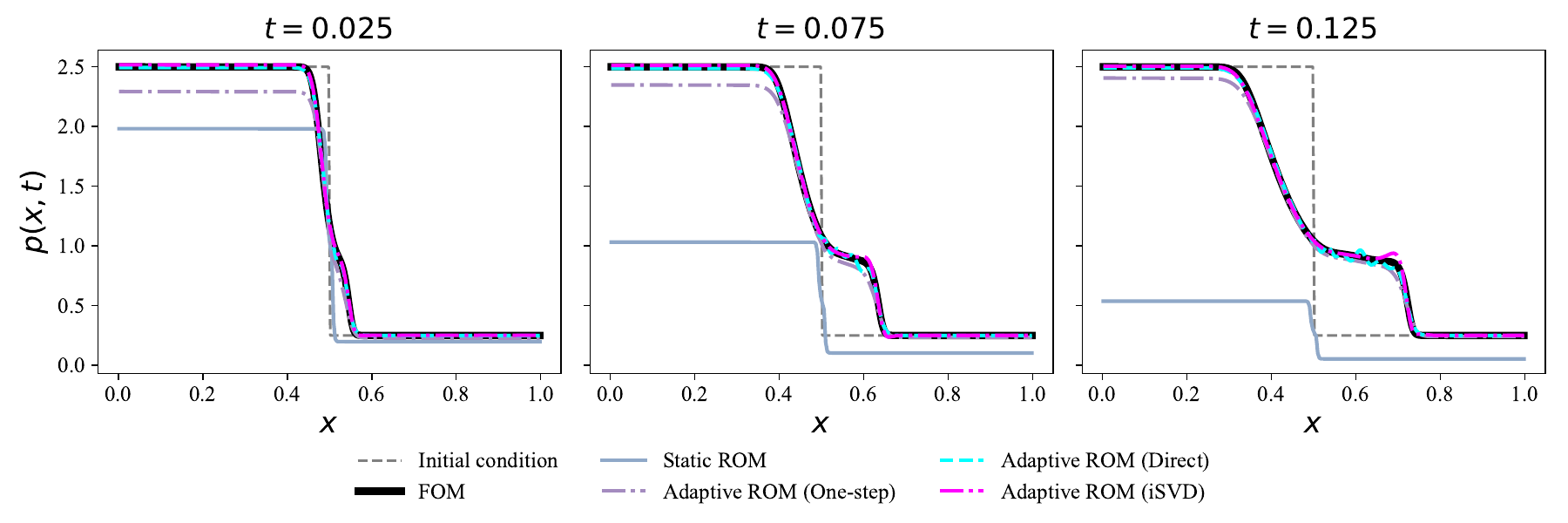}%
  }
  \caption{Sod shock tube solution profiles for adaptation window \(z=10\). All ROMs are trained on the interval \(t\in[0,0.001]\), and tested over \(t\in[0.001,0.125]\).}
  \label{fig:sod_profiles_z10}
\end{figure}

Figure~\ref{fig:sod_profiles_z10} compares the density, velocity, and pressure profiles produced by the FOM, the static ROM, the one-step adaptive ROM, the Direct adaptive ROM, and the proposed iSVD adaptive ROM at three instances of time. The static ROM fails in the expected way; once the shock, contact discontinuity, and expansion fan move beyond the structures represented in the initial training data, the fixed basis can no longer provide a reliable predictive representation. The one-step adaptive ROM improves substantially over this static baseline. However, it still shows visible degradation early on as the solution evolves. By contrast, the two history-aware methods remain much closer to the FOM throughout the prediction horizon. Both the Direct adaptive ROM and the proposed iSVD adaptive ROM capture the dominant wave pattern well. Between these two, the iSVD-based ROM is consistently the more faithful approximation. This is particularly visible at later times, where the Direct method begins to exhibit small oscillations, while iSVD remains more tightly matched to the FOM. For iSVD specifically, at the final reported time, a slight overshoot becomes visible near the shock in the velocity and pressure profiles. Even with that small discrepancy, however, the iSVD-based ROM still provides the best overall agreement with the FOM among the reduced models shown.

\begin{figure}
\centering
  \subfloat[Relative error in density.]{%
    \includegraphics[width=0.3\columnwidth]{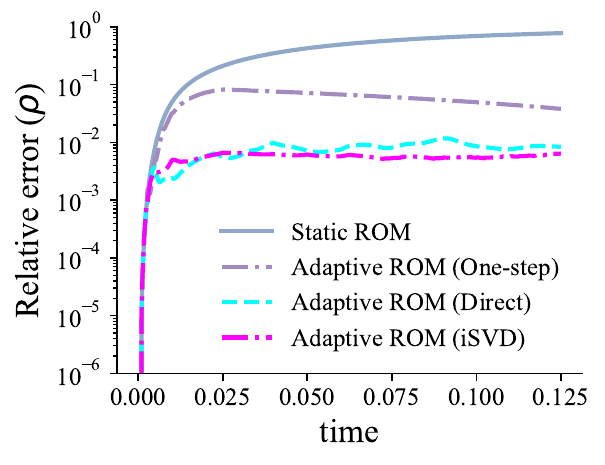}%
  }
  \subfloat[Relative error in velocity.]{%
    \includegraphics[width=0.3\columnwidth]{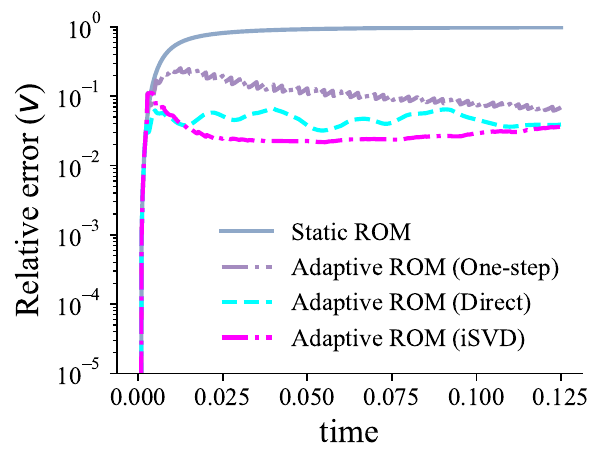}%
  }
  \subfloat[Relative error in pressure.]{%
    \includegraphics[width=0.3\columnwidth]{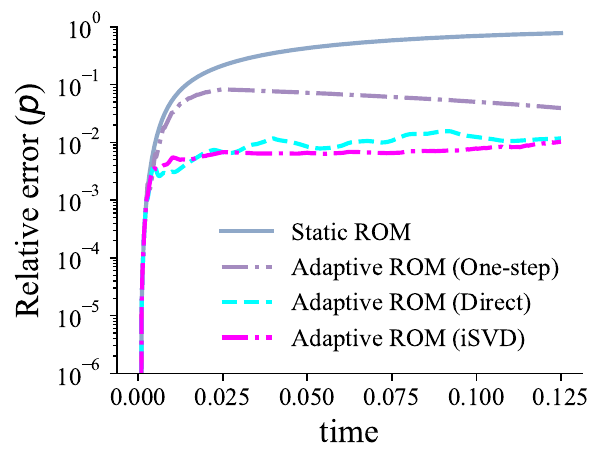}%
  }
  \caption{Relative error histories for the Sod shock tube at adaptation window \(z=10\). All ROMs are trained on the interval \(t\in[0,0.001]\), and tested over \(t\in[0.001,0.125]\).}
  \label{fig:sod_errors_z10}
\end{figure}

The quantitative comparison is given by the relative error histories in Figure~\ref{fig:sod_errors_z10}. These results support the same ordering suggested by the solution profiles. The static ROM is by far the least accurate model in all three primitive variables. The one-step adaptive ROM reduces the error relative to the static baseline, but its error remains consistently above those of the history-aware methods. The Direct adaptive ROM performs significantly better and remains competitive over the trajectory, but the proposed iSVD adaptive ROM maintains the lowest overall error across all three reported variables.

\subsubsection{Runtime and acceleration}

Table~\ref{tab:sod_timing} reports the wall-clock times and acceleration factors for the FOM, the one-step adaptive ROM, the Direct adaptive ROM, and the proposed iSVD adaptive ROM in the Sod shock tube setting. The reported values correspond to the adaptation window \(z=10\). All three adaptive ROMs provide speedup over the FOM. Among them, the proposed iSVD adaptive ROM is the fastest, narrowly outperforming the one-step method and more clearly improving upon the Direct adaptive ROM. These timing results are consistent with the trend already observed in the simpler Burgers problem, while the most compelling acceleration comparison remains the RDE case in the following section.

\begin{table}
\centering
\caption{Wall-clock timings and acceleration factors for the Sod shock tube case (with $z=10$). All wall-clock times were measured on a local machine equipped with an Apple M3 Pro chip, and are averaged over 10 runs to mitigate measurement noise.}
\label{tab:sod_timing}
\begin{tabular}{lcc}
\toprule
Model & Wall-clock time (seconds) & Acceleration factor \(\lambda_{\mathrm{acc}}\) \\
\midrule
FOM & $6.126$ & $1.0$ \\
Adaptive ROM (One-step) & $1.248$ & $4.91$ \\
Adaptive ROM (Direct) & $1.336$ & $4.58$ \\
Adaptive ROM (iSVD) & $1.236$ & $4.96$ \\
\bottomrule
\end{tabular}
\end{table}

\subsection{Rotating detonation engine}
\label{subsec:results_rde}

We finally consider a one-dimensional premixed rotating detonation engine (RDE)~\citep{Mohaghegh2026CompFlowLab}, which serves as the most challenging test case in this work. In contrast to Burgers equation and the Sod shock tube, the present problem combines strong convection, sharp wave fronts, detailed multi-species chemistry, and repeated wave propagation in a periodic domain. It is therefore a particularly demanding setting for adaptive reduced-order modeling, and it provides the most application-driven validation case for the proposed history-aware iSVD-based framework.

\subsubsection{Problem setup}

For this case, we follow the same one-dimensional RDE configuration used in the existing adaptive ROM study presented in~\cite{Mohaghegh2026selfadaptive}. The computational domain is one-dimensional in the azimuthal direction and has a length
$
L = 287\ \mathrm{mm},
$
discretized using \(2000\) uniformly distributed cells, so that
$
\Delta x = 0.1435\ \mathrm{mm}.
$
Periodic boundary conditions are imposed in the azimuthal direction. Combustion is modeled using a detailed \(\mathrm{H_2\!-\!O_2}\) reaction mechanism~\citep{Oran1982} with 10 species and 24 reactions. The full-order simulation is initialized from a ramp-shaped pressure profile designed to trigger a single detonation front and is then advanced from \(t=0\) to \(t=1.5\,\mathrm{ms}\) using the fine time step
$
\Delta t = 0.5\ \mathrm{ns},
$
resulting in a total of $N_t=3\,000\,000$ prediction steps.
The inlet is modeled through a simplified injection rule based on injector conditions \((p,T,v)\) and an isentropic critical-pressure condition following the previous work by~\cite{Huang2024parametric}. When the local pressure exceeds the critical value, injection is turned off; otherwise, constant inlet pressure, temperature, and velocity are imposed. To assess the method under representative operating conditions, we consider the same family of inlet conditions used in the reference study. In particular, we focus on the equivalence ratio $\phi = 1.0$, the inlet pressure $P_{\mathrm{in}} = 4$, and inlet velocity \(V_{\mathrm{in}}=200\,\mathrm{m/s}\). Moreover, the inlet temperature is fixed at \(300\,\mathrm{K}\). A concise statement of the governing equations and boundary model is given in Appendix~\ref{app:app_rde}. To simulate the system, an in-house CFD code\footnote{\url{https://github.com/alimike97/CompFlowLab}} is employed for both the FOM and ROMs.

\subsubsection{Adaptive Galerkin--FGS}

Because the full-order solver for this case is explicit, the reduced model is constructed in the Galerkin form rather than in the LSPG setting used for Burgers equation and the Sod shock tube. Also, hyper-reduction is performed using FGS instead of QDEIM to better capture localized flow structures appearing in RDE. For adaptive ROMs, we focus on the iSVD and the Direct method for basis adaptation, while both methods use the same coarse FOM correction signal mechanism. For the Direct baseline, we follow the implementation used in~\citep{mohaghegh2026feature} and~\citep{Mohaghegh2026selfadaptive}, where the basis is updated at each time step using the Direct method, while the sampling points are refreshed only every \(z\) time steps when a new lookahead coarse FOM signal becomes available. The proposed iSVD adaptive ROM follows the standard adaptation method described in Algorithm~\ref{alg:adaptive_rom}, where both basis and sampling points are adapted every \(z\) time steps. Therefore, the iSVD-based method adapts the basis less often compared to the version of the Direct method implemented for this test case.
Both models are initialized using
$
r = w_{\mathrm{init}} = z = 10
$
and
$
n_s = 20,
$
with the first 10 sample points selected by QDEIM and the remaining 10 selected by the feature-guided strategy. Pressure gradients are used as the feature indicator for sample placement, which is a natural choice for detonation wave tracking in this setting.

\begin{figure}
\centering
  \subfloat[Density profiles.]{%
    \includegraphics[width=0.75\columnwidth]{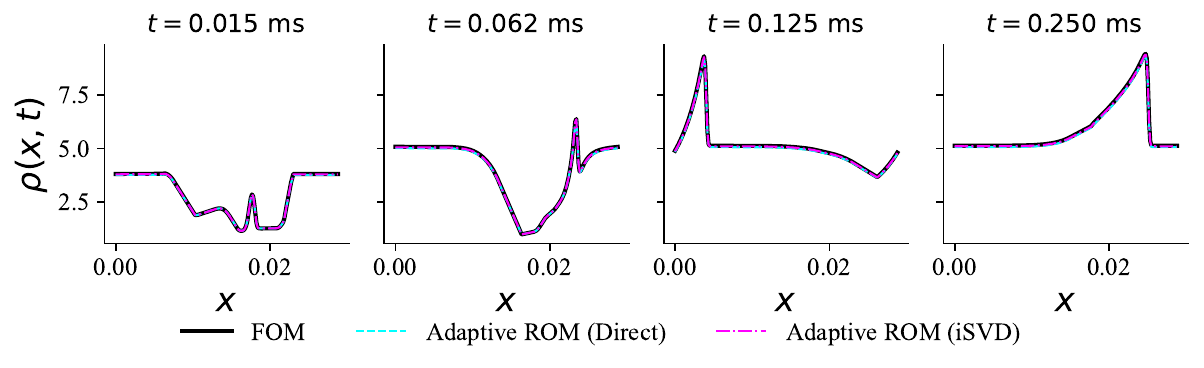}%
  }\\ \vspace{-10pt}
  \subfloat[Velocity profiles.]{%
    \includegraphics[width=0.75\columnwidth]{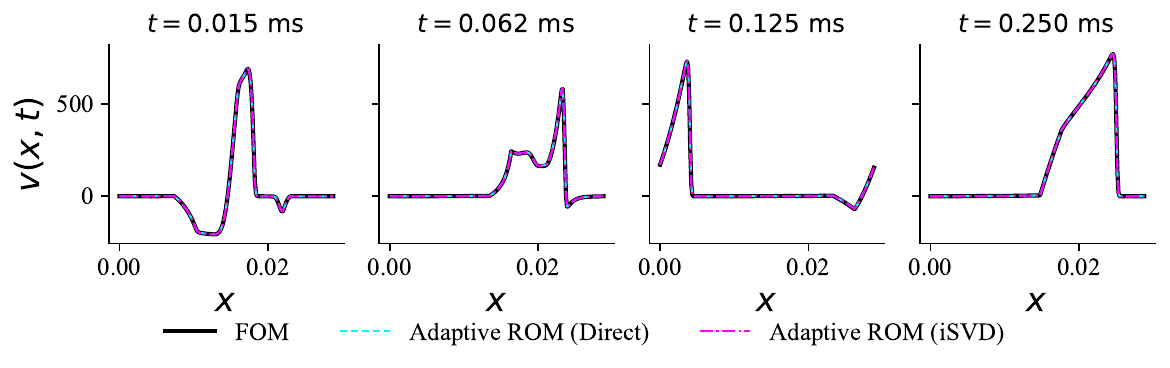}%
  }\\ \vspace{-10pt}
  \subfloat[Pressure profiles.]{%
    \includegraphics[width=0.75\columnwidth]{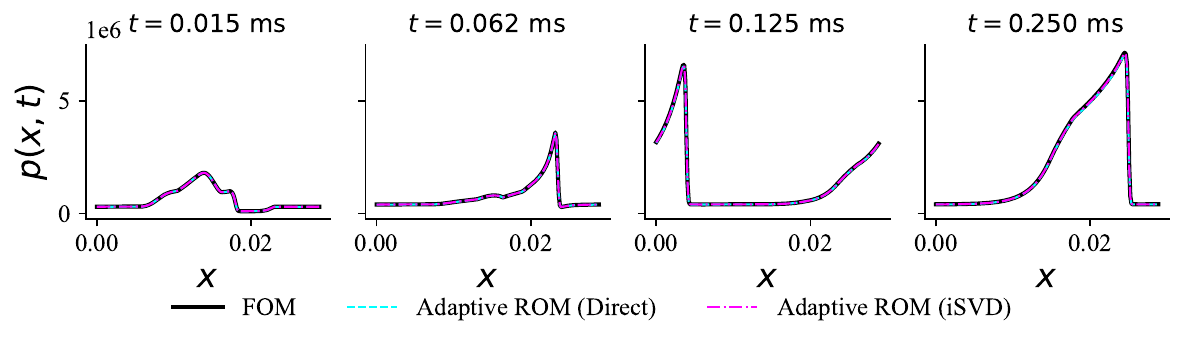}%
  }\\ \vspace{-10pt}
  \subfloat[Temperature profiles.]{%
    \includegraphics[width=0.75\columnwidth]{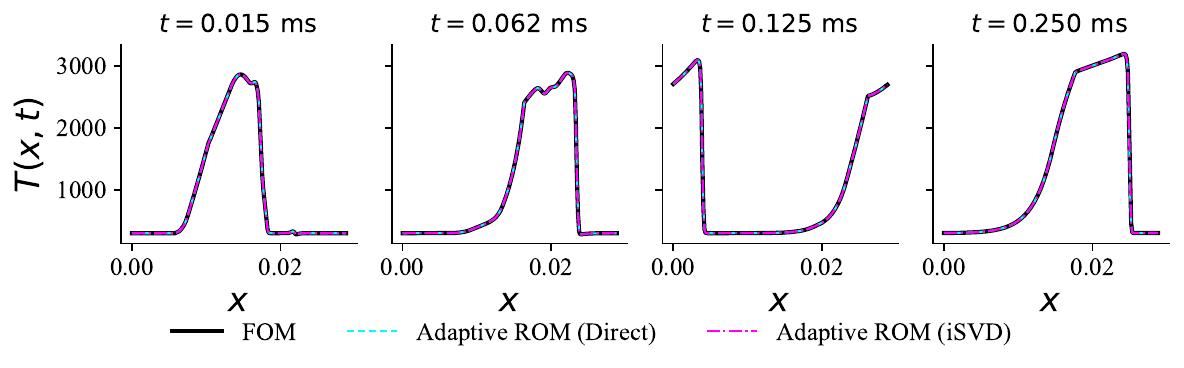}%
  }
  \caption{Solution profiles for the one-dimensional RDE case within its transient phase. Both ROMs are trained on the interval \(t\in[0,5\times10^{-6}\,\mathrm{ms}]\), and tested over \(t\in[5\times10^{-6}\,\mathrm{ms},1.5\,\mathrm{ms}]\).}
  \label{fig:rde_profiles}
\end{figure}

Figure~\ref{fig:rde_profiles} compares the density, velocity, pressure, and temperature profiles produced by the FOM, the Direct adaptive ROM, and the proposed iSVD adaptive ROM at four time instances spanning the transient phase of the solution (a total of $500,000$ prediction steps). Across all reported times, both adaptive ROMs provide a reasonable approximation of the FOM, indicating that both methods are capable of capturing the development of the detonation structure. The difference between the two adaptive updates becomes more visible at the final time, particularly when looking at the density and pressure profiles. Close inspection of these two reveals that the iSVD-based ROM remains more tightly aligned with the FOM. This distinction, even though small, is crucial since the sharpness and placement of the detonation front are central physical features of the solution.

\begin{figure}
\centering
  \subfloat[Relative error in density.]{%
    \includegraphics[width=0.35\columnwidth]{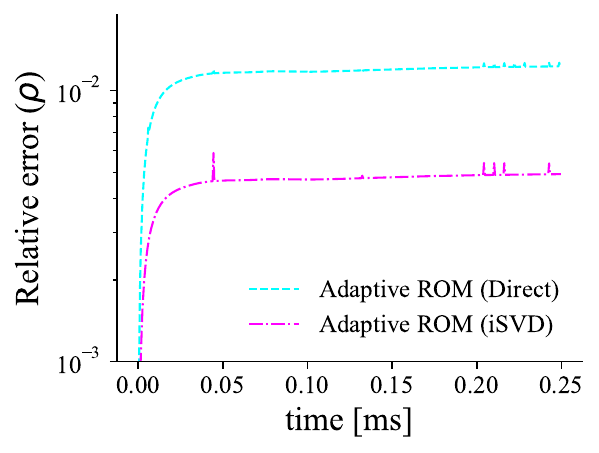}%
  }
  \subfloat[Relative error in velocity.]{%
    \includegraphics[width=0.35\columnwidth]{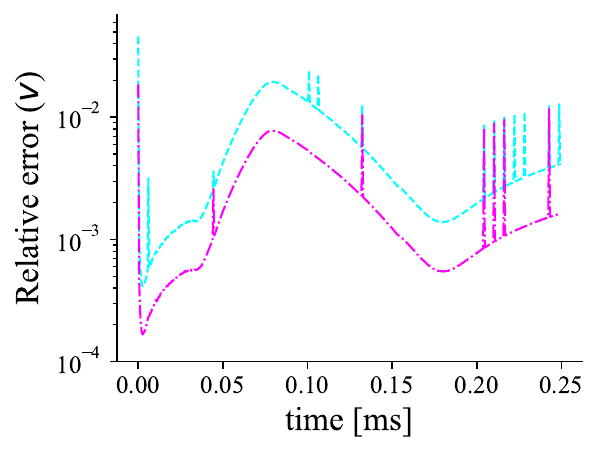}%
  }\\ \vspace{-10pt}
  \subfloat[Relative error in pressure.]{%
    \includegraphics[width=0.35\columnwidth]{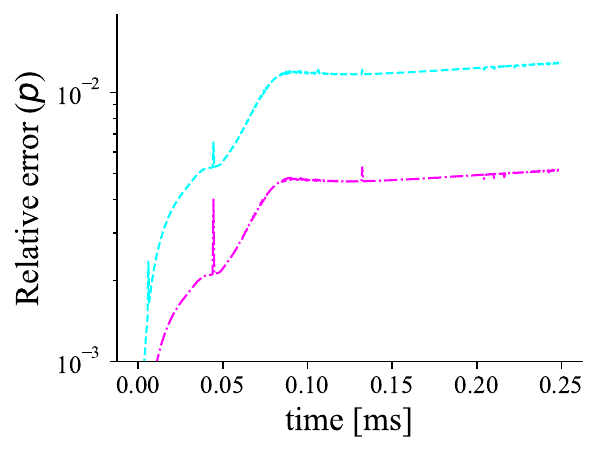}%
  }
  \subfloat[Relative error in temperature.]{%
    \includegraphics[width=0.35\columnwidth]{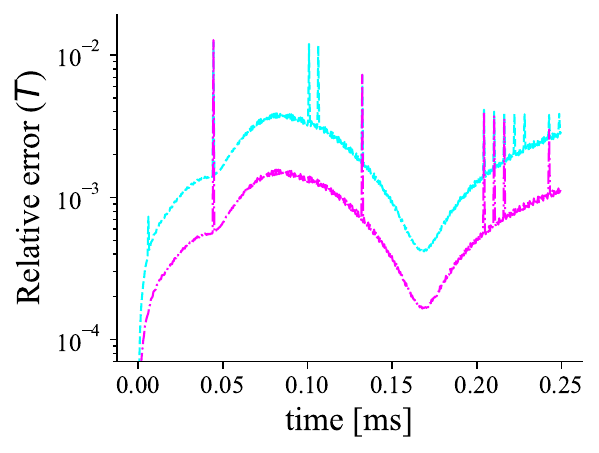}%
  }
  \caption{Relative error histories for the one-dimensional RDE case within its transient phase. Both ROMs are trained on the interval \(t\in[0,5\times10^{-6}\,\mathrm{ms}]\), and tested over \(t\in[5\times10^{-6}\,\mathrm{ms},1.5\,\mathrm{ms}]\).}
  \label{fig:rde_error_histories}
\end{figure}

The corresponding relative error histories are shown in Figure~\ref{fig:rde_error_histories}. This figure provides the clearest quantitative summary of the RDE comparison. Across all four reported variables, the iSVD adaptive ROM maintains lower error than the Direct adaptive ROM over the entire transient phase of the simulation. The separation is especially clear in density and pressure, where the iSVD curves remain consistently below the Direct curves from early times all the way through the final reported time. In velocity and temperature, the two methods are closer and both exhibit more oscillatory behavior, but the iSVD update still maintains the lower overall error level.

\begin{figure}
\centering
\includegraphics[width=0.8\linewidth]{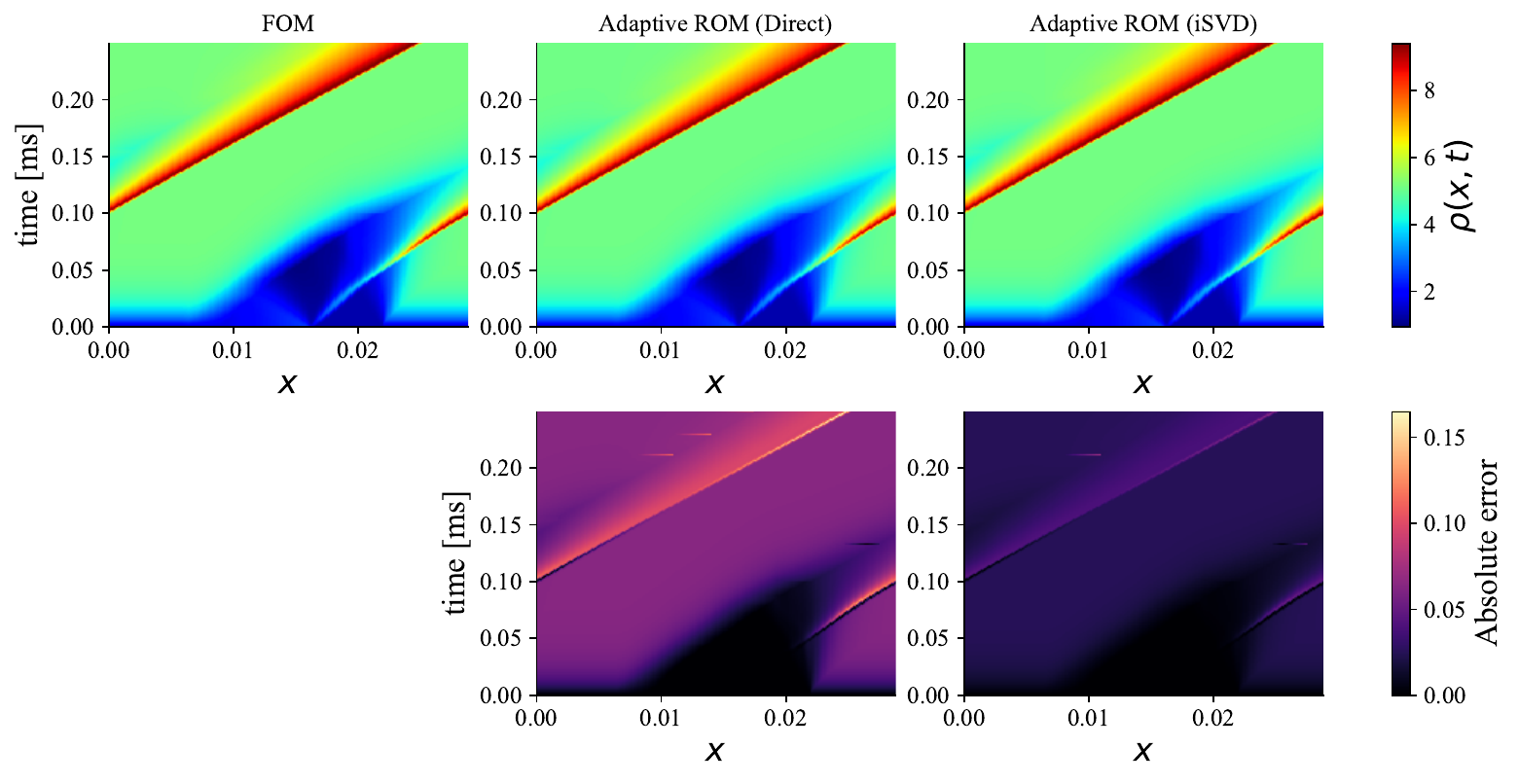}
\caption{Spacetime diagram for density in the RDE case. Both ROMs are trained on the interval \(t\in[0,5\times10^{-6}\,\mathrm{ms}]\), and tested over \(t\in[5\times10^{-6}\,\mathrm{ms},1.5\,\mathrm{ms}]\). The time axis is restricted to the transient regime before the solution settles into approximately cyclic behavior. Bottom row shows the corresponding absolute error.}
\label{fig:rde_spacetime_density}
\end{figure}

\begin{figure}
\centering
\includegraphics[width=0.8\linewidth]{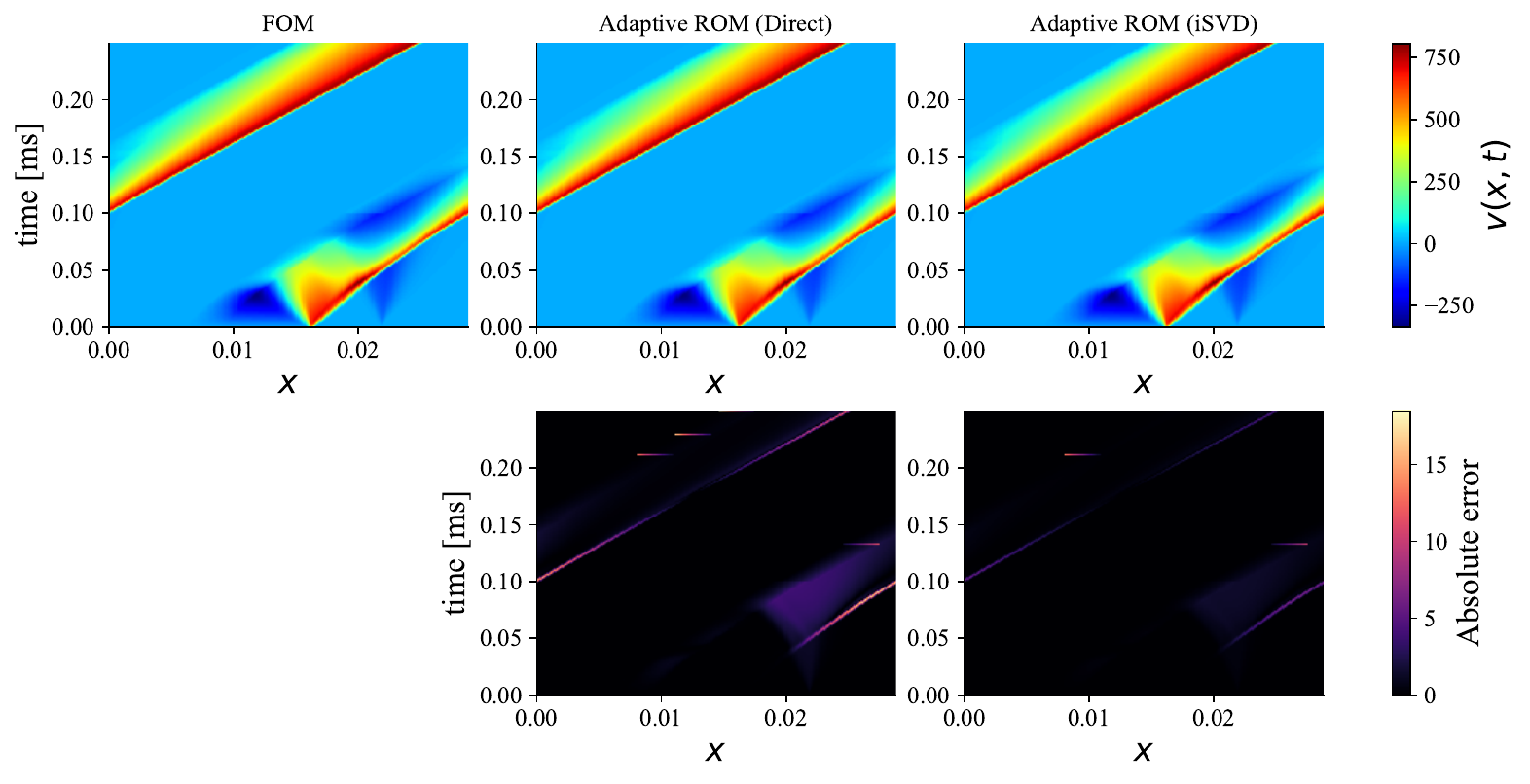}
\caption{Spacetime diagram for velocity in the RDE case. Both ROMs are trained on the interval \(t\in[0,5\times10^{-6}\,\mathrm{ms}]\), and tested over \(t\in[5\times10^{-6}\,\mathrm{ms},1.5\,\mathrm{ms}]\). The time axis is restricted to the transient regime before the solution settles into approximately cyclic behavior. Bottom row shows the corresponding absolute error.}
\label{fig:rde_spacetime_velocity}
\end{figure}

\begin{figure}
\centering
\includegraphics[width=0.8\linewidth]{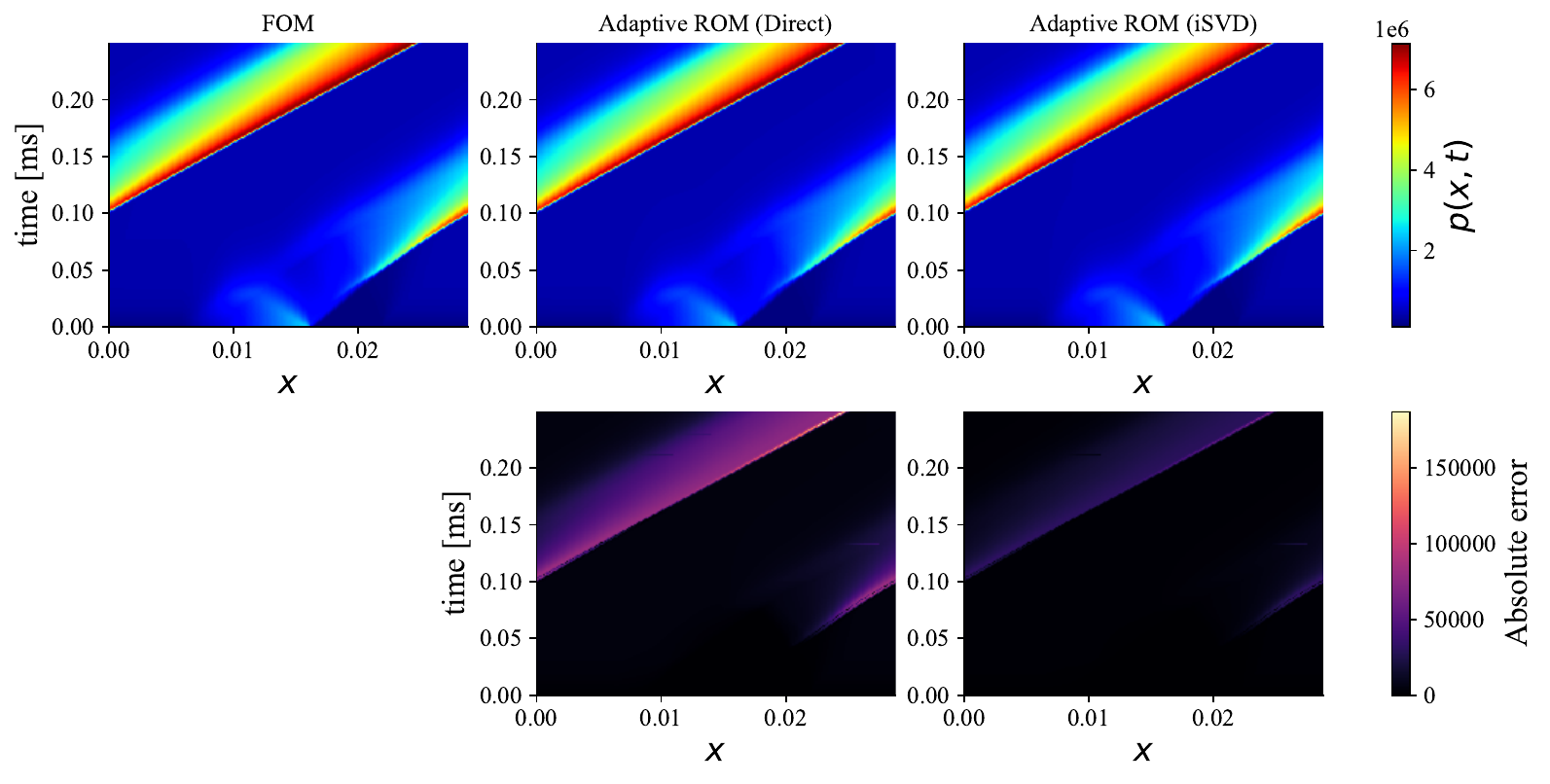}
\caption{Spacetime diagram for pressure in the RDE case. Both ROMs are trained on the interval \(t\in[0,5\times10^{-6}\,\mathrm{ms}]\), and tested over \(t\in[5\times10^{-6}\,\mathrm{ms},1.5\,\mathrm{ms}]\). The time axis is restricted to the transient regime before the solution settles into approximately cyclic behavior. Bottom row shows the corresponding absolute error.}
\label{fig:rde_spacetime_pressure}
\end{figure}

\begin{figure}
\centering
\includegraphics[width=0.8\linewidth]{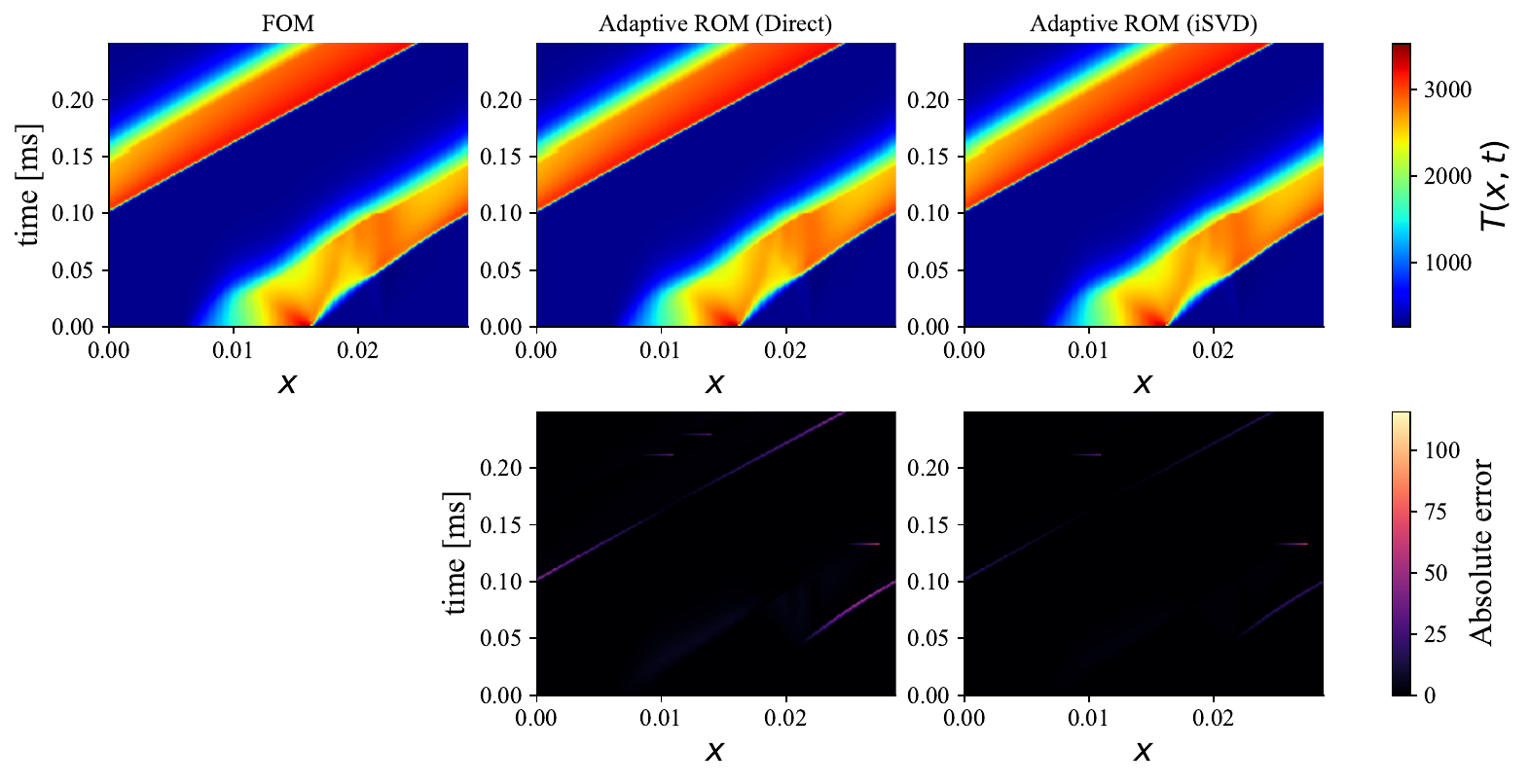}
\caption{Spacetime diagram for temperature in the RDE case. Both ROMs are trained on the interval \(t\in[0,5\times10^{-6}\,\mathrm{ms}]\), and tested over \(t\in[5\times10^{-6}\,\mathrm{ms},1.5\,\mathrm{ms}]\). The time axis is restricted to the transient regime before the solution settles into approximately cyclic behavior. Bottom row shows the corresponding absolute error.}
\label{fig:rde_spacetime_temperature}
\end{figure}

To further visualize the behavior of the adaptive ROMs, Figures~\ref{fig:rde_spacetime_density}--\ref{fig:rde_spacetime_temperature} show spacetime diagrams for density, velocity, pressure, and temperature. In each case, the top row includes the spacetime fields, while the bottom row shows the corresponding error fields. The time axis is restricted to the transient portion of the simulation, since beyond this interval the solution becomes approximately cyclic. These plots provide a complementary view of the RDE results. Visually, both adaptive ROMs appear to reproduce the dominant spacetime structures reasonably well, but the associated error fields reveal that the iSVD-based ROM maintains smaller errors throughout the transient evolution. This is especially evident in the density and pressure fields.

This result is particularly significant because the RDE case is the most difficult example considered in the paper. The advantage of iSVD is therefore not confined to the simpler Burgers or Sod settings; it persists in a chemically reactive, shock-dominated detonation problem with repeated wave propagation. Moreover, the advantage is not isolated to a single quantity of interest. The proposed history-aware update improves the reduced-order prediction in density, velocity, pressure, and temperature simultaneously. This strongly supports the main idea of iSVD, that carrying a compressed memory of past dynamics through the singular structure leads to a more reliable adaptive basis update than fitting the basis only to a recent data window. Additional RDE results over the entire prediction horizon (the full $3,000,000$ steps) are reported in Appendix~\ref{app:app_rde_additional_results}. There, we show solution profiles and relative error histories for all monitored quantities, namely density, velocity, pressure, temperature, and the mass fractions of the 10 chemical species, for the two adaptation windows \(z=5\) and \(z=10\).

\subsubsection{Runtime and acceleration}

In addition to accuracy, the RDE case provides a meaningful test of computational efficiency. Because this is the most demanding example in the paper, it is also the most compelling place to report wall-clock performance. Using the acceleration factor previously defined in \eqref{eq:acceleration_factor}, we compare the runtime of the FOM, the Direct adaptive ROM, and the proposed iSVD adaptive ROM over the same prediction horizon (all with $z=10$). Table~\ref{tab:rde_timing} reports the measured wall-clock times and the corresponding acceleration factors. The results show that the Direct adaptive ROM achieves an acceleration factor of approximately \(5.8\), while the proposed iSVD adaptive ROM scores \(10.7\). This reveals an important finding: the iSVD-based ROM is not only more accurate than the Direct adaptive ROM baseline, but also substantially faster. A major reason for this acceleration advantage is the different update frequency used by the two adaptive strategies in this RDE setting. Following the reference Direct adaptive ROM implementation, the Direct method updates the basis at every time step, while the sampling points are refreshed every \(z\) time steps when new lookahead FOM information becomes available. By contrast, the proposed iSVD ROM updates both the basis and the sampling points only every \(z\) time steps. Therefore, iSVD performs significantly fewer basis updates over the full prediction horizon. In the previous Burgers and Sod cases, where the Direct method and iSVD are both updated every \(z\) time steps, iSVD was still slightly faster, but the runtime difference was modest (see Tables \ref{tab:burgers_timing} and \ref{tab:sod_timing}). In the RDE case, however, the reduced update frequency of iSVD becomes a major contributor to the nearly \(2\) times speedup over the Direct adaptive ROM, and more than an order of magnitude faster execution than the FOM. The important observation is that this reduced update frequency does not come at the expense of accuracy. On the contrary, the iSVD adaptive ROM remains more accurate than the Direct method even though it adapts the basis less often. This suggests that the compressed spectral memory carried by iSVD allows the basis to remain predictive over multiple fine time steps without requiring a new basis update at every step.

This is one of the strongest results in the manuscript. The proposed iSVD update does not improve predictive fidelity at the expense of increased online cost; rather, on this problem, it improves both accuracy and runtime simultaneously. Since the RDE case is the most application-driven and computationally intensive example considered in the paper, this result provides strong evidence that the proposed history-aware adaptation is not only mathematically appealing, but also practically advantageous.

\begin{table}
\centering
\caption{Wall-clock timings and acceleration factors for the RDE case (with $z=10$). The experiments ware performed on University of Kansas Center for Research Computing on one compute node equipped with a Dual Intel Xeon 6240R CPU with 192GB and 2933 MHz DDR memory.}
\label{tab:rde_timing}
\begin{tabular}{lcc}
\toprule
Model & Wall-clock time (hours) & Acceleration factor \(\lambda_{\mathrm{acc}}\) \\
\midrule
FOM & $22.00$ & \(1.0\) \\
Adaptive ROM (Direct) & $3.79$ & \(5.8\) \\
Adaptive ROM (iSVD) & $2.05$ & \(10.7\) \\
\bottomrule
\end{tabular}
\end{table}

\subsubsection{Parametric study}

\begin{figure}
\centering
\includegraphics[width=0.8\linewidth]{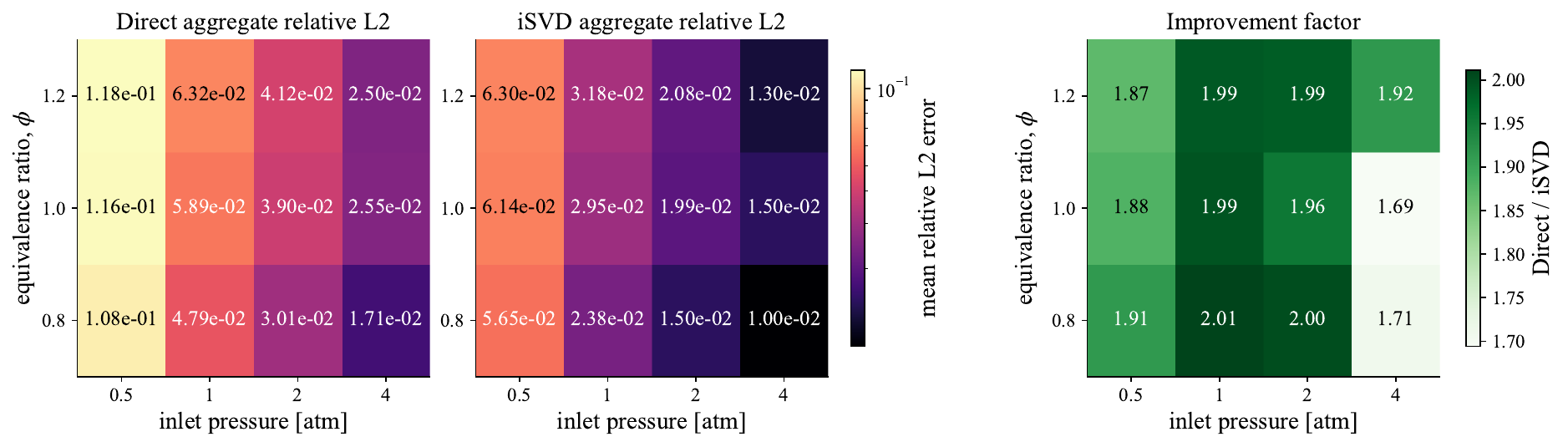}
\caption{Parametric comparison of the Direct and iSVD adaptive ROMs over equivalence ratios $\phi=\{0.8,1.0,1.2\}$ and inlet pressures $P_\mathrm{in}=\{0.5,1.0,2.0,4.0\}$ atm. Both ROMs across all cases are trained on the interval \(t\in[0,5\times10^{-6}\,\mathrm{ms}]\), and tested over \(t\in[5\times10^{-6}\,\mathrm{ms},1.5\,\mathrm{ms}]\). The first two panels show the aggregate relative $L_2$ error for the Direct and iSVD methods, respectively. The third panel shows the improvement factor, defined as the ratio of Direct error to iSVD error; values greater than unity indicate lower error for iSVD.
}
\label{fig:parametric_overall}
\end{figure}

We now perform a parametric study over three equivalence ratios, $\phi=\{0.8,1.0,1.2\}$, and four inlet pressures, $P_{\mathrm{in}}=\{0.5,1.0,2.0,4.0\}$ atm, resulting in twelve operating conditions. For each parametric configuration $\mu=(\phi,P_{\mathrm{in}})$, the Adaptive ROM solutions are compared against the corresponding FOM solution for the same operating condition. For variable $q_k$, ROM model $\in \{\mathrm{Direct},\mathrm{iSVD}\}$, and saved time index $t^n$, the instantaneous relative error is defined as
\begin{equation}
e_{k,\mathrm{ROM}}^{\mu}(t^n)
\coloneq
\frac{
\left\| q_{k,\mathrm{ROM}}^{\mu}(t^n)
-
q_{k,\mathrm{FOM}}^{\mu}(t^n)
\right\|_2
}{
\left\| q_{k,\mathrm{FOM}}^{\mu}(t^n) \right\|_2
},
\end{equation}
with the corresponding time-averaged error
\begin{equation}
\bar{e}_{k,\mathrm{ROM}}^{\mu}
\coloneq
\frac{1}{N_t}
\sum_{n=1}^{N_t}
e_{k,\mathrm{ROM}}^{\mu}(t^n).
\end{equation}
The aggregate error shown in Fig.~\ref{fig:parametric_overall} is computed by averaging the errors over all variables,
\begin{equation}
E_\mathrm{ROM}^{\mu}
\coloneq
\frac{1}{N_v}
\sum_{k=1}^{N_v}
\bar{e}_{k,\mathrm{ROM}}^{\mu},
\end{equation}
and the overall improvement factor is then defined as
$
\frac{
E_{\mathrm{Direct}}^{\mu}
}{
E_{\mathrm{iSVD}}^{\mu}
}.
$
As shown in Fig.~\ref{fig:parametric_overall}, the iSVD method yields an improvement factor greater than unity for all equivalence ratios and inlet pressures considered, indicating consistently lower error than the Direct method across the full parametric sweep.

\begin{figure}
\centering
\includegraphics[width=0.8\linewidth]{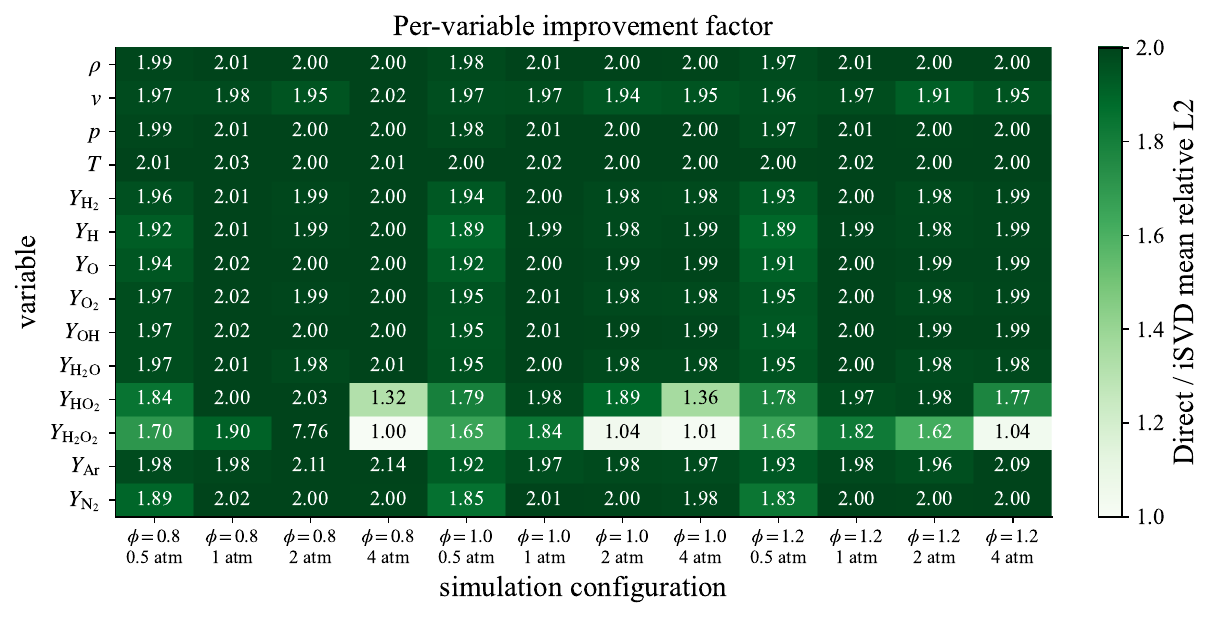}
\caption{Per-variable improvement factor for the iSVD adaptive ROM relative to the Direct adaptive ROM across the full parametric sweep. Both ROMs across all cases are trained on the interval \(t\in[0,5\times10^{-6}\,\mathrm{ms}]\), and tested over \(t\in[5\times10^{-6}\,\mathrm{ms},1.5\,\mathrm{ms}]\).}
\label{fig:parametric_variable}
\end{figure}

Fig.~\ref{fig:parametric_variable} further supports this trend, where the per-variable improvement factor
$
\frac{
\bar{e}_{k,\mathrm{Direct}}^{\mu}
}{
\bar{e}_{k,\mathrm{iSVD}}^{\mu}
}
$
is greater than unity for all variables and parameter cases. This demonstrates that the improved accuracy of iSVD is not limited to a single flow quantity, but is observed across the primitive variables and species mass fractions. The acceleration factors remain the same as previously reported and are therefore omitted here. Consequently, across all tested operating conditions, iSVD is not only more accurate than the Direct method, but also nearly twice as fast.

\section{Conclusion}
\label{sec:conclusion}

This work introduced a history-aware adaptive reduced-order modeling framework based on incremental singular value decomposition. The central idea is to couple an intrusive projection-based ROM with an online subspace tracking mechanism that periodically incorporates coarse-time-step full-order information and updates the reduced basis accordingly. This leads to an adaptive framework that evolves with the dynamics during prediction.

A key theme of the paper has been that successful adaptive reduced-order modeling is not only a matter of updating the basis online, but of doing so in a way that preserves useful information about the system's past. In the proposed framework, this role is played by iSVD. By carrying forward singular values associated with previously learned subspace, the update retains a compressed representation of the historical dynamics while assimilating new correction snapshots. This provides a principled mechanism for balancing memory and responsiveness, and gives a concrete mathematical meaning to the notion of a history-aware adaptive ROM.

The numerical experiments support this perspective consistently across all three test cases. We used the viscous Burgers problem as a development platform to study the predictive limitations of static ROMs, the benefits of online adaptation, and the sensitivity of the framework to its main hyperparameters. These results showed that static ROMs, even when enriched by larger reduced dimensions or longer offline training windows, remain fundamentally limited when asked to extrapolate far beyond the training regime. In contrast, the proposed adaptive framework substantially improves predictive accuracy. Among the basis adaptation methods tested, the methods that preserve some notion of the past outperform those that react only to the newest observation, and iSVD provides the strongest overall performance while maintaining the lightest compressed memory of system history among the history-aware approaches. The Burgers results also clarified an important conceptual point, that the coarse full-order correction signal need not be more accurate than the adaptive ROM itself in order to be useful. Its value lies in providing off-manifold information that guides the basis update, not in serving as a replacement for the reduced state.

The Sod shock tube extended these conclusions to a compressible hyperbolic setting with moving discontinuities. In this problem, the static ROM again failed once the dominant wave structures moved beyond the initial trial manifold, while the adaptive ROMs remained predictive. Most importantly, the proposed iSVD adaptive ROM consistently improved upon the Direct update in density, velocity, and pressure across the adaptation windows studied. These results showed that the advantages of the history-aware iSVD update are not limited to smooth model problems, but persist in a setting governed by shock propagation and sharp wave interactions.

The one-dimensional premixed rotating detonation engine provided the strongest and most application-driven validation case. In this case, the ROM is initialized from only \(5\,\mathrm{ns}\) of training data and then used to predict over a \(1.5\,\mathrm{ms}\) horizon, making the prediction interval approximately \(3\times 10^5\) times longer than the initial training window. In this highly stiff, chemically reactive, shock-dominated problem, the proposed iSVD adaptive ROM improved upon the current Direct adaptive ROM baseline in all reported state variables and also delivered a significantly larger acceleration factor. This final result is particularly important, as it shows that the proposed update is not only more accurate, but also more efficient in a demanding practical setting. Thus, the benefit of iSVD in adaptive reduced-order modeling is not merely theoretical; it translates directly into improved predictive fidelity and computational performance on a problem of real relevance.

Taken together, these results suggest that iSVD provides a particularly effective mechanism for online basis adaptation in predictive reduced-order modeling. It preserves a compressed history of the evolving dynamics, requires only small reduced-order linear algebra in the update, and integrates naturally into projection-based ROM frameworks. More broadly, the present work shows that adaptive ROMs need not rely on large offline datasets or expensive online basis reconstruction to remain predictive. A minimal offline phase, combined with principled online learning of the reduced subspace, can be sufficient to produce accurate and efficient reduced models well beyond their original training regime.

Several directions remain for future work. One important path is the development of adaptive strategies for choosing the correction frequency and the forgetting factor automatically, based on the evolving state of the simulation. It would also be of interest to study more general correction signal sources, partial-observation updates, and higher-dimensional multi-physics applications. Another path forward is extending the framework to non-intrusive ROMs, where adaptation must update not only the basis but also the learned reduced operators. Nevertheless, the present results already indicate that history-aware subspace tracking, and in particular iSVD, offers a promising path toward more reliable adaptive ROMs for predictive simulation.



\section*{Code Availability}
The Python implementation of the proposed framework, including all code needed to reproduce the 1D viscous Burgers and Sod shock tube experiments, is openly available at \url{https://github.com/APHedayat/iSVD-ROM} under the MIT license. The RDE simulations were performed using the in-house CFD code available at \url{https://github.com/alimike97/CompFlowLab}.

\section*{Acknowledgment}
Amirpasha Hedayat and Karthik Duraisamy were supported by the Office of Under Secretary of Defense for Research and Engineering (OUSD(RE)) grant N00014-21-1-295. Cheng Huang and Ali Mohaghegh acknowledge the supports from the Air Force Office of Scientific Research (AFOSR) under the grant FA9550-23-1-0211 (Program managers: Drs. Chiping Li and Fariba Fahroo) and support from the Center for Research Computing at the University of Kansas. Laura Balzano acknowledges NSF award CCF-2331590.

\appendix
\counterwithin{figure}{section}
\counterwithin{table}{section}
\counterwithin{equation}{section}
\counterwithin{algorithm}{section}

\section{Algorithms for alternative basis adaptation methods}
\label{app:app_basis_update_algorithms}

In this appendix, we collect the basis update algorithms used for comparison with iSVD. Throughout, \(\widehat{\bm y}_{k+1} \in \mathbb{R}^N\) denotes the preprocessed correction snapshot,
\[
\widehat{\bm y}_{k+1}
\coloneq
\bm D\left(\bm y_{k+1}-\bm q_{\mathrm{ref}}\right),
\]
and \(\bm \Phi_k \in \mathbb{R}^{N\times r}\) denotes the current orthonormal basis. Note also that we distinguish between the fine-time-step index $n$ and the adaptation event index $k$.

\begin{algorithm}
\caption{Windowed SVD update}
\label{alg:wsvd_update}
\begin{algorithmic}[1]
\State \textbf{Input:} preprocessed windowed data matrix \(\widehat{\bm Y}^{(w)}_{k+1} \in \mathbb{R}^{N\times w}\), target rank \(r\)

\State Compute the singular value decomposition
$$
\widehat{\bm Y}^{(w)}_{k+1}
=
\bm U^{(w)}_{k+1} \bm \Sigma^{(w)}_{k+1} \bm V^{(w)T}_{k+1}
$$
\State Retain the leading \(r\) left singular vectors:
$$
\bm \Phi_{k+1}
\coloneq
\bm U^{(w)}_{k+1}(:,1\!:\!r)
$$
\State \textbf{Return:} updated basis \(\bm \Phi_{k+1}\)
\end{algorithmic}
\end{algorithm}

\begin{algorithm}
\caption{Direct update}
\label{alg:direct_update}
\begin{algorithmic}[1]
\State \textbf{Input:} current basis \(\bm \Phi_k \in \mathbb{R}^{N\times r}\), preprocessed windowed data matrix \(\widehat{\bm Y}^{(w)}_{k+1} \in \mathbb{R}^{N\times w}\)

\State Form the reduced data matrix
$$
\bm Z_{k+1}
\coloneq
\bm \Phi_k^\dagger \widehat{\bm Y}^{(w)}_{k+1}
$$
\State Compute the updated basis
$$
\widetilde{\bm \Phi}_{k+1}
\coloneq
\widehat{\bm Y}^{(w)}_{k+1}\bm Z_{k+1}^\dagger
$$
\State Orthonormalize the updated basis:
$$
\bm \Phi_{k+1}
\coloneq
\operatorname{orth}\!\left(\widetilde{\bm \Phi}_{k+1}\right)
$$
\State \textbf{Return:} updated basis \(\bm \Phi_{k+1}\)
\end{algorithmic}
\end{algorithm}

\begin{algorithm}
\caption{One-step (rank-one) update}
\label{alg:onestep_update}
\begin{algorithmic}[1]
\State \textbf{Input:} current basis \(\bm \Phi_k \in \mathbb{R}^{N\times r}\), previous preprocessed correction snapshot \(\widehat{\bm y}_{k} \in \mathbb{R}^N\), current reduced coordinates \(\bm a_{n+1}^{-} \in \mathbb{R}^r\)

\State Form the residual between the new snapshot and the current reduced reconstruction:
$$
\bm r_{k+1}
\coloneq
\widehat{\bm y}_{k}
-
\bm \Phi_k \bm a_{n+1}^{-}
$$
\State Form the rank-one correction:
$$
\Delta \bm \Phi_{k+1}
\coloneq
\frac{
\bm r_{k+1}(\bm a_{n+1}^{-})^T
}{
\|\bm a_{n+1}^{-}\|_2^2
}
$$
\State Update and orthonormalize the basis:
$$
\bm \Phi_{k+1}
\coloneq
\operatorname{orth}\!\left(
\bm \Phi_k + \Delta \bm \Phi_{k+1}
\right)
$$
\State \textbf{Return:} updated basis \(\bm \Phi_{k+1}\)
\end{algorithmic}
\end{algorithm}

\begin{algorithm}
\caption{Oja's update}
\label{alg:oja_update}
\begin{algorithmic}[1]
\State \textbf{Input:} current basis \(\bm \Phi_k \in \mathbb{R}^{N\times r}\), previous preprocessed correction snapshot \(\widehat{\bm y}_{k} \in \mathbb{R}^N\), learning rate \(\eta > 0\)

\State Form the streaming update:
$$
\widetilde{\bm \Phi}_{k+1}
\coloneq
\bm \Phi_k
+
\eta \,
\widehat{\bm y}_{k}
\bigl(
\widehat{\bm y}_{k}^T \bm \Phi_k
\bigr)
$$
\State Orthonormalize the updated basis:
$$
\bm \Phi_{k+1}
\coloneq
\operatorname{orth}\!\left(\widetilde{\bm \Phi}_{k+1}\right)
$$
\State \textbf{Return:} updated basis \(\bm \Phi_{k+1}\)
\end{algorithmic}
\end{algorithm}

\begin{algorithm}
\caption{GROUSE update}
\label{alg:grouse_update}
\begin{algorithmic}[1]
\State \textbf{Input:} current basis \(\bm \Phi_k \in \mathbb{R}^{N\times r}\), previous preprocessed correction snapshot \(\widehat{\bm y}_{k} \in \mathbb{R}^N\), learning rate \(\eta > 0\)

\State Compute the reduced coefficients
$$
\bm w_{k+1}
\coloneq
\arg\min_{\bm w\in\mathbb{R}^r}
\left\|
\widehat{\bm y}_{k}
-
\bm \Phi_k \bm w
\right\|_2^2
$$
\State Form the projected component and residual:
$$
\bm p_{k+1}
\coloneq
\bm \Phi_k \bm w_{k+1},
\qquad
\bm r_{k+1}
\coloneq
\widehat{\bm y}_{k}
-
\bm p_{k+1}
$$
\State Compute the geodesic step size
$$
\alpha_{k+1}
\coloneq
\eta \|\bm p_{k+1}\|_2 \|\bm r_{k+1}\|_2
$$
\State Form the update direction
$$
\bm d_{k+1}
\coloneq
\left[
\bigl(\cos\alpha_{k+1}-1\bigr)\frac{\bm p_{k+1}}{\|\bm p_{k+1}\|_2}
+
\sin\alpha_{k+1}\frac{\bm r_{k+1}}{\|\bm r_{k+1}\|_2}
\right]
\frac{\bm w_{k+1}^T}{\|\bm w_{k+1}\|_2}
$$
\State Update the basis:
$$
\bm \Phi_{k+1}
\coloneq
\bm \Phi_k + \bm d_{k+1}
$$
\State \textbf{Return:} updated basis \(\bm \Phi_{k+1}\)
\end{algorithmic}
\end{algorithm}

\section{Governing equations for Sod shock tube}
\label{app:app_sod}

For the Sod shock tube considered in Section~\ref{subsec:results_sod}, the FOM is the one-dimensional compressible Euler system written in conservative form,
\begin{equation}
\frac{\partial \bm U}{\partial t}
+
\frac{\partial \bm F(\bm U)}{\partial x}
=
\bm 0,
\qquad
x \in [0,1],
\label{eq:sod_euler_appendix}
\end{equation}
with conserved state
\begin{equation}
\bm U(x,t)
=
\begin{bmatrix}
\rho \\
m \\
E
\end{bmatrix},
\label{eq:sod_state_appendix}
\end{equation}
where \(\rho\) is the density, \(m=\rho u\) is the momentum, \(u\) is the velocity, and \(E\) is the total energy. The inviscid flux is
\begin{equation}
\bm F(\bm U)
=
\begin{bmatrix}
m \\
\dfrac{m^2}{\rho} + p \\
u(E+p)
\end{bmatrix},
\label{eq:sod_flux_appendix}
\end{equation}
where the pressure is given by the ideal-gas equation of state
\begin{equation}
p
=
(\gamma-1)\left(
E - \frac{1}{2}\rho u^2
\right),
\qquad
\gamma = 1.4.
\label{eq:sod_eos_appendix}
\end{equation}

The initial condition is the standard Sod Riemann problem with a discontinuity at the midpoint of the domain,
\begin{equation}
(\rho,u,p)(x,0)
=
\begin{cases}
(1.0,\,0,\,1.0), & x < 0.5, \\[1mm]
(0.125,\,0,\,0.1), & x > 0.5.
\end{cases}
\label{eq:sod_ic_appendix}
\end{equation}
Equivalently, the initial conserved state is
\begin{equation}
\bm U(x,0)
=
\begin{bmatrix}
\rho \\
\rho u \\
\dfrac{p}{\gamma-1} + \dfrac{1}{2}\rho u^2
\end{bmatrix}.
\label{eq:sod_conserved_ic_appendix}
\end{equation}

The domain is partitioned into \(N_x=256\) uniform finite-volume cells of width
\begin{equation}
\Delta x = \frac{1}{N_x},
\label{eq:sod_dx_appendix}
\end{equation}
and the unknown \(\bm U_i^n\) denotes the cell-average approximation over cell \(i\) at time level \(t^n\). The semi-discrete finite-volume form is
\begin{equation}
\frac{d\bm U_i}{dt}
+
\frac{\hat{\bm F}_{i+1/2}-\hat{\bm F}_{i-1/2}}{\Delta x}
=
\bm 0,
\qquad
i=1,\dots,N_x,
\label{eq:sod_fv_semidiscrete_appendix}
\end{equation}
where \(\hat{\bm F}_{i+1/2}\) is the numerical flux at the interface between cells \(i\) and \(i+1\).

In the present implementation, the numerical flux is a local Lax--Friedrichs (Rusanov-type) flux,
\begin{equation}
\hat{\bm F}_{i+1/2}
=
\frac{1}{2}\left[
\bm F(\bm U_i) + \bm F(\bm U_{i+1})
\right]
-
\frac{1}{2}s_{i+1/2}\left(
\bm U_{i+1}-\bm U_i
\right),
\label{eq:sod_rusanov_flux_appendix}
\end{equation}
where the interface dissipation speed is taken as
\begin{equation}
s_{i+1/2}
=
\max\!\left(
|u_i|+c_i,\,
|u_{i+1}|+c_{i+1}
\right),
\label{eq:sod_rusanov_speed_appendix}
\end{equation}
with sound speed
\begin{equation}
c = \sqrt{\frac{\gamma p}{\rho}}.
\label{eq:sod_sound_speed_appendix}
\end{equation}
At the two boundaries, ghost states are formed by copying the nearest interior state, corresponding to a zero-gradient boundary treatment in the finite-volume discretization.

Time integration is performed with backward Euler. Thus, given \(\bm U^n\), the next state \(\bm U^{n+1}\) is obtained from
\begin{equation}
\bm U^{n+1}
-
\bm U^n
+
\Delta t\,\bm R(\bm U^{n+1})
=
\bm 0,
\label{eq:sod_backward_euler_appendix}
\end{equation}
where \(\bm R(\bm U)\) denotes the finite-volume residual assembled from the flux differences in \eqref{eq:sod_fv_semidiscrete_appendix}. In the numerical experiments of Section~\ref{subsec:results_sod}, the time step is
\begin{equation}
\Delta t = 2.5\times 10^{-4},
\label{eq:sod_dt_appendix}
\end{equation}
and the solution is advanced for \(N_t=500\) time steps.

The nonlinear system \eqref{eq:sod_backward_euler_appendix} is solved by Newton iterations. At each Newton step, primitive variables are recovered from the current conservative state through
\begin{equation}
u = \frac{m}{\rho},
\qquad
p = (\gamma-1)\left(E-\frac{1}{2}\rho u^2\right),
\label{eq:sod_primitives_appendix}
\end{equation}
and small positive floors are enforced on density and pressure in the implementation to avoid nonphysical states during the nonlinear iterations. The Jacobian used in Newton's method is assembled from the discrete residual of the backward-Euler finite-volume system together with the Jacobians of the Euler fluxes.

This fully discrete finite-volume/backward-Euler solver defines the full-order model used to generate the reference trajectories and correction signals for the Sod shock tube experiments reported in Section~\ref{subsec:results_sod}.

\section{Additional Sod shock tube results}
\label{app:app_sod_additional_results}

This appendix collects the additional Sod shock tube results for the adaptation windows
$
z\in\{5,15,20\}.
$
These figures complement the results at \(z=10\) and show that the same qualitative ordering persists across the other adaptation frequencies considered.

\begin{figure}
\centering
  \subfloat[Density profiles.]{%
    \includegraphics[width=0.75\columnwidth]{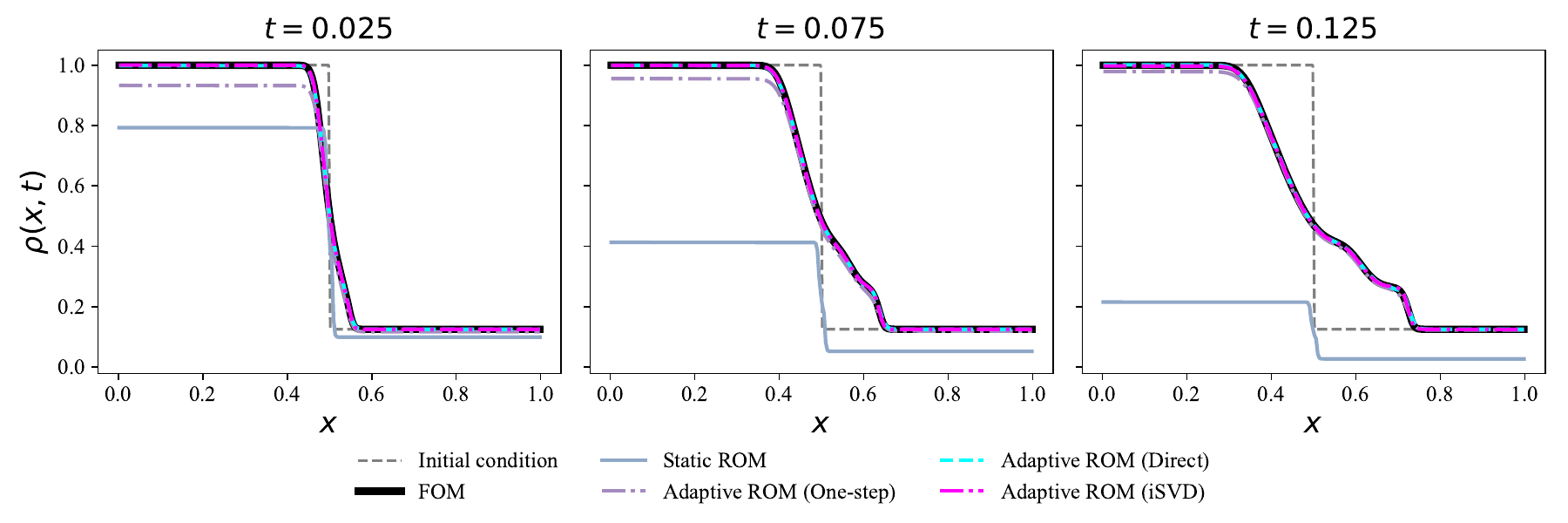}%
  }\\ \vspace{-10pt}
  \subfloat[Velocity profiles.]{%
    \includegraphics[width=0.75\columnwidth]{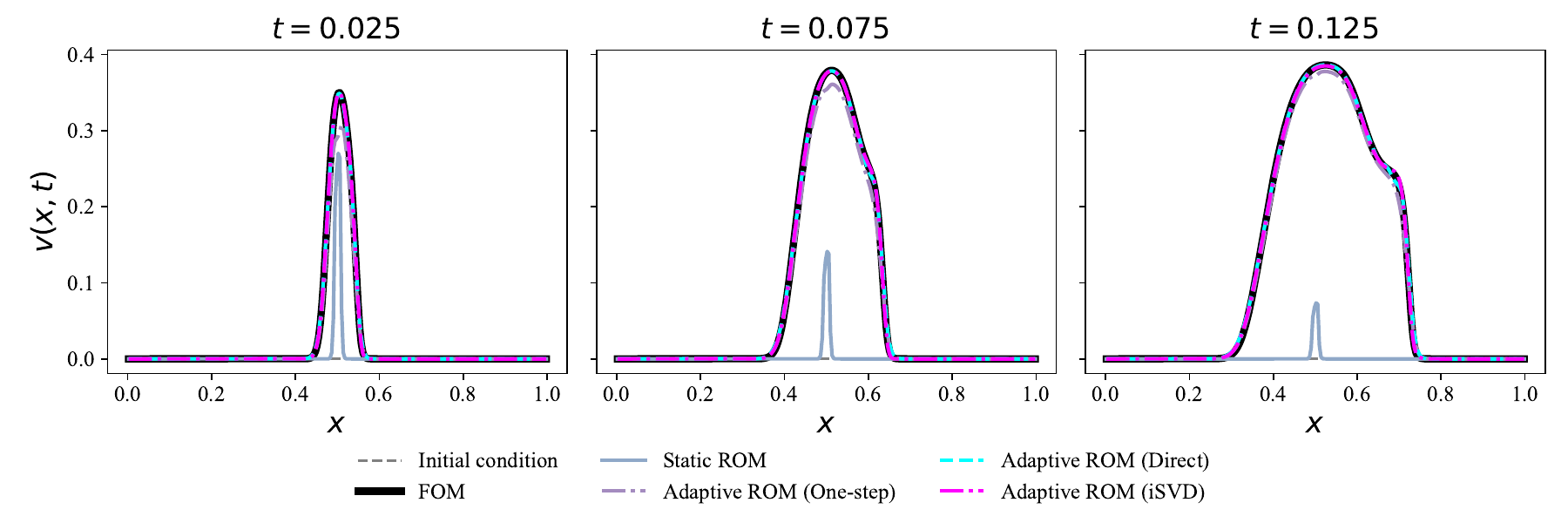}%
  }\\ \vspace{-10pt}
  \subfloat[Pressure profiles.]{%
    \includegraphics[width=0.75\columnwidth]{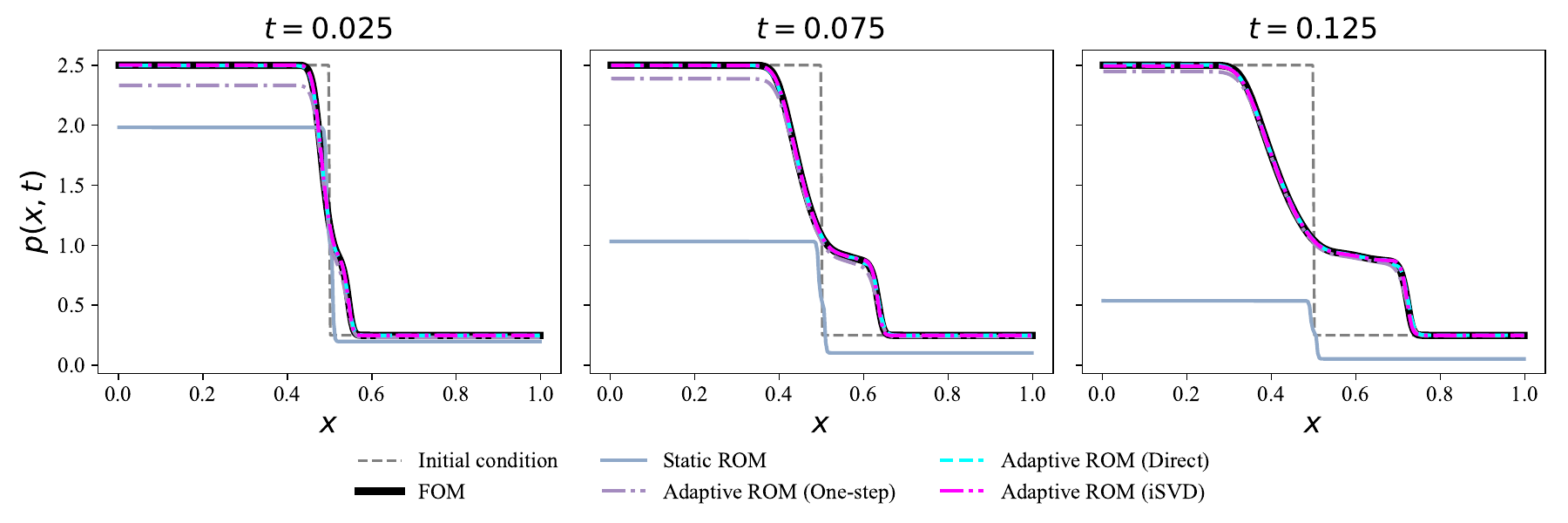}%
  }
  \caption{Sod shock tube solution profiles for \(z=5\). All ROMs are trained on the interval \(t\in[0,0.001]\), and tested over \(t\in[0.001,0.125]\).}
  \label{fig:app_sod_profiles_z5}
\end{figure}

\begin{figure}
\centering
  \subfloat[Density profiles.]{%
    \includegraphics[width=0.75\columnwidth]{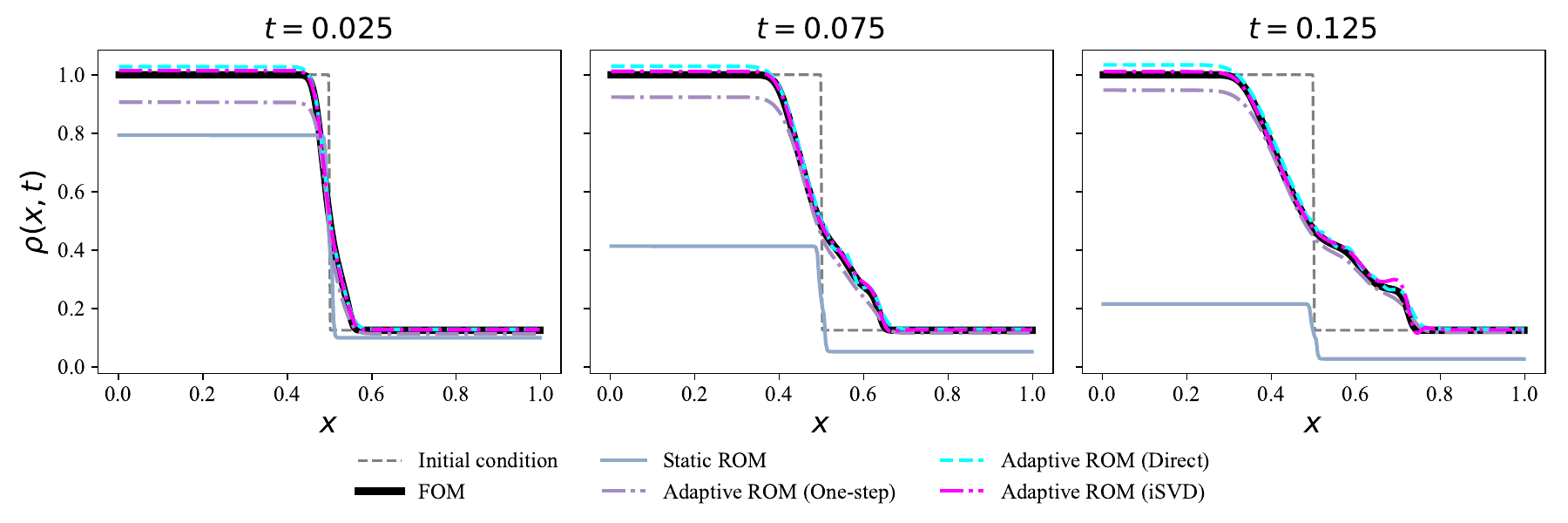}%
  }\\ \vspace{-10pt}
  \subfloat[Velocity profiles.]{%
    \includegraphics[width=0.75\columnwidth]{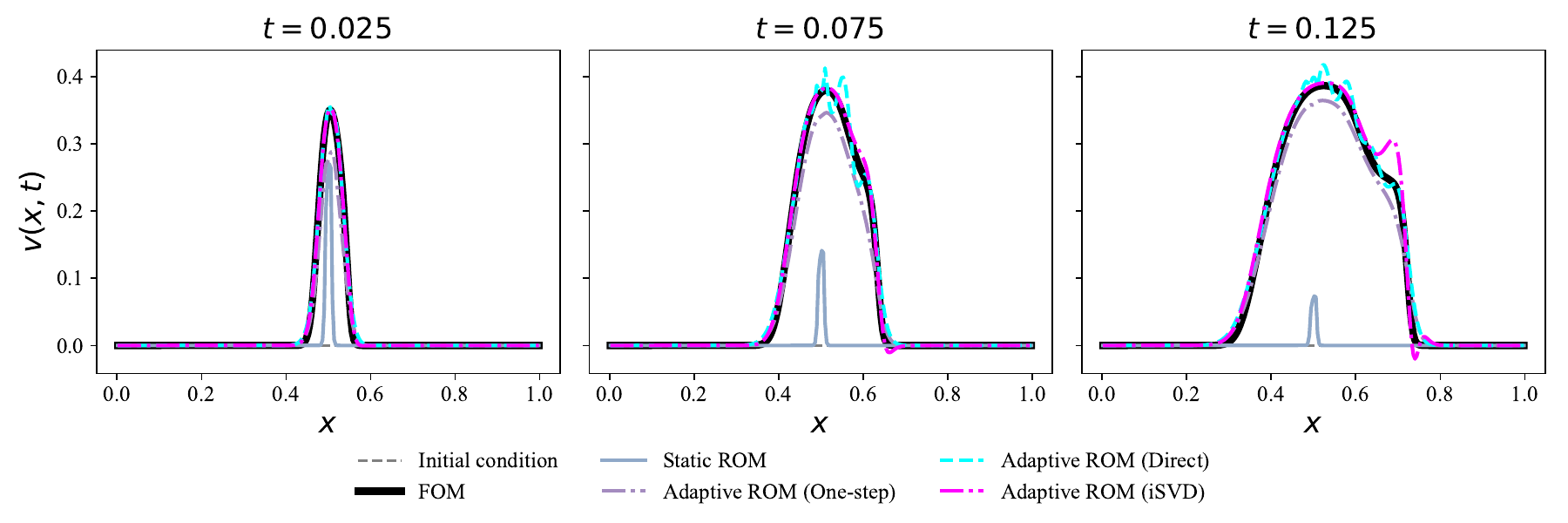}%
  }\\ \vspace{-10pt}
  \subfloat[Pressure profiles.]{%
    \includegraphics[width=0.75\columnwidth]{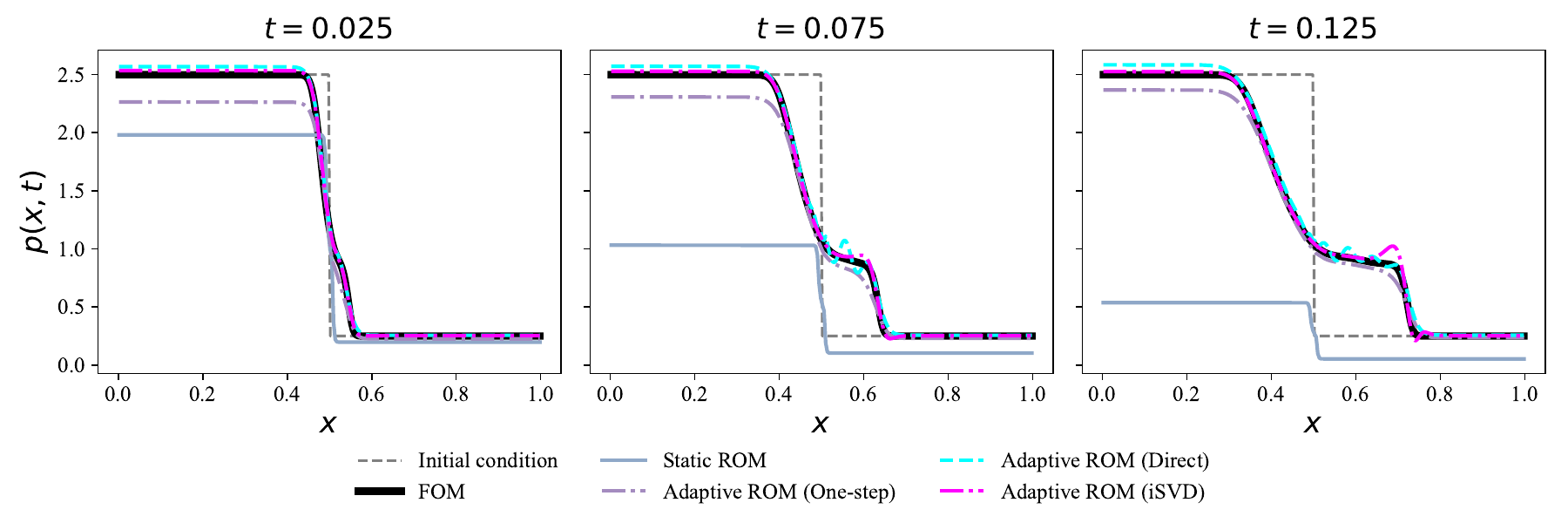}%
  }
  \caption{Sod shock tube solution profiles for \(z=15\). All ROMs are trained on the interval \(t\in[0,0.001]\), and tested over \(t\in[0.001,0.125]\).}
  \label{fig:app_sod_profiles_z15}
\end{figure}

\begin{figure}
\centering
  \subfloat[Density profiles.]{%
    \includegraphics[width=0.75\columnwidth]{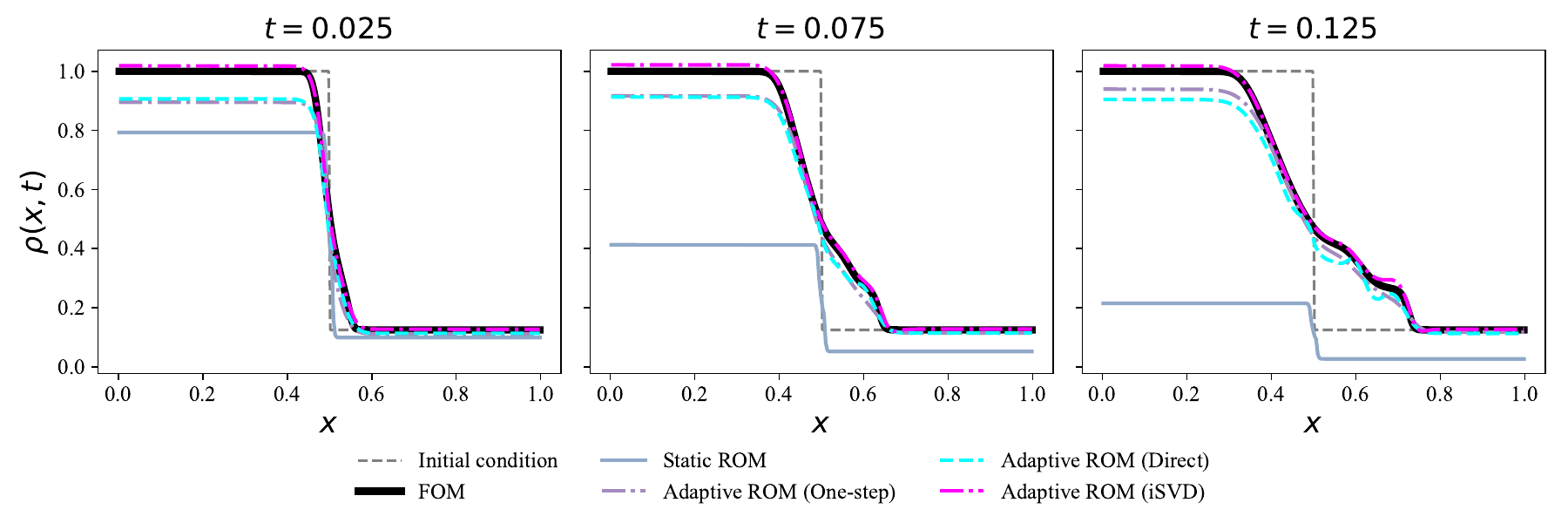}%
  }\\ \vspace{-10pt}
  \subfloat[Velocity profiles.]{%
    \includegraphics[width=0.75\columnwidth]{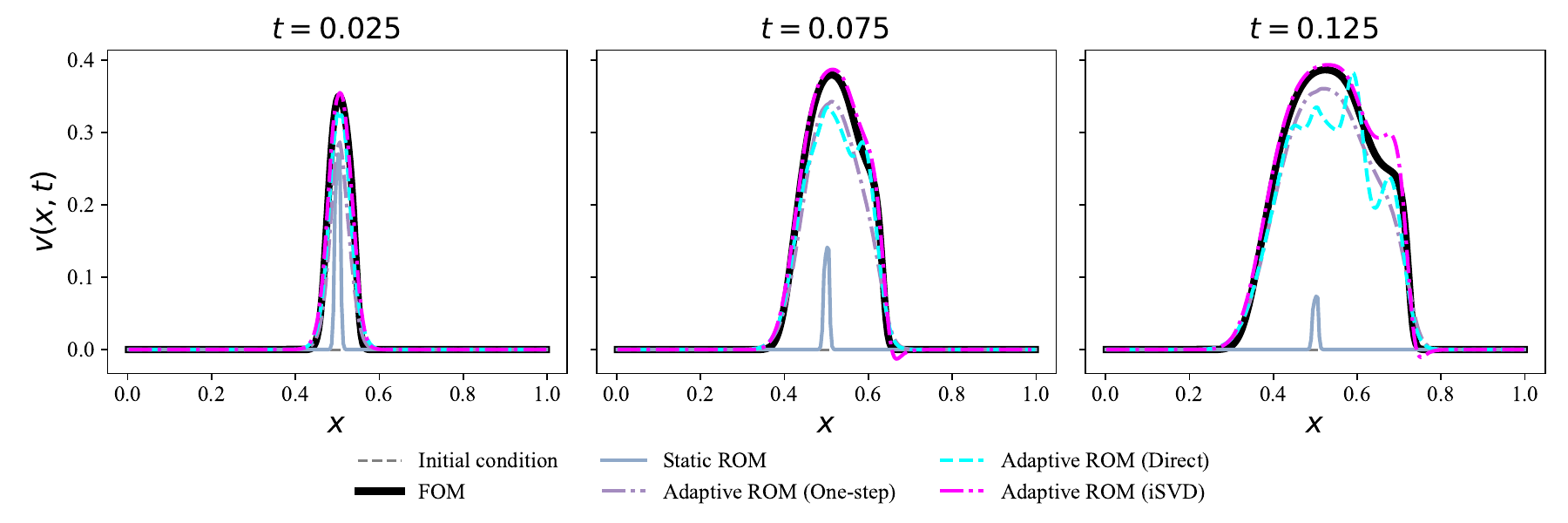}%
  }\\ \vspace{-10pt}
  \subfloat[Pressure profiles.]{%
    \includegraphics[width=0.75\columnwidth]{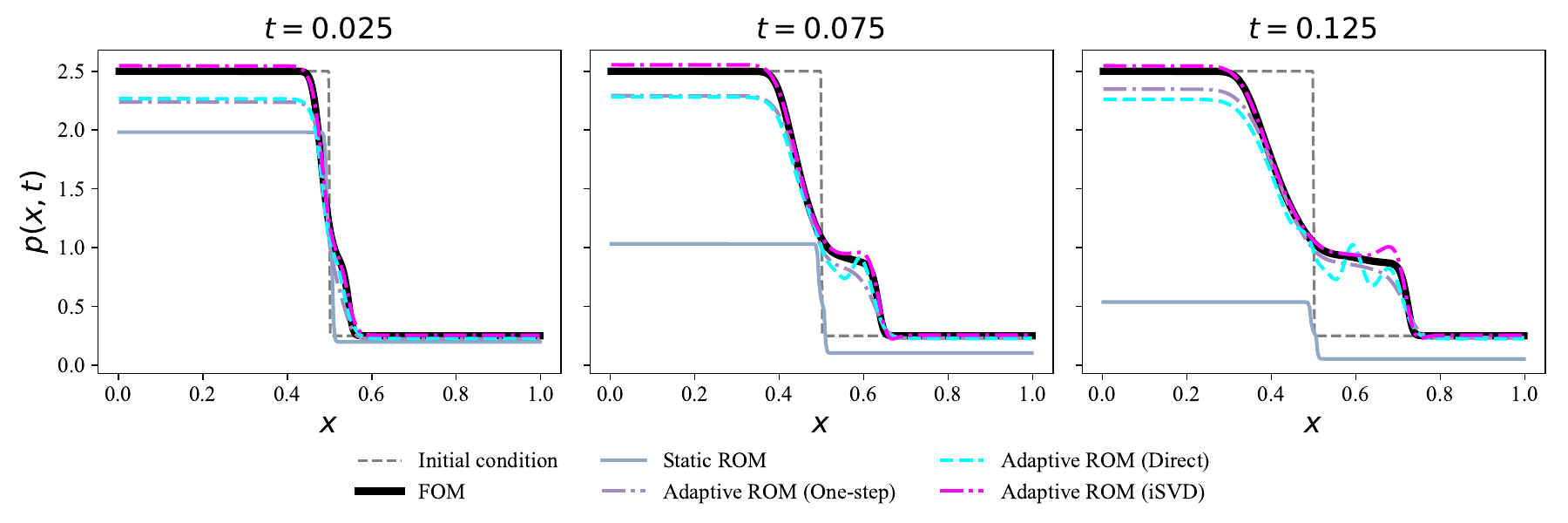}%
  }
  \caption{Sod shock tube solution profiles for \(z=20\). All ROMs are trained on the interval \(t\in[0,0.001]\), and tested over \(t\in[0.001,0.125]\).}
  \label{fig:app_sod_profiles_z20}
\end{figure}

Figures~\ref{fig:app_sod_profiles_z5}--\ref{fig:app_sod_profiles_z20} compare the density, velocity, and pressure profiles of the FOM, the static ROM, the one-step adaptive ROM, the Direct adaptive ROM, and the proposed iSVD adaptive ROM. Across all three additional adaptation windows, the static ROM again fails once the moving shock-tube structures leave the initial training manifold. The one-step adaptive ROM consistently improves over the static baseline, but remains clearly less accurate than the two history-aware methods. Both Direct and iSVD remain predictive, with iSVD generally producing the sharpest and most faithful profiles. As the adaptation window grows, the iSVD adaptive ROM continues to provide the best overall profile agreement with the FOM.

\begin{figure}
\centering
  \subfloat[Relative error in density.]{%
    \includegraphics[width=0.3\columnwidth]{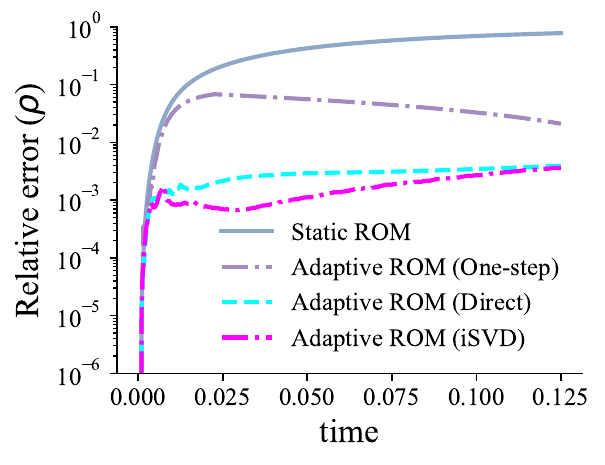}%
  }
  \subfloat[Relative error in velocity.]{%
    \includegraphics[width=0.3\columnwidth]{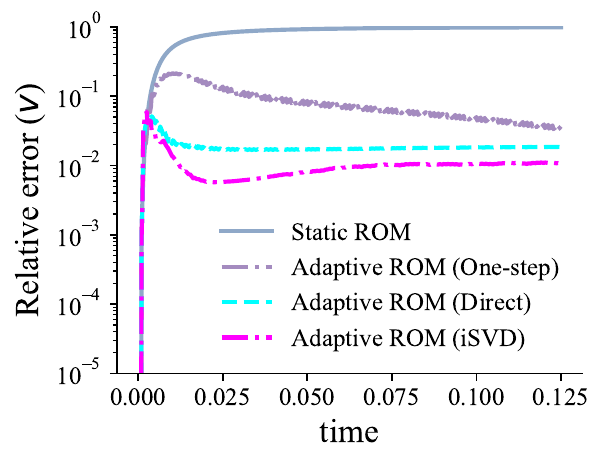}%
  }
  \subfloat[Relative error in pressure.]{%
    \includegraphics[width=0.3\columnwidth]{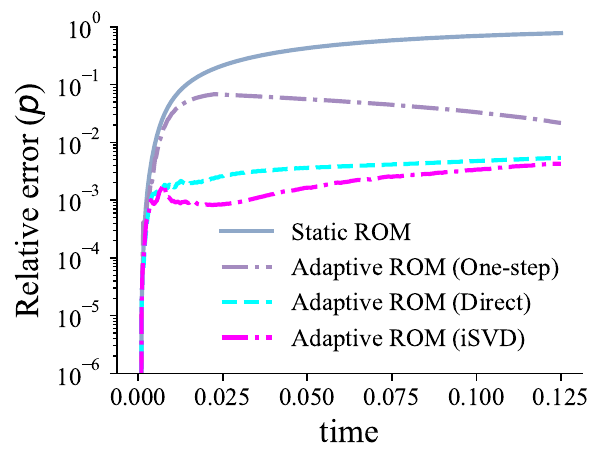}%
  }
  \caption{Relative error histories for the Sod shock tube at \(z=5\). All ROMs are trained on the interval \(t\in[0,0.001]\), and tested over \(t\in[0.001,0.125]\).}
  \label{fig:app_sod_errors_z5}
\end{figure}

\begin{figure}
\centering
  \subfloat[Relative error in density.]{%
    \includegraphics[width=0.3\columnwidth]{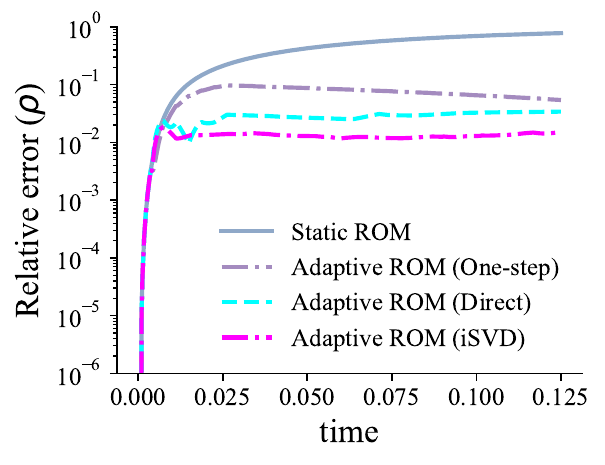}%
  }
  \subfloat[Relative error in velocity.]{%
    \includegraphics[width=0.3\columnwidth]{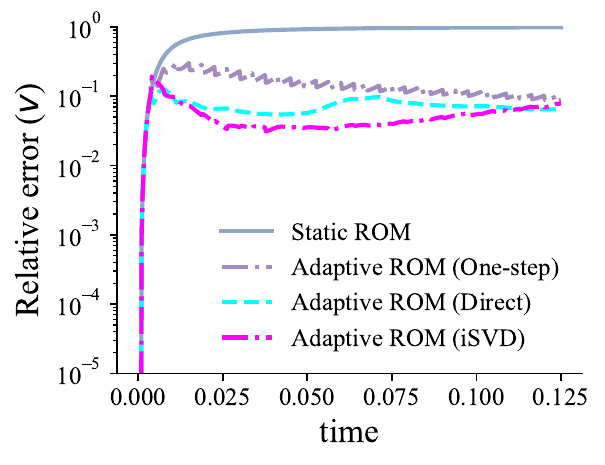}%
  }
  \subfloat[Relative error in pressure.]{%
    \includegraphics[width=0.3\columnwidth]{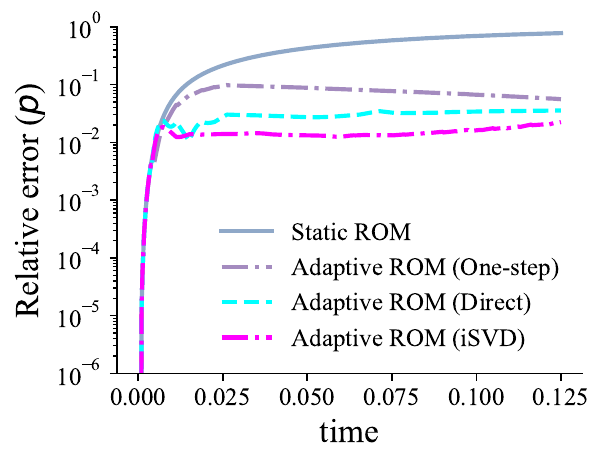}%
  }
  \caption{Relative error histories for the Sod shock tube at \(z=15\). All ROMs are trained on the interval \(t\in[0,0.001]\), and tested over \(t\in[0.001,0.125]\).}
  \label{fig:app_sod_errors_z15}
\end{figure}

\begin{figure}
\centering
  \subfloat[Relative error in density.]{%
    \includegraphics[width=0.3\columnwidth]{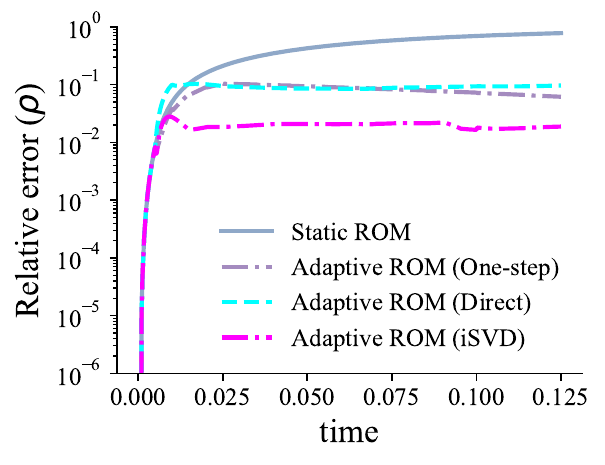}%
  }
  \subfloat[Relative error in velocity.]{%
    \includegraphics[width=0.3\columnwidth]{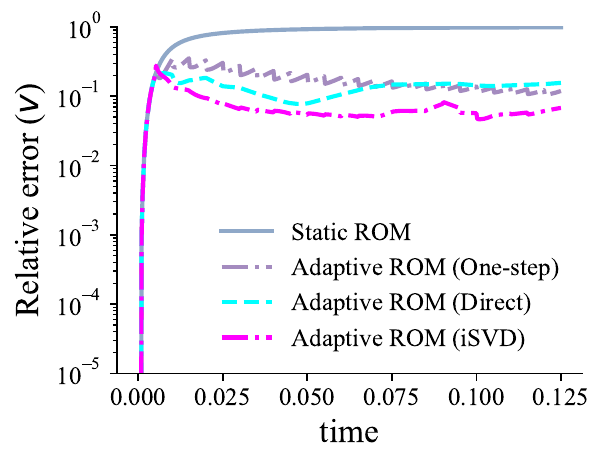}%
  }
  \subfloat[Relative error in pressure.]{%
    \includegraphics[width=0.3\columnwidth]{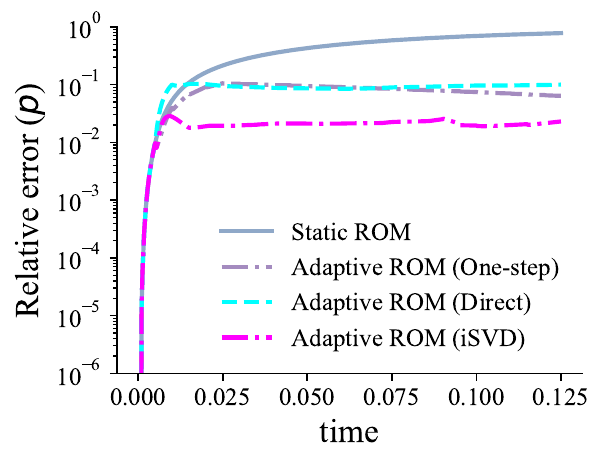}%
  }
  \caption{Relative error histories for the Sod shock tube at \(z=20\). All ROMs are trained on the interval \(t\in[0,0.001]\), and tested over \(t\in[0.001,0.125]\).}
  \label{fig:app_sod_errors_z20}
\end{figure}

The corresponding relative error histories are shown in Figures~\ref{fig:app_sod_errors_z5}--\ref{fig:app_sod_errors_z20}. The static ROM is the least accurate model, the one-step adaptive ROM lies in an intermediate position, and the two history-aware methods remain the strongest models. Among them, iSVD consistently yields the lowest overall errors. Thus, the conclusions drawn in the main body for the case with \(z=10\) remain valid across the broader range of adaptation windows studied here.

\section{Governing equations for rotating detonation engine}
\label{app:app_rde}

For the one-dimensional rotating detonation engine case considered in Section~\ref{subsec:results_rde}, the full-order model is based on the one-dimensional conservation equations for mass, momentum, total energy, and species transport. In conservative form,
\begin{equation}
\frac{\partial \bm q}{\partial t}
+
\frac{\partial \bm f}{\partial x}
-
\frac{\partial \bm f_v}{\partial x}
=
\bm H.
\label{eq:rde_app_conservation}
\end{equation}
The conservative state vector is
\begin{equation}
\bm q
=
\begin{bmatrix}
\rho \\
\rho u \\
\rho h_0 - p \\
\rho Y_l
\end{bmatrix},
\label{eq:rde_app_state}
\end{equation}
where \(\rho\) is the density, \(u\) is the azimuthal velocity, \(p\) is the pressure, \(Y_l\) is the mass fraction of the \(l\)-th species, and the total enthalpy is
\begin{equation}
h_0 = h + \frac{1}{2}u^2
=
\sum_l h_l Y_l + \frac{1}{2}u^2.
\label{eq:rde_app_total_enthalpy}
\end{equation}

The inviscid and viscous fluxes are written as
\begin{equation}
\bm f
=
\begin{bmatrix}
\rho u \\
\rho u^2 + p \\
\rho u h_0 \\
\rho u Y_l
\end{bmatrix},
\qquad
\bm f_v
=
\begin{bmatrix}
0 \\
\tau_{xx} \\
u\tau_{xx} - q_x \\
\rho D_l \dfrac{\partial Y_l}{\partial x}
\end{bmatrix},
\label{eq:rde_app_fluxes}
\end{equation}
where \(D_l\) denotes the effective diffusion coefficient of the \(l\)-th species into the mixture.

The heat flux is
\begin{equation}
q_x
=
-K\frac{\partial T}{\partial x}
+
\sum_{l=1}^{N_s}
\rho D_l \frac{\partial Y_l}{\partial x} h_l
+
q_{\mathrm{source}},
\label{eq:rde_app_heat_flux}
\end{equation}
where \(K\) is the thermal conductivity and \(T\) is the temperature. The viscous stress is
\begin{equation}
\tau_{xx}
=
\frac{4}{3}\mu \frac{\partial u}{\partial x},
\label{eq:rde_app_stress}
\end{equation}
with \(\mu\) the molecular viscosity. The source vector contains the chemistry source terms in the species equations,
\begin{equation}
\bm H
=
\begin{bmatrix}
0 \\
0 \\
0 \\
\dot{\omega}_l
\end{bmatrix}.
\label{eq:rde_app_source}
\end{equation}

The one-dimensional RDE uses a periodic domain in the azimuthal direction together with a simplified injection model. Let \(p_{\mathrm{cr}}\) denote the critical pressure defined from the stagnation pressure \(p_0\) and reactant specific-heat ratio \(\gamma_R\) by
\begin{equation}
p_{\mathrm{cr}}
=
p_0
\left(
\frac{2}{\gamma_R+1}
\right)^{-\frac{\gamma_R}{\gamma_R-1}}.
\label{eq:rde_app_pcr}
\end{equation}
Then the inlet rule is modeled as
\begin{equation}
\begin{cases}
\text{no injection}, & p > p_{\mathrm{cr}}, \\[1mm]
\text{inject constant } (V_{\mathrm{in}},P_{\mathrm{in}},T_{\mathrm{in}}), & p \le p_{\mathrm{cr}},
\end{cases}
\label{eq:rde_app_injection_rule}
\end{equation}
with
\begin{equation}
T_{\mathrm{in}} = 300\,\mathrm{K}.
\label{eq:rde_app_tin}
\end{equation}

In the present configuration, combustion is represented using a detailed \(\mathrm{H_2\!-\!O_2}\) kinetic mechanism with \(10\) species and \(24\) reactions. The full-order simulations are initialized from a ramp-shaped profile and advanced explicitly using a four-stage third-degree strong-stability-preserving Runge--Kutta scheme. Spatial discretization is performed by a cell-centered second-order finite-volume method with Roe fluxes, a Barth--Jespersen limiter, and ghost-cell boundary treatment.

These equations and numerical ingredients define the full-order model from which the reduced models in Section~\ref{subsec:results_rde} are constructed and tested.

\section{Additional RDE results}
\label{app:app_rde_additional_results}

\begin{figure}
\centering
\includegraphics[width=0.235\linewidth]{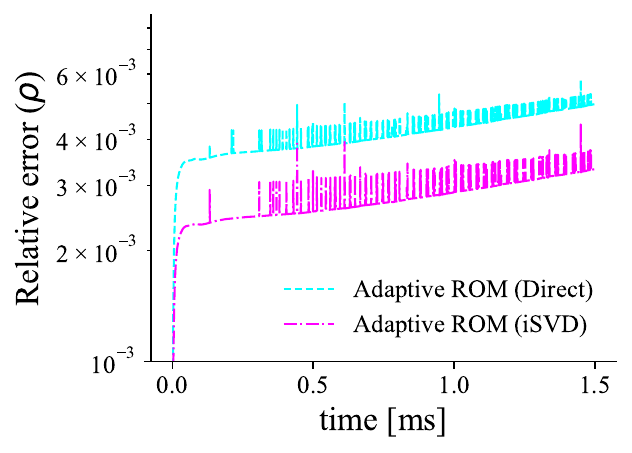}
\includegraphics[width=0.235\linewidth]{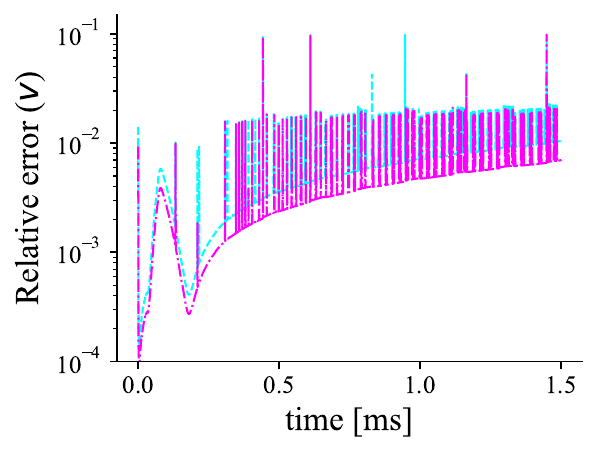}
\includegraphics[width=0.235\linewidth]{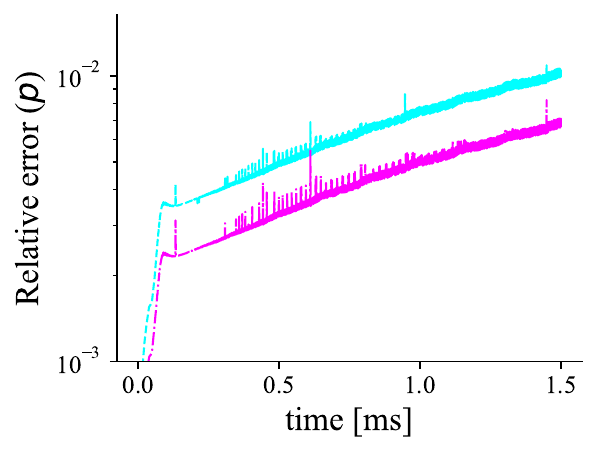}
\includegraphics[width=0.235\linewidth]{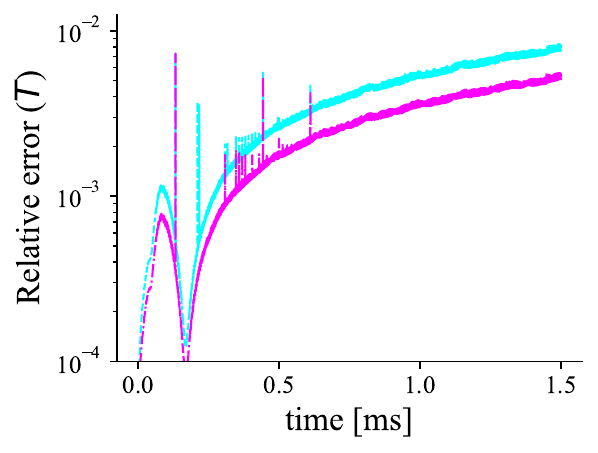}\\[2pt]

\includegraphics[width=0.235\linewidth]{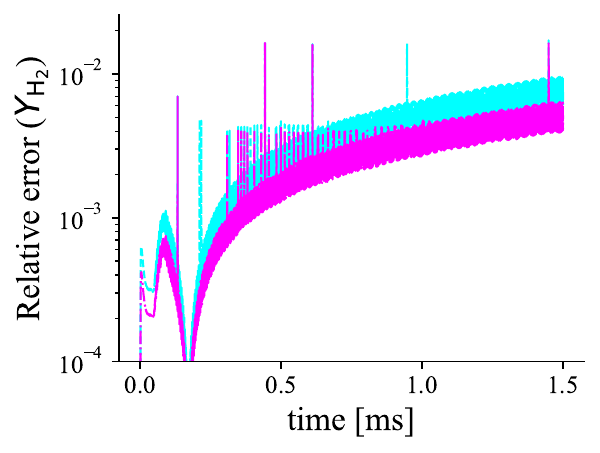}
\includegraphics[width=0.235\linewidth]{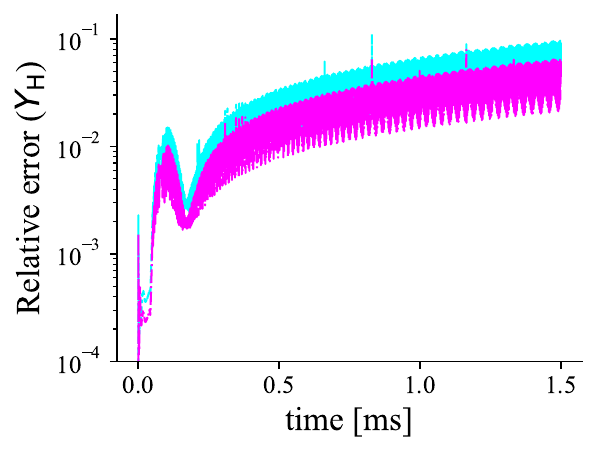}
\includegraphics[width=0.235\linewidth]{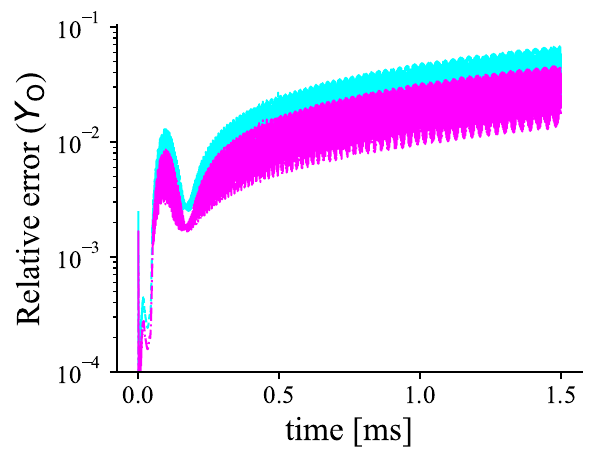}
\includegraphics[width=0.235\linewidth]{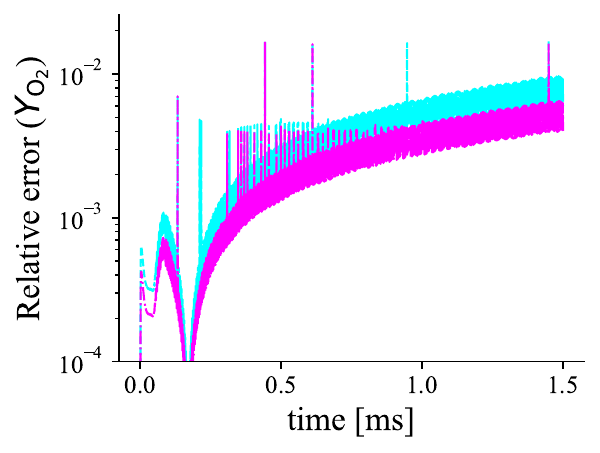}\\[2pt]

\includegraphics[width=0.235\linewidth]{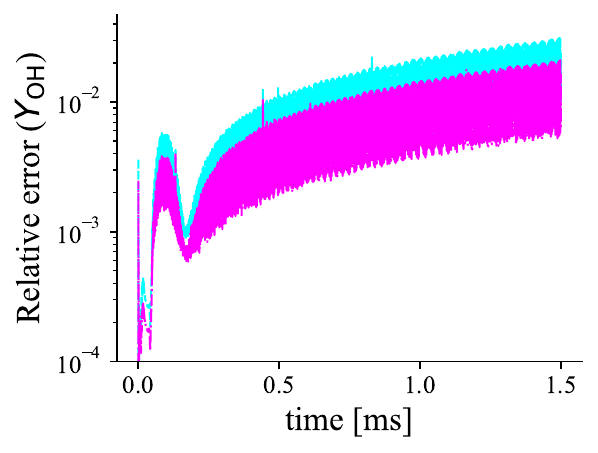}
\includegraphics[width=0.235\linewidth]{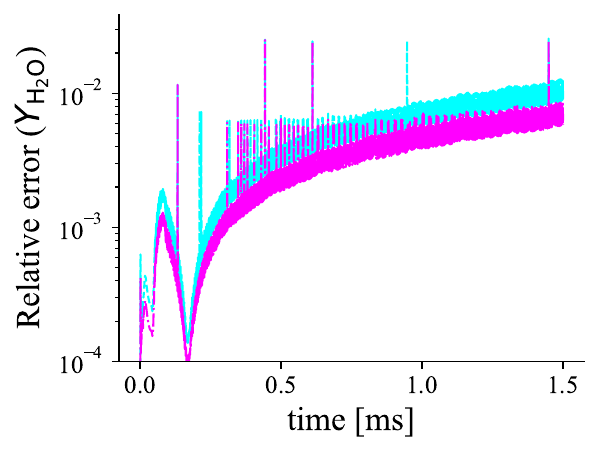}
\includegraphics[width=0.235\linewidth]{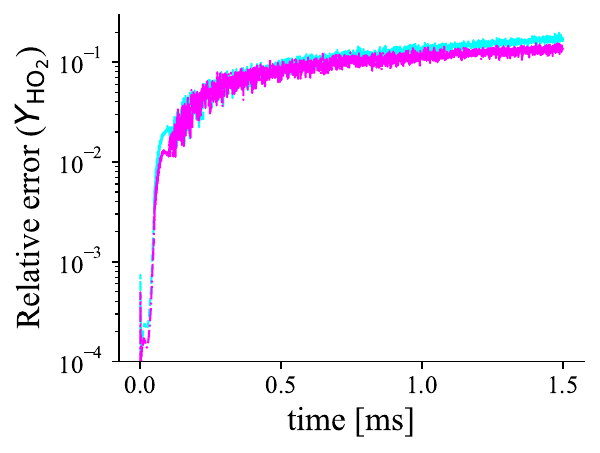}
\includegraphics[width=0.235\linewidth]{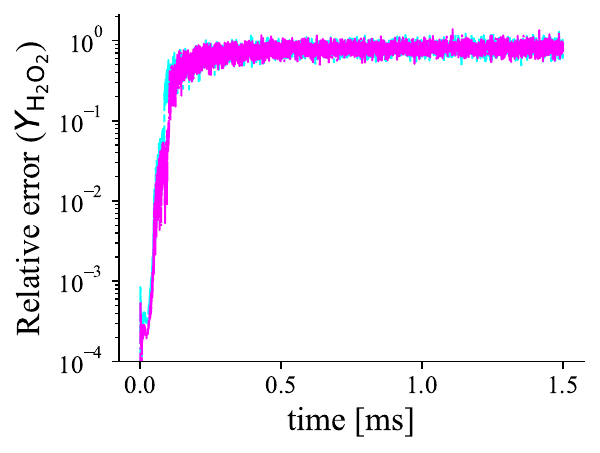}\\[2pt]

\includegraphics[width=0.235\linewidth]{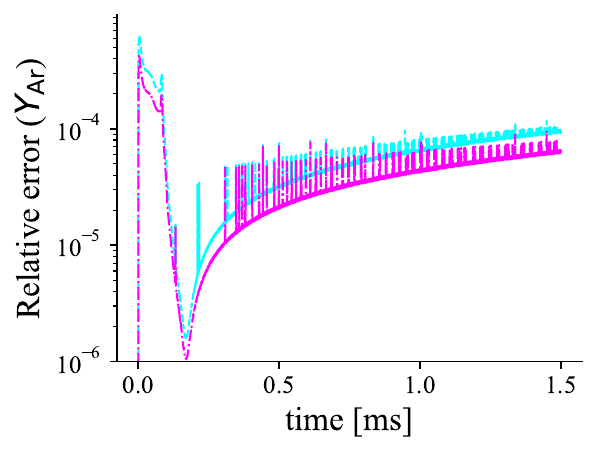}
\includegraphics[width=0.235\linewidth]{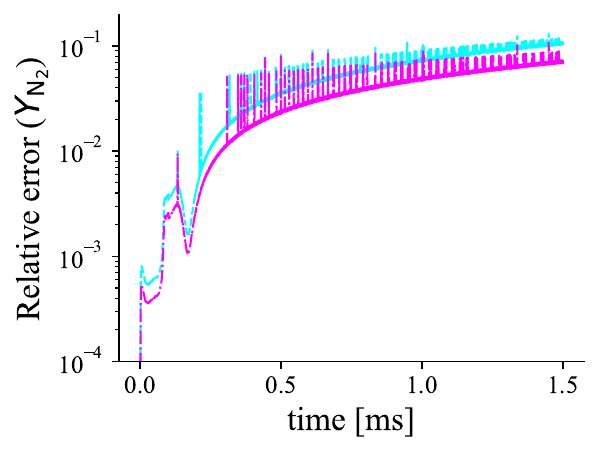}
\caption{Additional RDE relative error histories for \(z=5\). Both ROMs are trained on the interval \(t\in[0,5\times10^{-6}\,\mathrm{ms}]\), and tested over \(t\in[5\times10^{-6}\,\mathrm{ms},1.5\,\mathrm{ms}]\).}
\label{fig:app_rde_errors_z5}
\end{figure}

\begin{figure}
\centering
\includegraphics[width=0.235\linewidth]{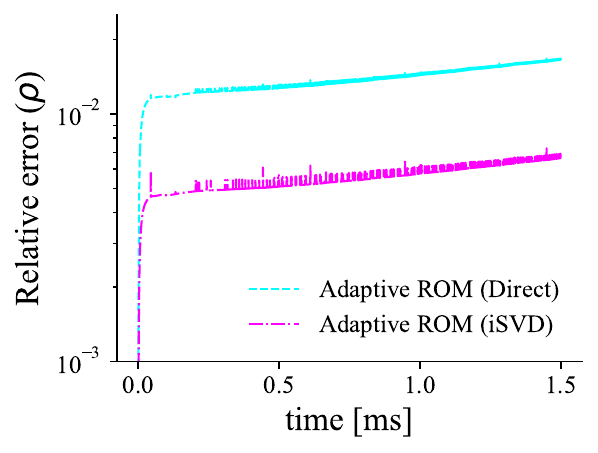}
\includegraphics[width=0.235\linewidth]{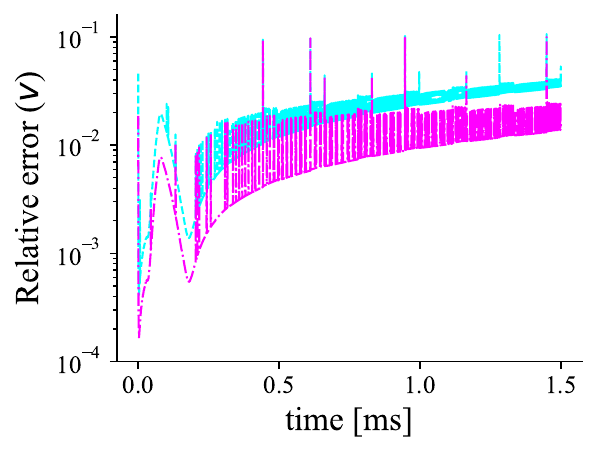}
\includegraphics[width=0.235\linewidth]{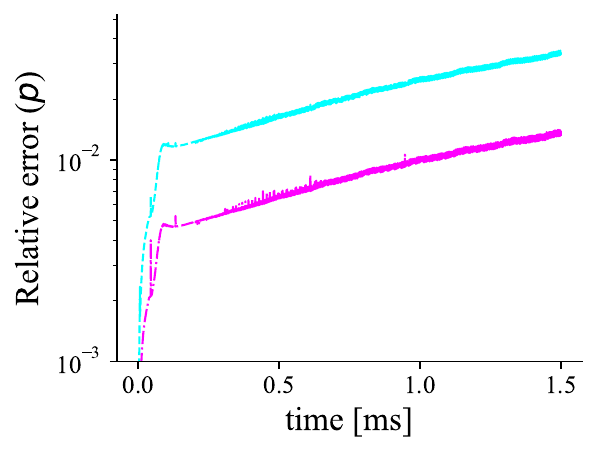}
\includegraphics[width=0.235\linewidth]{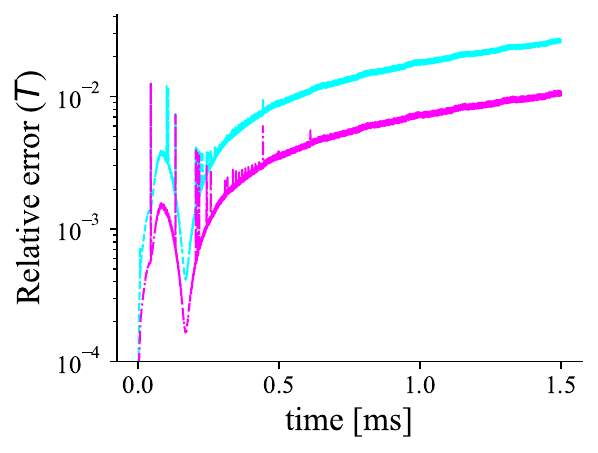}\\[2pt]

\includegraphics[width=0.235\linewidth]{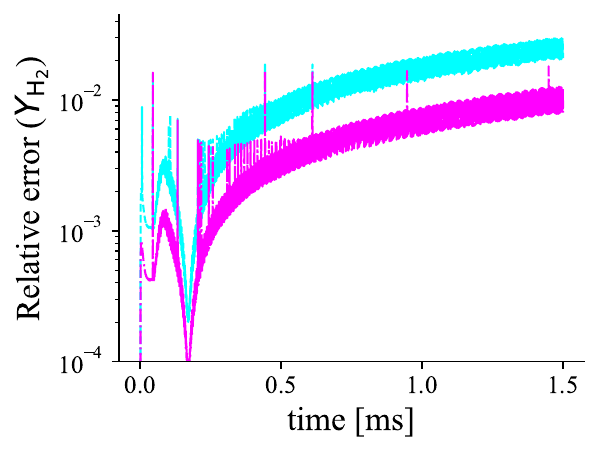}
\includegraphics[width=0.235\linewidth]{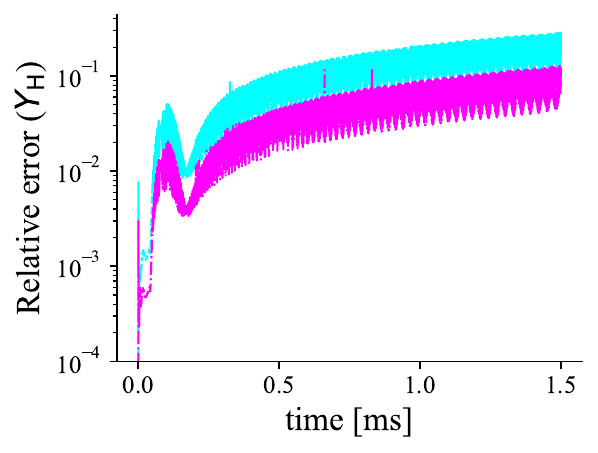}
\includegraphics[width=0.235\linewidth]{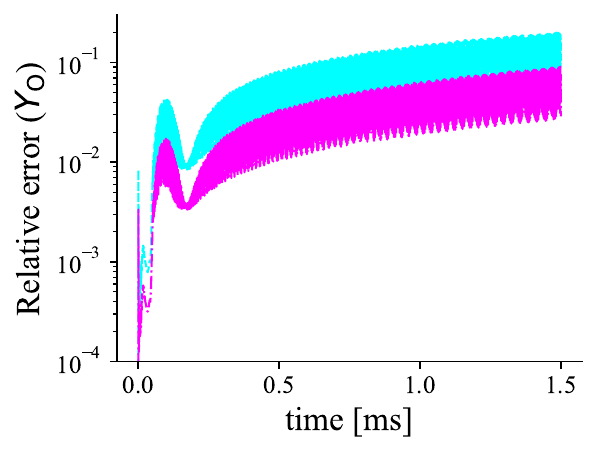}
\includegraphics[width=0.235\linewidth]{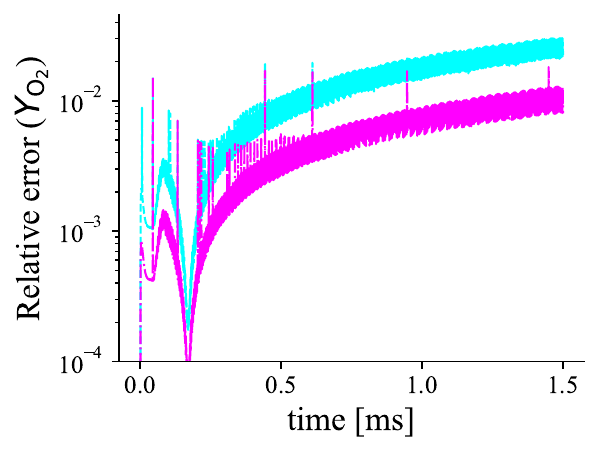}\\[2pt]

\includegraphics[width=0.235\linewidth]{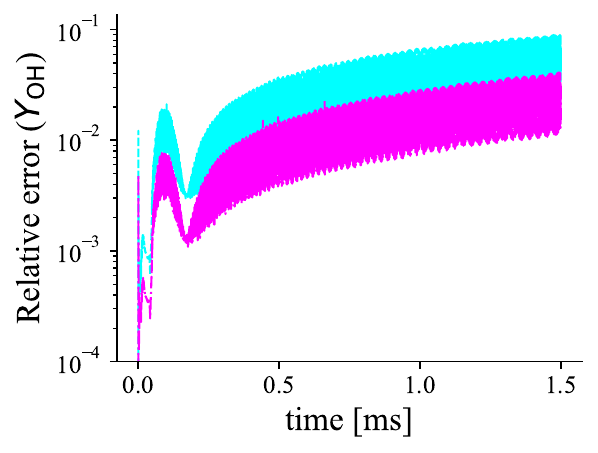}
\includegraphics[width=0.235\linewidth]{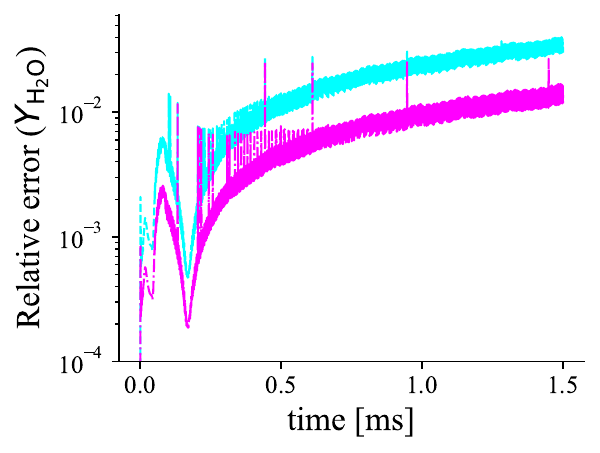}
\includegraphics[width=0.235\linewidth]{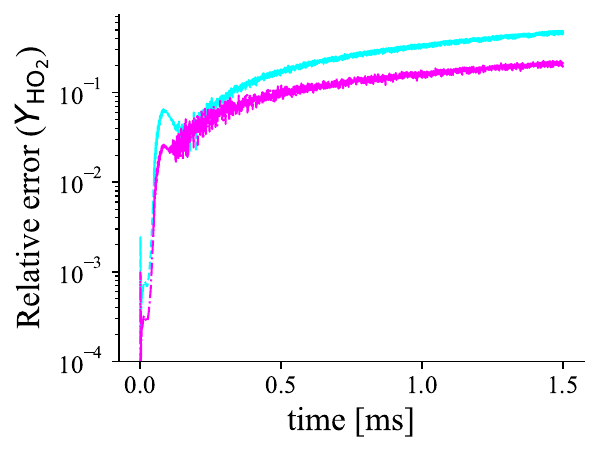}
\includegraphics[width=0.235\linewidth]{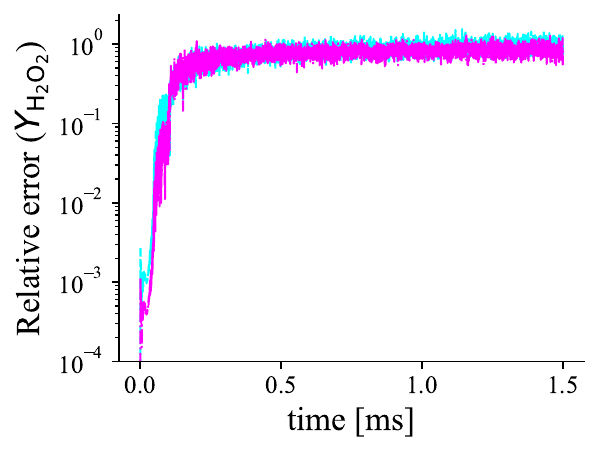}\\[2pt]

\includegraphics[width=0.235\linewidth]{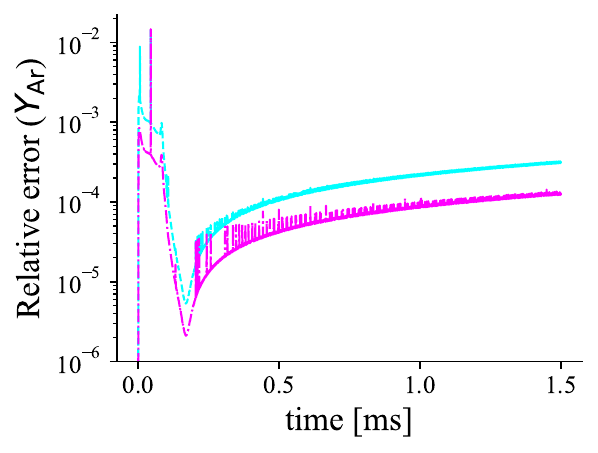}
\includegraphics[width=0.235\linewidth]{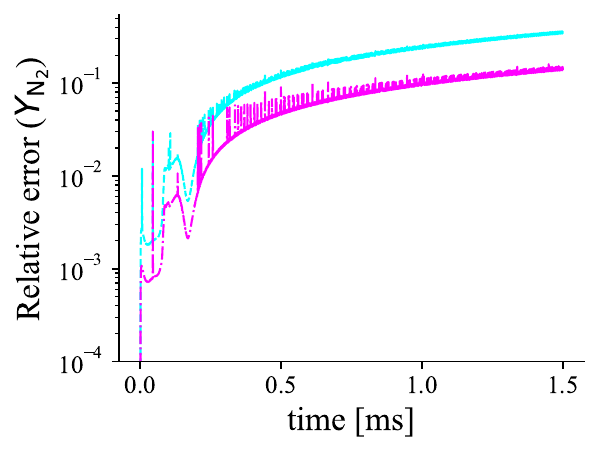}
\caption{Additional RDE relative error histories for \(z=10\). Both ROMs are trained on the interval \(t\in[0,5\times10^{-6}\,\mathrm{ms}]\), and tested over \(t\in[5\times10^{-6}\,\mathrm{ms},1.5\,\mathrm{ms}]\).}
\label{fig:app_rde_errors_z10}
\end{figure}

This appendix reports additional results for the one-dimensional RDE case. Here, we show results for all 14 monitored quantities: density, velocity, pressure, temperature, and the mass fractions of the 10 chemical species. Results are reported for the two adaptation windows \(z=5\) (Figures~\ref{fig:app_rde_profiles_z5} and \ref{fig:app_rde_errors_z5}) and \(z=10\) (Figures~\ref{fig:app_rde_profiles_z10} and \ref{fig:app_rde_errors_z10}). These supplementary figures support and extend the conclusions of the main text, showing that although both adaptive ROMs reproduce the dominant detonation dynamics, the proposed iSVD adaptive ROM remains more accurate than the Direct adaptive ROM across the broader set of variables.

\begin{figure}
\centering
\includegraphics[width=0.8\linewidth]{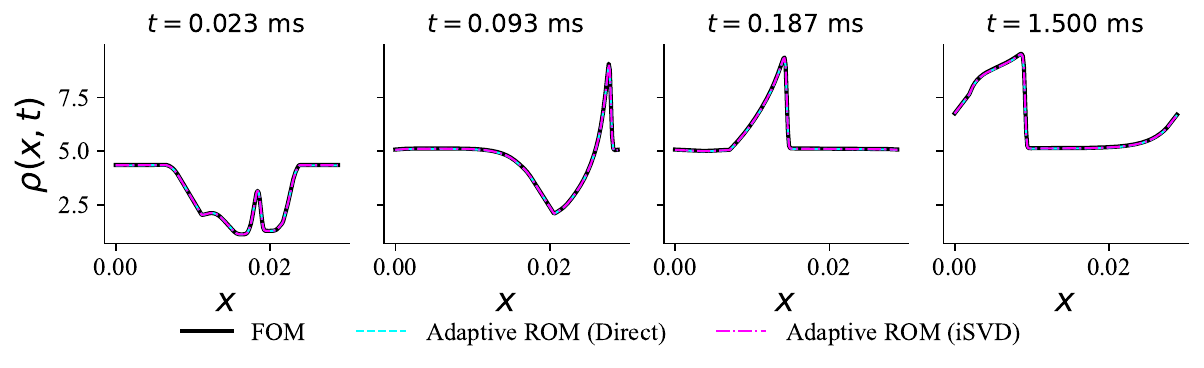}\\[2pt]
\includegraphics[width=0.8\linewidth]{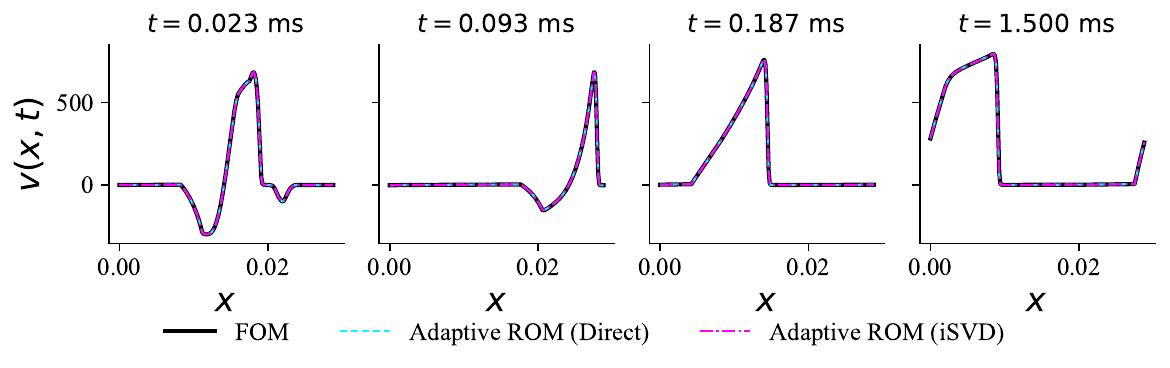}\\[2pt]
\includegraphics[width=0.8\linewidth]{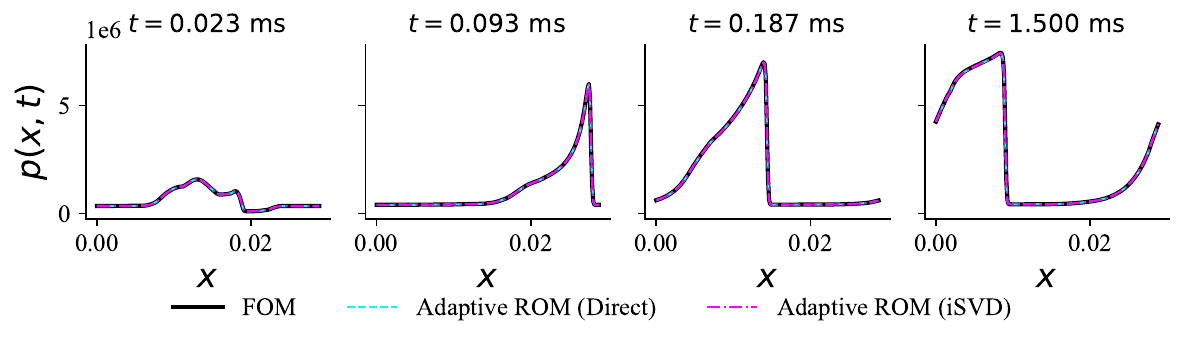}\\[2pt]
\includegraphics[width=0.8\linewidth]{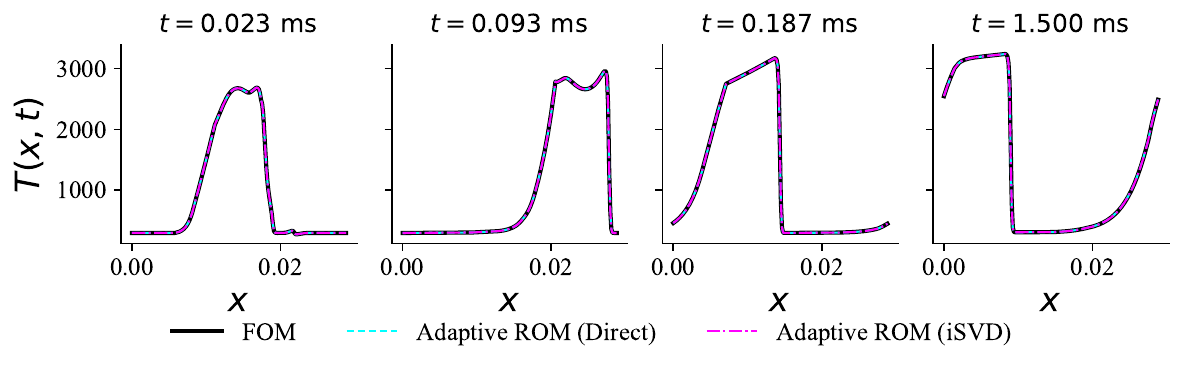}\\[2pt]
\includegraphics[width=0.8\linewidth]{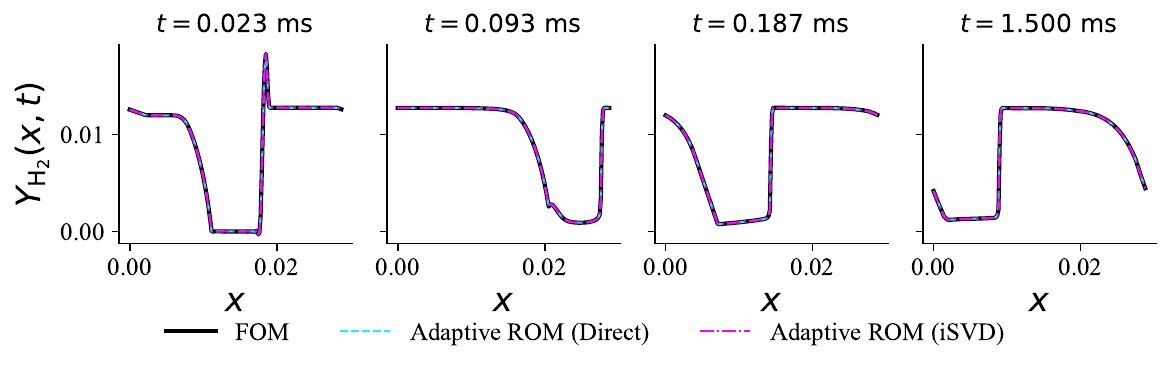}
\caption{Additional RDE solution profiles for \(z=5\). Both ROMs are trained on the interval \(t\in[0,5\times10^{-6}\,\mathrm{ms}]\), and tested over \(t\in[5\times10^{-6}\,\mathrm{ms},1.5\,\mathrm{ms}]\).}
\label{fig:app_rde_profiles_z5}
\end{figure}

\begin{figure}
\ContinuedFloat
\centering
\includegraphics[width=0.8\linewidth]{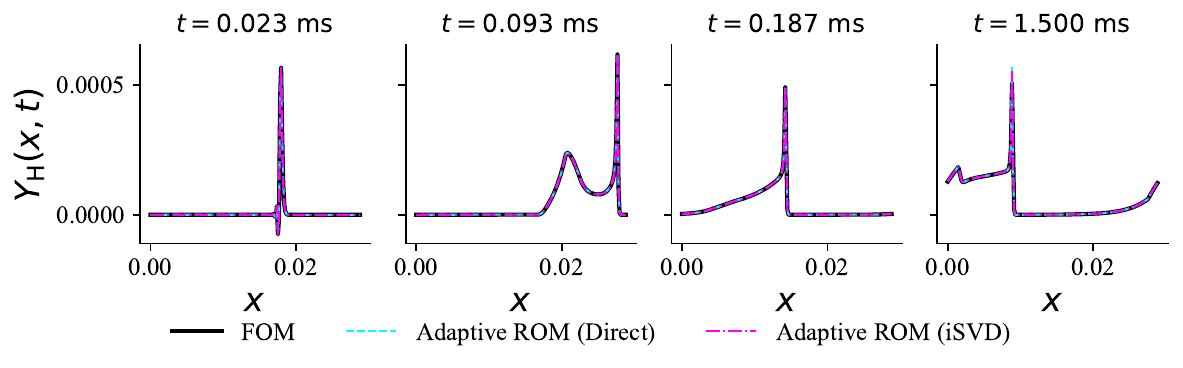}\\[2pt]
\includegraphics[width=0.8\linewidth]{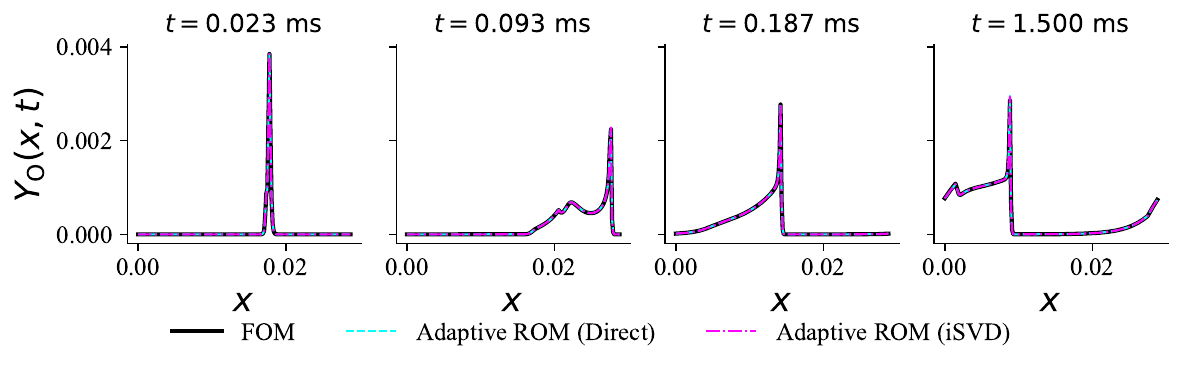}\\[2pt]
\includegraphics[width=0.8\linewidth]{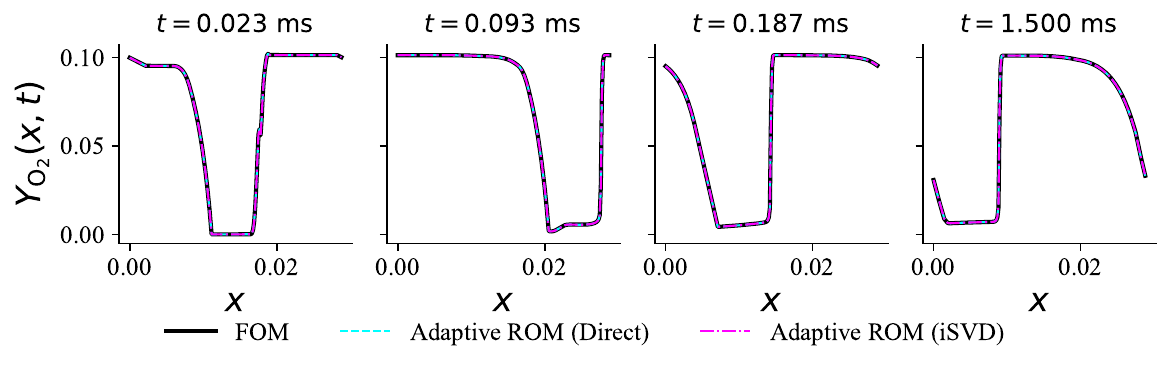}\\[2pt]
\includegraphics[width=0.8\linewidth]{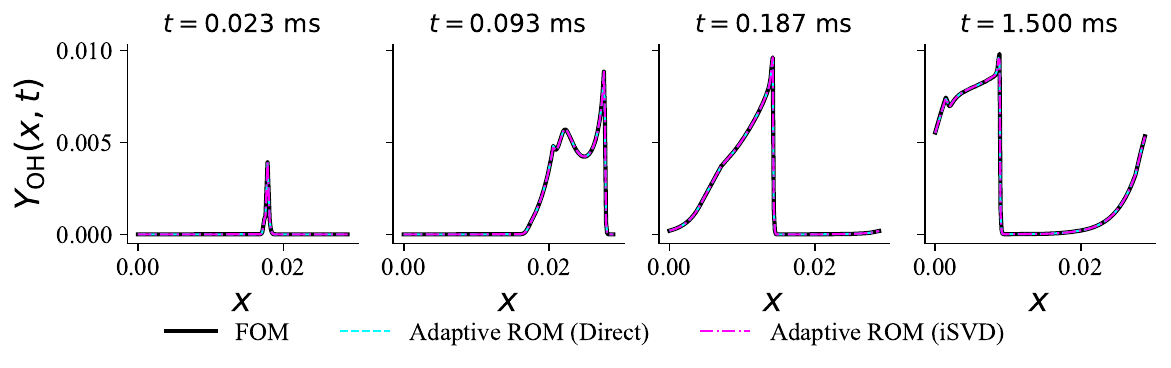}\\[2pt]
\includegraphics[width=0.8\linewidth]{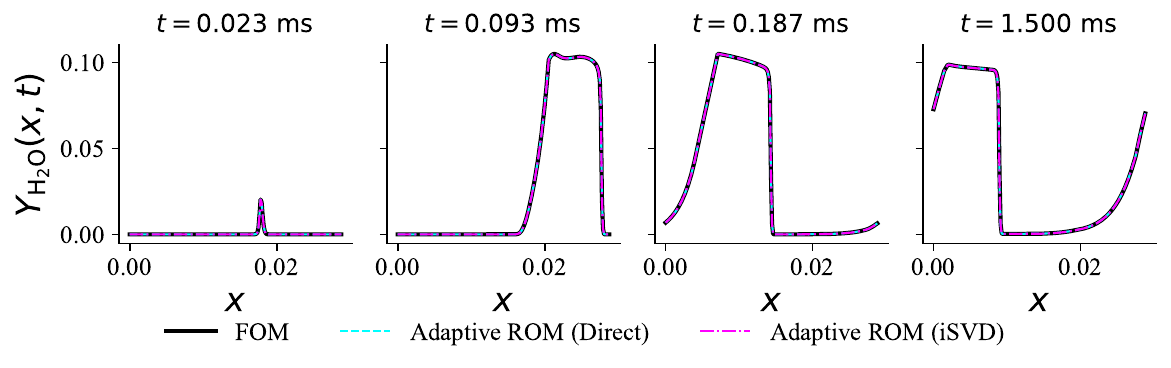}
\caption[]{Additional RDE solution profiles for \(z=5\). Both ROMs are trained on the interval \(t\in[0,5\times10^{-6}\,\mathrm{ms}]\), and tested over \(t\in[5\times10^{-6}\,\mathrm{ms},1.5\,\mathrm{ms}]\) (continued).}
\end{figure}

\begin{figure}
\ContinuedFloat
\centering
\includegraphics[width=0.8\linewidth]{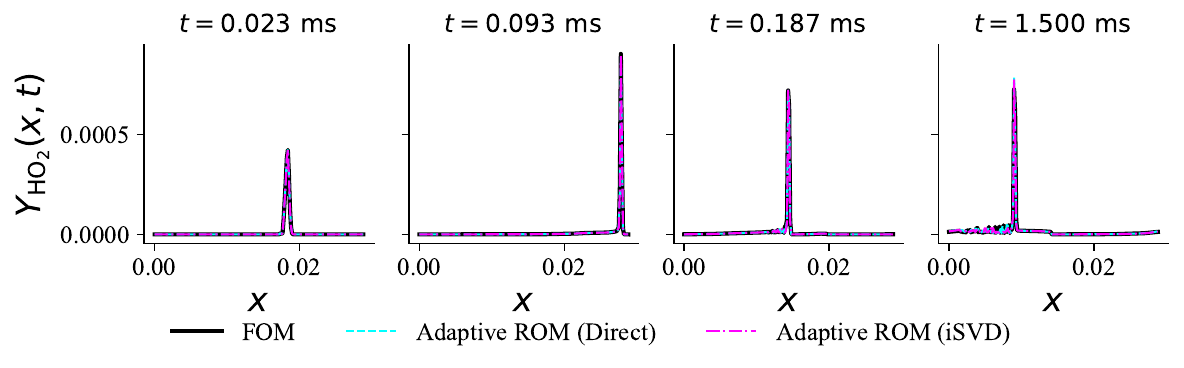}\\[2pt]
\includegraphics[width=0.8\linewidth]{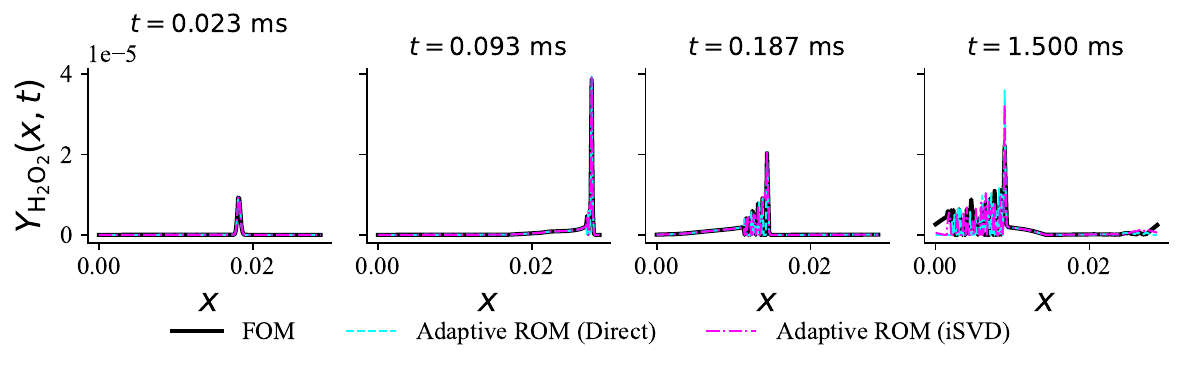}\\[2pt]
\includegraphics[width=0.8\linewidth]{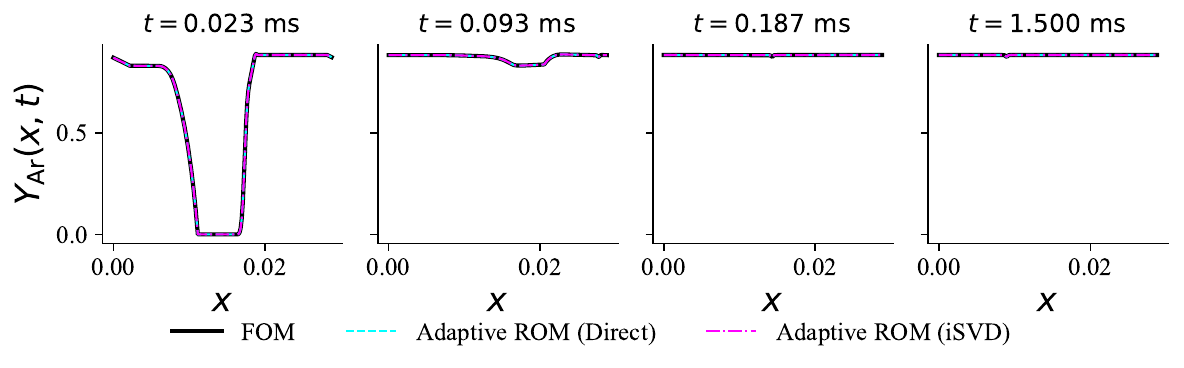}\\[2pt]
\includegraphics[width=0.8\linewidth]{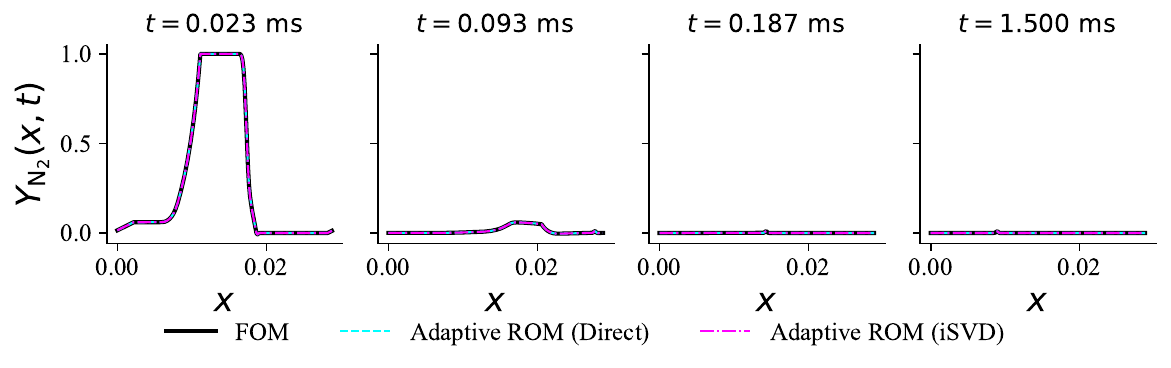}
\caption[]{Additional RDE solution profiles for \(z=5\). Both ROMs are trained on the interval \(t\in[0,5\times10^{-6}\,\mathrm{ms}]\), and tested over \(t\in[5\times10^{-6}\,\mathrm{ms},1.5\,\mathrm{ms}]\) (continued).}
\end{figure}

\begin{figure}
\centering
\includegraphics[width=0.8\linewidth]{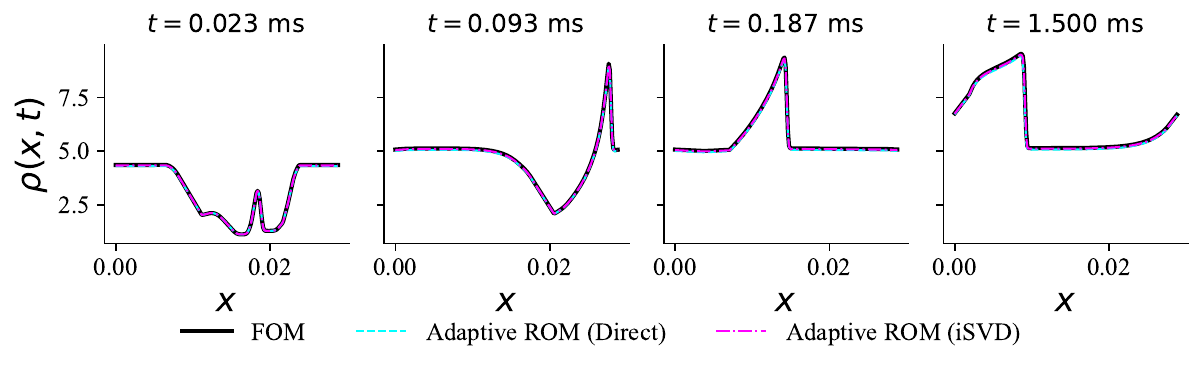}\\[2pt]
\includegraphics[width=0.8\linewidth]{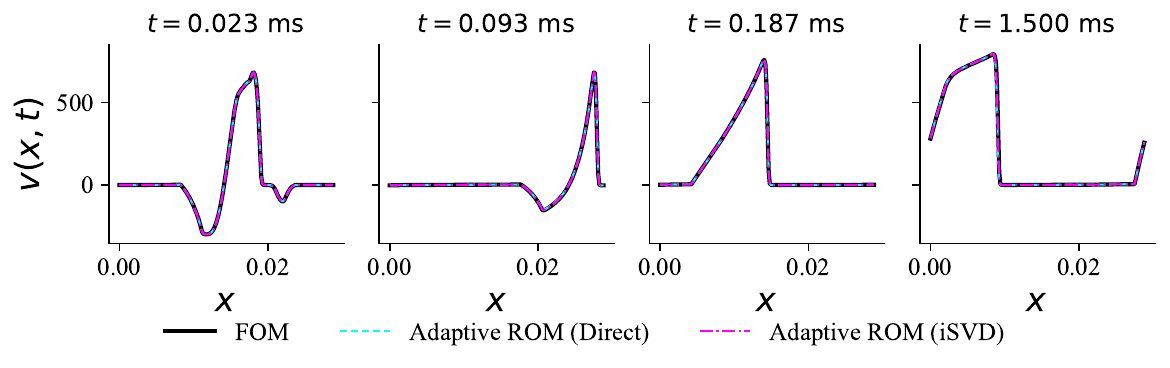}\\[2pt]
\includegraphics[width=0.8\linewidth]{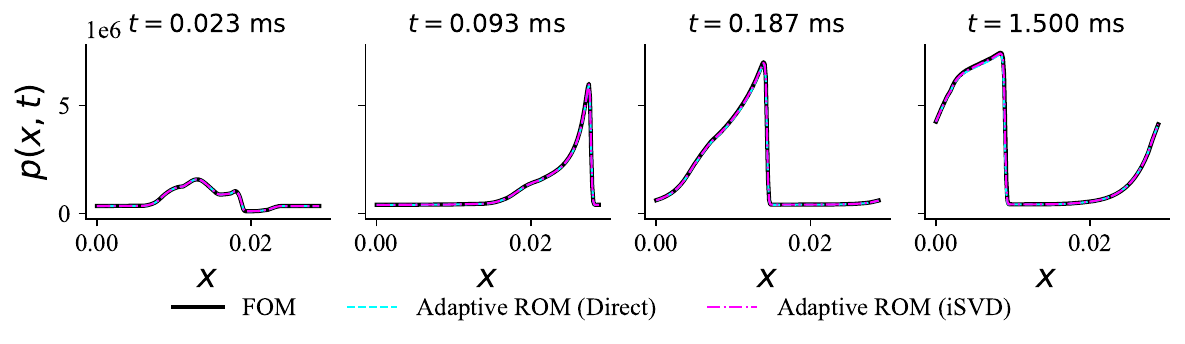}\\[2pt]
\includegraphics[width=0.8\linewidth]{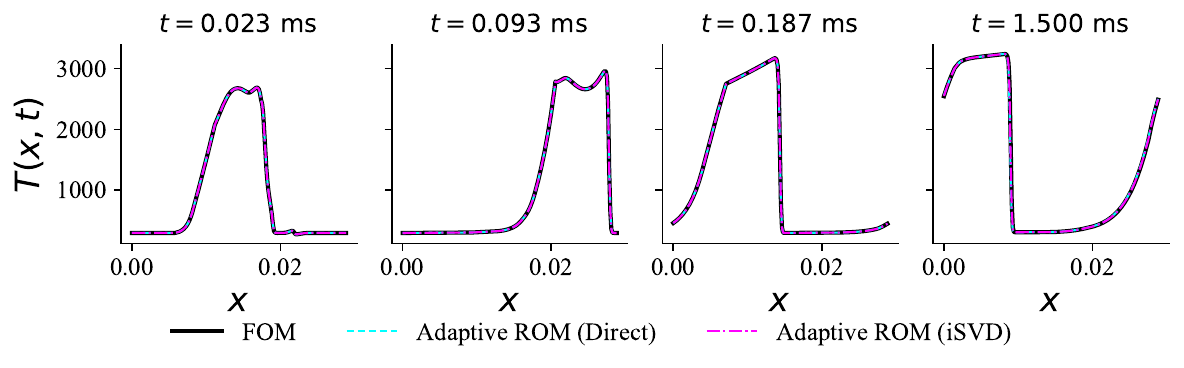}\\[2pt]
\includegraphics[width=0.8\linewidth]{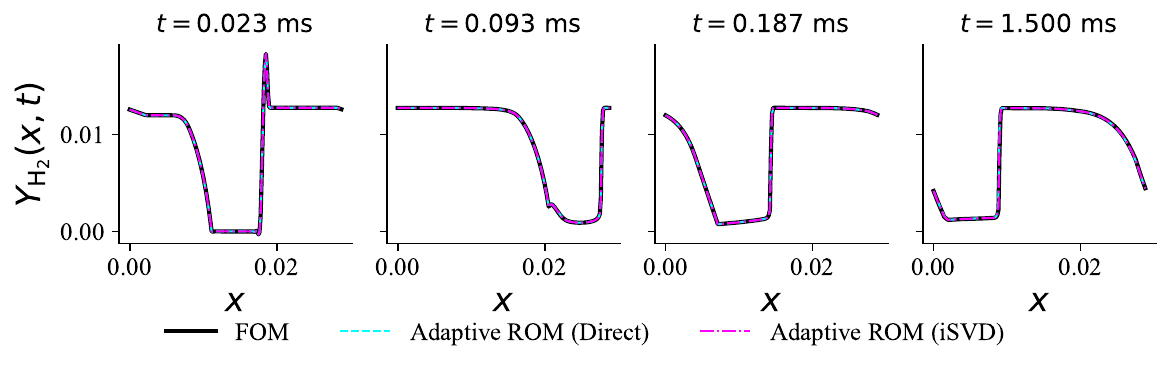}
\caption{Additional RDE solution profiles for \(z=10\). Both ROMs are trained on the interval \(t\in[0,5\times10^{-6}\,\mathrm{ms}]\), and tested over \(t\in[5\times10^{-6}\,\mathrm{ms},1.5\,\mathrm{ms}]\).}
\label{fig:app_rde_profiles_z10}
\end{figure}

\begin{figure}
\ContinuedFloat
\centering
\includegraphics[width=0.8\linewidth]{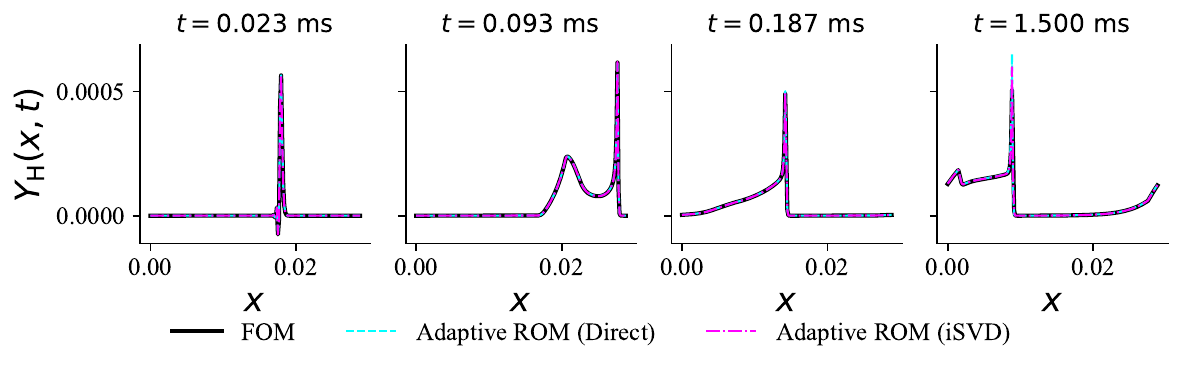}\\[2pt]
\includegraphics[width=0.8\linewidth]{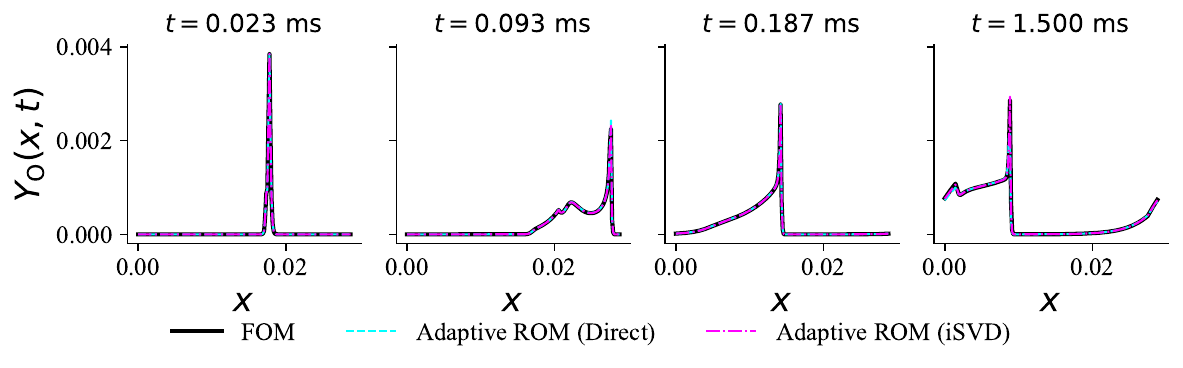}\\[2pt]
\includegraphics[width=0.8\linewidth]{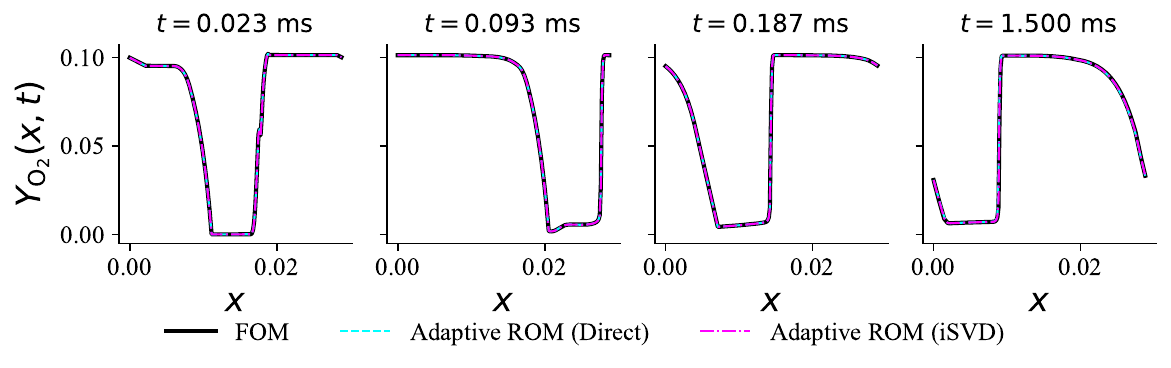}\\[2pt]
\includegraphics[width=0.8\linewidth]{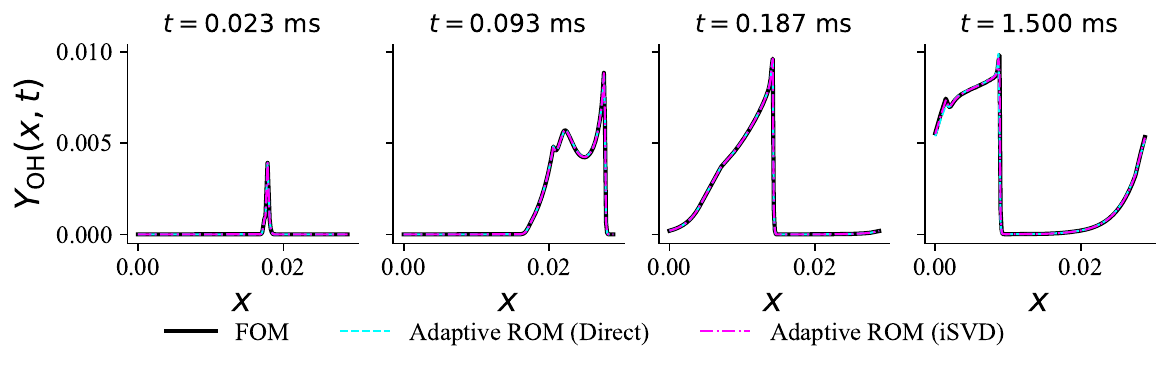}\\[2pt]
\includegraphics[width=0.8\linewidth]{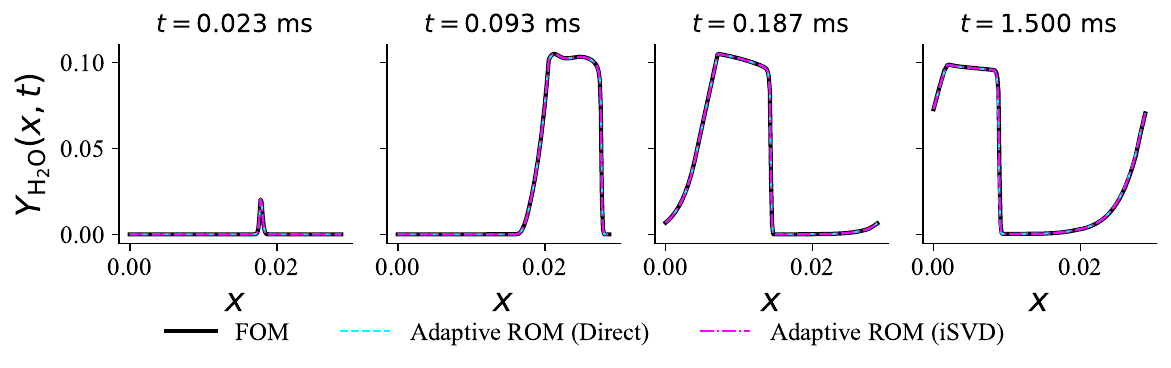}
\caption[]{Additional RDE solution profiles for \(z=10\). Both ROMs are trained on the interval \(t\in[0,5\times10^{-6}\,\mathrm{ms}]\), and tested over \(t\in[5\times10^{-6}\,\mathrm{ms},1.5\,\mathrm{ms}]\) (continued).}
\end{figure}

\begin{figure}
\ContinuedFloat
\centering
\includegraphics[width=0.8\linewidth]{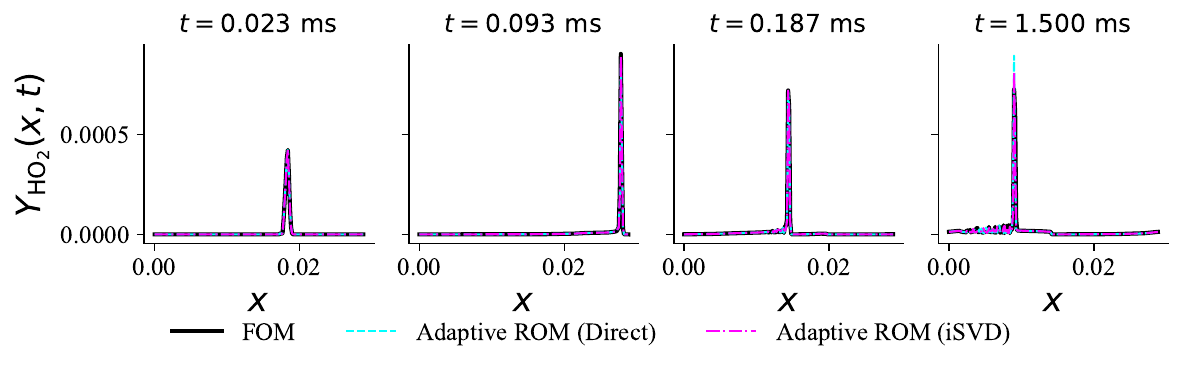}\\[2pt]
\includegraphics[width=0.8\linewidth]{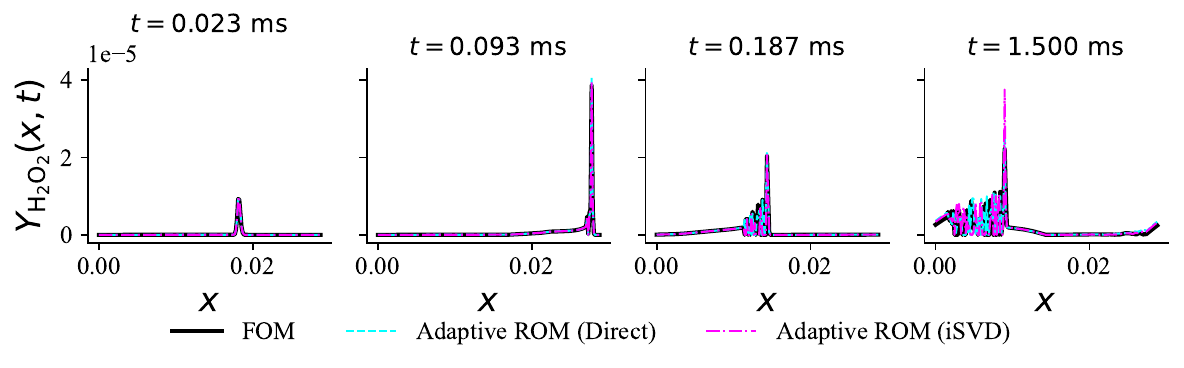}\\[2pt]
\includegraphics[width=0.8\linewidth]{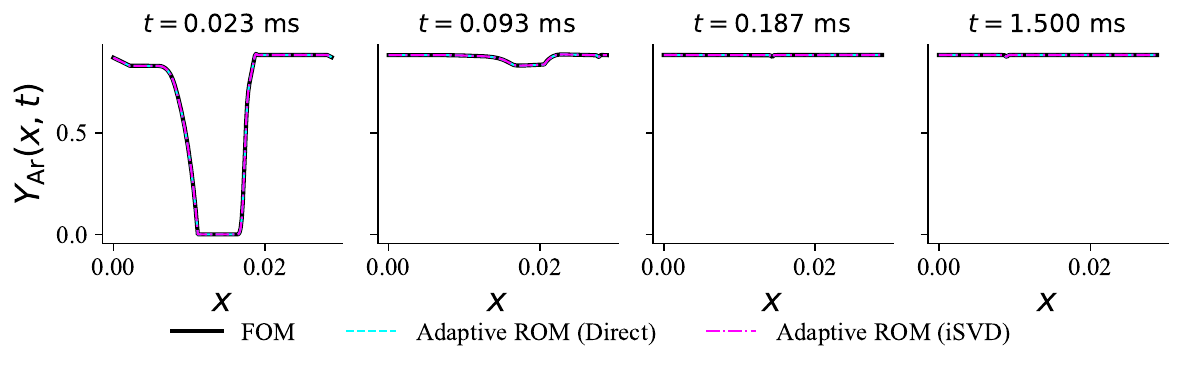}\\[2pt]
\includegraphics[width=0.8\linewidth]{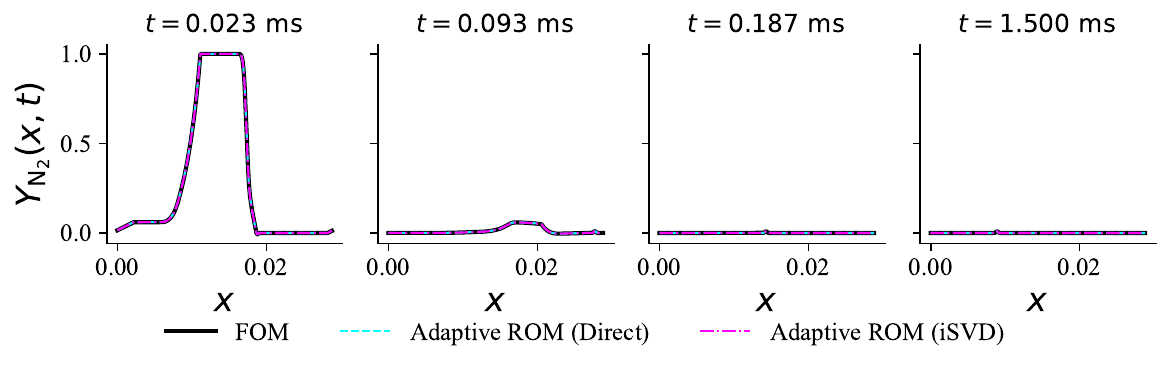}
\caption[]{Additional RDE solution profiles for \(z=10\). Both ROMs are trained on the interval \(t\in[0,5\times10^{-6}\,\mathrm{ms}]\), and tested over \(t\in[5\times10^{-6}\,\mathrm{ms},1.5\,\mathrm{ms}]\) (continued).}
\end{figure}

\clearpage

\bibliographystyle{elsarticle-num}
\bibliography{references}

\end{document}